\PassOptionsToPackage{table}{xcolor}
\documentclass[11pt]{article}
\usepackage{ACL2023}

\makeatletter
\AtBeginDocument{%
  \@twocolumnfalse            
  \let\orig@twocolumn\twocolumn
  \renewcommand{\twocolumn}[1][]{\onecolumn} 
  \onecolumn                  
}
\makeatother

\usepackage{microtype}
\usepackage{xcolor}
\usepackage{graphicx}
\usepackage{booktabs}
\usepackage{array}
\usepackage{tabularx}
\usepackage{longtable}
\usepackage{multirow}
\usepackage{float}
\usepackage{wrapfig}
\usepackage{enumitem}
\usepackage{amsmath,amssymb,breqn}
\usepackage{algorithm}
\usepackage{algpseudocode}
\usepackage{framed}
\usepackage{comment}
\usepackage{natbib}
\usepackage{multibib}
\usepackage{pifont}
\usepackage{arydshln}
\usepackage{tikz}
\usetikzlibrary{trees,shapes,shapes.geometric,arrows,decorations.markings,positioning,arrows.meta}
\usepackage[many]{tcolorbox}
\usepackage{subcaption}
\usepackage{varwidth}
\usepackage{setspace}
\usepackage{extsizes}
\usepackage{cuted}
\usepackage{flushend}
\usepackage{dblfloatfix}
\usepackage{fixltx2e}
\usepackage{soul}
\usepackage{hyperref}
\definecolor{darkblue}{rgb}{0,0,0.5}
\hypersetup{colorlinks=true, citecolor=darkblue, linkcolor=darkblue, urlcolor=darkblue}
\usepackage[capitalise,nameinlink]{cleveref}
\usepackage{wasysym} 

\usepackage[many]{tcolorbox}

\newtcolorbox{defin}{colback=Teal!5!White,enhanced,title=Alignment Faking: Bayesian–Stackelberg Equilibria,	attach boxed title to top left={xshift=-4mm},boxrule=0pt,after skip=1cm,before skip=1cm,right skip=0cm,breakable,fonttitle=\bfseries,toprule=0pt,bottomrule=0pt,rightrule=0pt,leftrule=3pt,arc=0mm,skin=enhancedlast jigsaw,sharp corners,colframe=Teal!55!black,colbacktitle=Teal!55!black,boxed title style={
		frame code={ 
			\fill[Teal!25!black](frame.south west)--(frame.north west)--(frame.north east)--([xshift=3mm]frame.east)--(frame.south east)--cycle;
			\draw[line width=1mm,Teal!25!black]([xshift=2mm]frame.north east)--([xshift=5mm]frame.east)--([xshift=2mm]frame.south east);
			\draw[line width=1mm,Teal!25!black]([xshift=5mm]frame.north east)--([xshift=8mm]frame.east)--([xshift=5mm]frame.south east);
			\fill[Teal!25!black](frame.south west)--+(4mm,-2mm)--+(4mm,2mm)--cycle;
		}
	}
}
\usetikzlibrary{shapes.geometric, arrows}
\usetikzlibrary{decorations.markings}

\definecolor{first}{RGB}{210,255,140}
\definecolor{second}{RGB}{136, 162, 190}
\definecolor{third}{RGB}{129, 222, 228}
\definecolor{fourth}{RGB}{132, 84, 246}
\definecolor{fifth}{RGB}{250, 223, 112}
\definecolor{sixth}{RGB}{203, 193, 172}
\definecolor{seventh}{RGB}{88, 112, 246}
\definecolor{eighth}{RGB}{245, 192, 106}
\definecolor{nine}{RGB}{171, 162, 111}
\definecolor{ten}{RGB}{217, 217, 217}

\definecolor{paired-light-blue}{RGB}{198, 219, 239}
\definecolor{paired-dark-blue}{RGB}{49, 130, 188}
\definecolor{paired-light-orange}{RGB}{251, 208, 162}
\definecolor{paired-dark-orange}{RGB}{230, 85, 12}
\definecolor{paired-light-green}{RGB}{199, 233, 193}
\definecolor{paired-dark-green}{RGB}{49, 163, 83}
\definecolor{paired-light-purple}{RGB}{218, 218, 235}
\definecolor{paired-dark-purple}{RGB}{117, 107, 176}
\definecolor{paired-light-gray}{RGB}{217, 217, 217}
\definecolor{paired-dark-gray}{RGB}{99, 99, 99}
\definecolor{paired-light-pink}{RGB}{222, 158, 214}
\definecolor{paired-dark-pink}{RGB}{123, 65, 115}
\definecolor{paired-light-red}{RGB}{231, 150, 156}
\definecolor{paired-dark-red}{RGB}{131, 60, 56}
\definecolor{paired-light-yellow}{RGB}{231, 204, 149}
\definecolor{paired-dark-yellow}{RGB}{141, 109, 49}
\definecolor{Teal}{RGB}{0, 50, 50}
\definecolor{White}{RGB}{250, 250, 250}
\definecolor{bg1}{HTML}{FF9966}
\definecolor{bg2}{HTML}{CCE5FF}
\definecolor{bg3}{HTML}{FFCC99}
\definecolor{bg4}{HTML}{FFC107}
\definecolor{bg5}{HTML}{FFCCCC}
\definecolor{bg6}{HTML}{D5E8D4}
\definecolor{bg7}{HTML}{eeeeee}
\definecolor{bg8}{HTML}{cdeb8b}
\definecolor{bg9}{HTML}{dae8fc}
\definecolor{bg10}{HTML}{a2e6eb}
\definecolor{bg31}{HTML}{FFCDD2} 
\definecolor{bg32}{HTML}{F8BBD0}
\definecolor{bg33}{HTML}{E1BEE7} 
\definecolor{bg34}{HTML}{D7CCC8} 
\definecolor{bg35}{HTML}{B2DFDB} 
\definecolor{bg36}{HTML}{A5D6A7} 
\definecolor{bg37}{HTML}{FFF9C4} 
\definecolor{bg38}{HTML}{FFECB3} 
\definecolor{bg111}{HTML}{CB6843}
\definecolor{bg112}{HTML}{D77C5C}
\definecolor{bg113}{HTML}{E28E6E}
\definecolor{bg114}{HTML}{E89F7D}
\definecolor{bg115}{HTML}{EDAE8A}
\definecolor{bg116}{HTML}{F0BA95}
\definecolor{bg117}{HTML}{F3C29F}
\definecolor{bg118}{HTML}{F6CCAA}
\definecolor{bg119}{HTML}{F8D5B3}
\definecolor{bg120}{HTML}{FADCBD}
\definecolor{bg121}{HTML}{FCE6C7}
\definecolor{bg39}{HTML}{FFE0B2} 
\definecolor{bg40}{HTML}{3CB371} 

\definecolor{bg43}{HTML}{ffe5d9}
\definecolor{bg15}{HTML}{7FFFD4}
\definecolor{bg17}{HTML}{F0FFFF}
\definecolor{bg18}{HTML}{F5FFFA}
\definecolor{bg19}{HTML}{F8F8FF}
\definecolor{bg20}{HTML}{FFFFFF}
\definecolor{bg21}{HTML}{E1F5FE}
\definecolor{bg22}{HTML}{B3E5FC}
\definecolor{bg23}{HTML}{81D4FA}
\definecolor{bg24}{HTML}{4FC3F7}
\definecolor{bg25}{HTML}{29B6F6}
\definecolor{bg26}{HTML}{03A9F4}
\definecolor{bg27}{HTML}{039BE5}
\definecolor{bg28}{HTML}{0288D1}
\definecolor{bg29}{HTML}{0277BD}
\definecolor{bg30}{HTML}{01579B}
\definecolor{bg16}{HTML}{FFCC99} 
\definecolor{pg51}{HTML}{E8F5E9} 
\definecolor{pg52}{HTML}{C8E6C9} 
\definecolor{pg53}{HTML}{B9F6CA} 
\definecolor{pg54}{HTML}{A9DFBF} 
\definecolor{pg55}{HTML}{BCF5A6} 
\definecolor{pg56}{HTML}{BEF1CE} 
\definecolor{pg57}{HTML}{CEF6EC} 
\definecolor{pg58}{HTML}{B7F0B1} 
\definecolor{pg59}{HTML}{B1F2B5} 
\definecolor{pg60}{HTML}{9DF3C4} 
\definecolor{pg61}{HTML}{DEF7E0} 
\definecolor{pg62}{HTML}{E8F8DC} 
\definecolor{pg63}{HTML}{EBF7E7} 
\definecolor{pg64}{HTML}{F0FDF4} 
\definecolor{pg65}{HTML}{F1FEE7} 
\definecolor{pg66}{HTML}{F7FFF6} 
\definecolor{pg67}{HTML}{FCFFE7} 
\definecolor{pg68}{HTML}{F4FFD2} 
\definecolor{pg69}{HTML}{EEFFE2} 
\definecolor{pg70}{HTML}{E3FDF5} 
\definecolor{connect-color}{RGB}{0,0,0}
\definecolor{middle-color}{RGB}{255,255,255}
\definecolor{leaf-color}{RGB}{173,216,230}
\definecolor{line-color}{RGB}{25,25,112}
\definecolor{soothingPurple}{RGB}{195, 160, 201}
\definecolor{hidden-draw}{RGB}{20,68,106}
\definecolor{hidden-pink}{RGB}{255,245,247}
\definecolor{dark-red}{RGB}{233, 150, 122}
\definecolor{light-red}{RGB}{255,182,193}
\definecolor{medium-red}{RGB}{205,92,92}
\definecolor{light-yellow}{RGB}{255, 239, 153}
\definecolor{light-blue}{RGB}{173, 216, 230}
\definecolor{paired-light-yellow}{HTML}{FFFF88}
\definecolor{paired-light-blue}{HTML}{CCE5FF}
\definecolor{paired-light-orange}{HTML}{FFCC99}
\definecolor{paired-dark-yellow}{HTML}{FFF2CC}
\definecolor{paired-light-pink}{HTML}{FFCCCC}
\definecolor{paired-cyan}{HTML}{D5E8D4}
\definecolor{paired-gray}{HTML}{eeeeee}
\definecolor{paired-green}{HTML}{cdeb8b}
\definecolor{paired-blue}{HTML}{dae8fc}
\definecolor{paired-dark-cyan}{HTML}{a2e6eb}
\definecolor{paired-dark-pink}{HTML}{e7b2d2}
\definecolor{paired-purple}{HTML}{9999ff}
\definecolor{paired-pink}{HTML}{cc99ff}
\definecolor{paired-orange}{HTML}{ffcc99}

\definecolor{a1}{RGB}{241,233,191}
\definecolor{a2}{RGB}{255,241,218}

\definecolor{a3}{RGB}{255,239,213}
\definecolor{a4}{RGB}{250,235,215}
\definecolor{a5}{RGB}{255,239,219}
\definecolor{a6}{RGB}{255,246,225}
\definecolor{a7}{RGB}{246,227,201}
\definecolor{a8}{RGB}{254,235,226}
\definecolor{a9}{RGB}{247,220,111}
\definecolor{a10}{RGB}{199,211,189}
\definecolor{a11}{RGB}{209,196,233}
\definecolor{a12}{RGB}{214,234,248}
\definecolor{a13}{RGB}{232,245,233}
\definecolor{a14}{RGB}{237,248,177}
\definecolor{a15}{RGB}{255,228,225}
\definecolor{a16}{RGB}{255,228,181}
\definecolor{a17}{RGB}{255,222,173}
\definecolor{a18}{RGB}{255,218,185}
\definecolor{a19}{RGB}{255,203,164}
\definecolor{a20}{RGB}{247,202,201}

\definecolor{a21}{RGB}{241,254,255}
\definecolor{a22}{RGB}{230,252,252}
\definecolor{a23}{RGB}{179,236,255}
\definecolor{a24}{RGB}{174,226,249}
\definecolor{a25}{RGB}{208,234,246}
\definecolor{a26}{RGB}{189,226,219}
\definecolor{a27}{RGB}{177,204,201}

\definecolor{a28}{RGB}{216,195,216}
\definecolor{a29}{RGB}{195,155,211}
\definecolor{a30}{RGB}{208,152,223}
\definecolor{a31}{RGB}{255,183,209}
\definecolor{a32}{RGB}{255,167,209}
\definecolor{a33}{RGB}{254,235,167}
\definecolor{a34}{RGB}{255,222,137}
\definecolor{a35}{RGB}{254,180,154}
\definecolor{a36}{RGB}{247,148,161}
\definecolor{a37}{RGB}{239,154,154}
\definecolor{a38}{RGB}{255,130,171}
\definecolor{a39}{RGB}{255,105,180}
\definecolor{a40}{RGB}{251,142,172}

\newtcolorbox{societal_harm}{
  colback=soothingPurple, 
  colframe=black, 
  boxrule=0pt,
  enhanced,
  title=Societal harm,
  attach boxed title to top right={yshift=-3mm},
  fonttitle=\bfseries,
  toprule=1pt,
  bottomrule=1pt,
  rightrule=1pt,
  leftrule=1pt,
  arc=1mm
}

\newtcolorbox{privacy_violation}{
  colback=soothingPurple, 
  colframe=black, 
  boxrule=0pt,
  enhanced,
  title=Privacy Violation,
  attach boxed title to top right={yshift=-3mm},
  fonttitle=\bfseries,
  toprule=1pt,
  bottomrule=1pt,
  rightrule=1pt,
  leftrule=1pt,
  arc=1mm
}

\newtcolorbox{disinformation_deception}{
  colback=soothingPurple, 
  colframe=black, 
  boxrule=0pt,
  enhanced,
  title=Disinformation \& Deception,
  attach boxed title to top right={yshift=-3mm},
  fonttitle=\bfseries,
  toprule=1pt,
  bottomrule=1pt,
  rightrule=1pt,
  leftrule=1pt,
  arc=1mm
}

\newtcolorbox{answer_disparity}{
  colback=soothingPurple, 
  colframe=black, 
  boxrule=0pt,
  enhanced,
  title=Answer disparity,
  attach boxed title to top right={yshift=-3mm},
  fonttitle=\bfseries,
  toprule=1pt,
  bottomrule=1pt,
  rightrule=1pt,
  leftrule=1pt,
  arc=1mm
}

\newtcolorbox{wrong_classification}{
  colback=soothingPurple, 
  colframe=black, 
  boxrule=0pt,
  enhanced,
  title=Wrong classification,
  attach boxed title to top right={yshift=-3mm},
  fonttitle=\bfseries,
  toprule=1pt,
  bottomrule=1pt,
  rightrule=1pt,
  leftrule=1pt,
  arc=1mm
}

\newtcolorbox{goal_hijacking}{
  colback=soothingPurple, 
  colframe=black, 
  boxrule=0pt,
  enhanced,
  title=Goal hijacking,
  attach boxed title to top right={yshift=-3mm},
  fonttitle=\bfseries,
  toprule=1pt,
  bottomrule=1pt,
  rightrule=1pt,
  leftrule=1pt,
  arc=1mm
}

\newtcolorbox{control_generation}{
  colback=soothingPurple, 
  colframe=black, 
  boxrule=0pt,
  enhanced,
  title=Control generation,
  attach boxed title to top right={yshift=-3mm},
  fonttitle=\bfseries,
  toprule=1pt,
  bottomrule=1pt,
  rightrule=1pt,
  leftrule=1pt,
  arc=1mm
}

\newtcolorbox{prompt_leaking}{
  colback=soothingPurple, 
  colframe=black, 
  boxrule=0pt,
  enhanced,
  title=Prompt leaking,
  attach boxed title to top right={yshift=-3mm},
  fonttitle=\bfseries,
  toprule=1pt,
  bottomrule=1pt,
  rightrule=1pt,
  leftrule=1pt,
  arc=1mm
}

\usepackage{lipsum}
\usepackage{tikz}
\usetikzlibrary{trees,shapes}

\usepackage{times}
\usepackage{latexsym}

\usepackage[T5]{fontenc}

\usepackage[utf8]{inputenc}

\usepackage{microtype}

\usepackage{inconsolata}
\usepackage{soul}

\usepackage{inconsolata}
\usepackage{algorithm}
\usepackage{algpseudocode}
\usepackage{amsmath,amssymb}
\usepackage{soul}
\usepackage{tikz}

\tikzset{rndblock/.style={rounded corners,rectangle,draw,scale=0.8,outer sep=0pt}}

\usetikzlibrary{shapes.geometric}
\usepackage{framed}
\usepackage{enumitem}
\newlist{RQ}{enumerate}{1}
\setlist[RQ]{label=\textbf{RQ\,\arabic*},ref={RQ\,\arabic*}}
\usepackage{comment}
\usepackage{natbib}
\usepackage{multibib}
\makeatletter
\usepackage{booktabs}
\usepackage[inkscapeformat=png]{svg}
\usepackage{graphicx}
\usepackage{caption}
\usepackage{subcaption}
\usepackage{tabularx}
\usepackage{soul}
\usepackage{float}
\usepackage{enumitem}
\usepackage{pifont}
\usepackage{arydshln}
\usepackage{lipsum}

\usepackage[many]{tcolorbox}

\usetikzlibrary{shapes.geometric, arrows}
\usetikzlibrary{decorations.markings}

\usepackage{fancybox}

\usepackage{hyperref}
 \definecolor{darkblue}{rgb}{0, 0, 0.5}
  \hypersetup{colorlinks=true, citecolor=darkblue, linkcolor=darkblue, urlcolor=darkblue}

\definecolor{vgreen}{HTML}{60A917}
\definecolor{vred}{HTML}{CE3A29}

\usepackage{xstring}
\usepackage{longtable}

\usepackage{tabularray}

\DefTblrTemplate{firsthead,middlehead,lasthead}{default}{
}
\DefTblrTemplate{firstfoot}{default}{
  \UseTblrTemplate{contfoot}{default}
  \UseTblrTemplate{caption}{default}
}
\DefTblrTemplate{middlefoot}{default}{
  \UseTblrTemplate{contfoot}{default}
  \UseTblrTemplate{capcont}{default}
}
\DefTblrTemplate{lastfoot}{default}{
  \UseTblrTemplate{note}{default}
  \UseTblrTemplate{remark}{default}
  \UseTblrTemplate{capcont}{default}
}

\newcolumntype{P}[1]{>{\centering\arraybackslash}p{#1}}

\usepackage{color}
\tcbuselibrary{skins}

\usepackage[export]{adjustbox} 

\usepackage{setspace}
\usepackage[capitalise,nameinlink]{cleveref}

\crefname{section}{Sec.}{Sec.}

\usepackage{microtype}
\usepackage{hyperref}
\usepackage{graphicx}
\usepackage{comment}
\usepackage{amsmath}
\usepackage{amssymb}
\usepackage{algorithm}
\usepackage{algpseudocode}
\usepackage{colortbl}
\usepackage[export]{adjustbox} 
\usepackage{varwidth}
\usepackage{enumitem}
\setlist{leftmargin=1mm}
\usepackage{pifont}
\usepackage{booktabs}
\usepackage{multirow}
\usepackage{subcaption}
\usepackage{resizegather}
\usepackage{breqn}
\usepackage[capitalise]{cleveref}
\usepackage{graphicx}
\usepackage{tikz}
\usetikzlibrary{shapes.geometric, arrows}
\usetikzlibrary{decorations.markings}
\usepackage{soul}
\usepackage{wrapfig,graphicx,lipsum}
\usepackage{extsizes}
\usepackage{cuted}
\usepackage{flushend}
\usepackage{float}
\usepackage{changepage,threeparttable}
\usepackage{setspace}
\usepackage{caption}
\usepackage{booktabs}
\usepackage{dblfloatfix} 
\usepackage{fixltx2e}
\usepackage[normalem]{ulem}

\usepackage{environ}
\usepackage{soul}
\usepackage{float}
\usepackage{enumitem}
\usepackage{pifont}
\usepackage{arydshln}
\usepackage{lipsum}

\usepackage[many]{tcolorbox}

\usetikzlibrary{shapes.geometric, arrows}
\usetikzlibrary{decorations.markings}

\usepackage{fancybox}

\usepackage{hyperref}
 \definecolor{darkblue}{rgb}{0, 0, 0.5}
  \hypersetup{colorlinks=true, citecolor=darkblue, linkcolor=darkblue, urlcolor=darkblue}

\definecolor{vgreen}{HTML}{60A917}
\definecolor{vred}{HTML}{CE3A29}

\usepackage{xstring}
\usepackage{longtable}

\usepackage{tabularray}

\DefTblrTemplate{firsthead,middlehead,lasthead}{default}{
}
\DefTblrTemplate{firstfoot}{default}{
  \UseTblrTemplate{contfoot}{default}
  \UseTblrTemplate{caption}{default}
}
\DefTblrTemplate{middlefoot}{default}{
  \UseTblrTemplate{contfoot}{default}
  \UseTblrTemplate{capcont}{default}
}
\DefTblrTemplate{lastfoot}{default}{
  \UseTblrTemplate{note}{default}
  \UseTblrTemplate{remark}{default}
  \UseTblrTemplate{capcont}{default}
}

\usepackage{color}
\tcbuselibrary{skins}

\usepackage[export]{adjustbox} 

\usepackage{setspace}
\usepackage[capitalise,nameinlink]{cleveref}

\crefname{section}{Sec.}{Sec.}

\usepackage{microtype}
\usepackage{hyperref}
\usepackage{graphicx}
\usepackage{comment}
\usepackage{amsmath}
\usepackage{amssymb}
\usepackage{algorithm}
\usepackage{algpseudocode}
\usepackage{colortbl}
\usepackage[export]{adjustbox} 
\usepackage{varwidth}
\usepackage{enumitem}
\setlist{leftmargin=1mm}
\usepackage{pifont}
\usepackage{booktabs}
\usepackage{multirow}
\usepackage{subcaption}
\usepackage{resizegather}
\usepackage{breqn}
\usepackage[capitalise]{cleveref}
\usepackage{graphicx}
\usepackage{tikz}
\usetikzlibrary{shapes.geometric, arrows}
\usetikzlibrary{decorations.markings}
\usepackage{soul}
\usepackage{wrapfig,graphicx,lipsum}
\usepackage{extsizes}
\usepackage{cuted}
\usepackage{flushend}
\usepackage{float}
\usepackage{changepage,threeparttable}
\usepackage{setspace}
\usepackage{caption}
\usepackage{booktabs}
\usepackage{dblfloatfix} 
\usepackage{fixltx2e}
\usepackage[normalem]{ulem}

\usepackage{tcolorbox}
\usepackage{amsmath}
\usepackage{amssymb}
\usepackage{caption}

\usepackage{environ}

\newlength{\myl}
\expandafter\let\expandafter\origequation\csname equation*\endcsname
\expandafter\let\expandafter\endorigequation\csname endequation*\endcsname
\long\def\[#1\]{\begin{equation*}#1\end{equation*}}
\RenewEnviron{equation*}{
  \settowidth{\myl}{$\displaystyle\BODY$} 
  \origequation
    \ifdim\myl>\linewidth
      \resizebox{\linewidth}{!}{$\displaystyle\BODY$}
    \else
      \BODY 
    \fi
  \endorigequation
}

\makeatletter
\newcommand{\DrawLine}{%
  \begin{tikzpicture}
  \path[use as bounding box] (0,0) -- (\linewidth,0);
  \draw[color=blue!75!black,dashed,dash phase=.5pt]
        (0-\kvtcb@leftlower-\kvtcb@boxsep,0)--
        (\linewidth+\kvtcb@rightlower+\kvtcb@boxsep,0);
  \end{tikzpicture}%
  }
\makeatother

\usepackage{pgfplots}
\usepackage{tikz}
\usetikzlibrary{positioning, arrows.meta}
\usepackage{longtable}
\usepackage{booktabs}
\usepackage{array}
\usepackage{adjustbox}
\usepackage{longtable}
\usepackage{pifont}

\usepackage{soul}            
\usepackage{fontawesome5}    
\IfFileExists{fontawesome5.sty}{
  \usepackage{fontawesome5} 
}{
  
}

\definecolor{algoPurple}{HTML}{6A51A3}
\definecolor{algoBlue}{HTML}{1F77B4}
\definecolor{algoGreen}{HTML}{2E8B57}
\definecolor{algoOrange}{HTML}{E67E22}

\makeatletter
\@ifundefined{faBalanceScale}{
  \@ifundefined{faCalculator}{
    \@ifundefined{faChartLine}{
    }{}
  }{}
}{}

\@ifundefined{faBrain}{
  \@ifundefined{faLightbulb}{
    \@ifundefined{faCompass}{
      
    }{}
  }{}
}{}

\@ifundefined{faUsersCog}{
  \@ifundefined{faUsers}{
    \@ifundefined{faProjectDiagram}{
      
    }{}
  }{}
}{}

\@ifundefined{faThumbsUp}{
  \@ifundefined{faHandshake}{
    \@ifundefined{faCheckDouble}{
      
    }{}
  }{}
}{}
\makeatother

\definecolor{AbsBack}{HTML}{EEF2FF}   
\definecolor{AbsFrame}{HTML}{5A67D8}  
\definecolor{AbsTitle}{HTML}{3B49B1}  

\newtcolorbox{abstractbox}{
  enhanced, breakable,
  colback=AbsBack, colframe=AbsFrame!85,
  boxrule=0.7pt,
  borderline={0.5pt}{0pt}{AbsFrame!40},
  arc=8pt, left=10pt, right=10pt, top=10pt, bottom=2pt,
  drop fuzzy shadow=AbsFrame!25
}

\newcommand{\AbstractTitle}{\textbf{\textcolor{AbsTitle}{\fontsize{18}{18}\selectfont Abstract}}}

\usepackage{iftex}
\ifPDFTeX
  \PackageError{pragya-fonts}{You must compile with XeLaTeX or LuaLaTeX}{}
\fi

\usepackage{fontspec}
\defaultfontfeatures{Ligatures=TeX, Scale=MatchLowercase}

\newfontfamily\PragyaHeadline[
  Path=fonts/,
  UprightFont = PragyaHeadline-Regular.ttf,
  BoldFont    = PragyaHeadline-Bold.ttf
]{Pragya Headline}

\usepackage{titlesec}

\titleformat{\section}
  {\PragyaHeadline\bfseries\Large\color{AbsTitle}}
  {\textcolor{AbsTitle}{\thesection}}
  {0.6em}
  {}

\titleformat{\subsection}
  {\PragyaHeadline\bfseries\large\color{AbsTitle}}
  {\textcolor{AbsTitle}{\thesubsection}}
  {0.5em}
  {}

\titleformat{\subsubsection}
  {\PragyaHeadline\bfseries\normalsize\color{AbsTitle}}
  {\textcolor{AbsTitle}{\thesubsubsection}}
  {0.5em}
  {}

\titlespacing*{\section}{0pt}{1.0ex plus .2ex}{0.6ex}
\titlespacing*{\subsection}{0pt}{0.8ex plus .2ex}{0.4ex}
\titlespacing*{\subsubsection}{0pt}{0.6ex plus .1ex}{0.3ex}

\usepackage{empheq}              
\usepackage[normalem]{ulem}      

\usepackage{amsthm}        

\usepackage{colortbl}
\usepackage{booktabs}
\usepackage{multirow}

\definecolor{paired-light-blue}{HTML}{C6DBEF}
\definecolor{paired-mid-blue}{HTML}{6BAED6}
\definecolor{paired-dark-blue}{HTML}{2171B5}

\definecolor{paired-light-orange}{HTML}{FFE5CC}
\definecolor{paired-mid-orange}{HTML}{FFCC99}
\definecolor{paired-dark-orange}{HTML}{FB9A29}

\usepackage{tikz}
\usetikzlibrary{matrix,positioning,calc,arrows.meta}

\usepackage{pgfplots}
\pgfplotsset{compat=1.18} 
\usepackage{tikz}
\usetikzlibrary{arrows.meta,decorations.pathreplacing,positioning}
\usepackage{float} 
\usepgfplotslibrary{groupplots}
\usepackage{algorithm}
\usepackage{algpseudocode} 

\usepackage{algorithmicx}
\usepackage{algpseudocode}

\tcbset{
  sphalgo/.style={
    breakable,
    enhanced,
    colback=white,
    colframe=black,
    boxrule=0.8pt,
    arc=2mm,
    left=1.2mm,right=1.2mm,top=1.2mm,bottom=1.2mm,
    before skip=6pt, after skip=6pt,
    fonttitle=\bfseries,
    colbacktitle=black,  
    coltitle=white,      
    title={#1},
  }
}

\algrenewcommand\algorithmicindent{1.0em}
\makeatletter
\renewcommand\ALG@name{Algorithm}
\makeatother

\title{\textcolor{white}{.}}
  
\usepackage{amsfonts}
\usepackage{nicefrac}
\usepackage{mathtools}
\usepackage{tocloft}
\definecolor{appendixpurple}{RGB}{156,81,152}
\newcommand{\AppendixMainEntry}[3]{%
    \noindent
    {\color{appendixpurple}\bfseries\large
    \hyperref[#1]{#2}}%
    \nobreak\leaders\hbox to 0.6em{\hss.\hss}\hfill
    {\bfseries\hyperref[#1]{\pageref*{#1}}}\par\vspace{0.6em}
}
\newcommand{\AppendixSubEntry}[3]{%
    \noindent\hspace*{2.2em}%
    {\color{appendixpurple}\bfseries\Large #2}\hspace{0.8em}%
    {\color{appendixpurple}\hyperref[#1]{#3}}%
    \nobreak\leaders\hbox to 0.6em{\hss.\hss}\hfill
    \hyperref[#1]{\pageref*{#1}}\par\vspace{0.35em}
}

\begin{document}

\begin{figure*}[t]
  \centering
  \includegraphics[width=.98\linewidth]{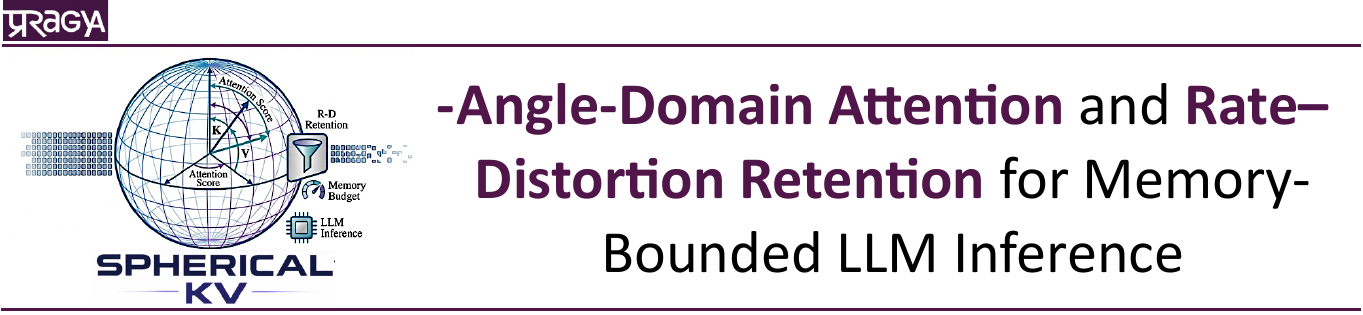}
  \vspace{-1.5em}
\end{figure*}

\renewcommand{\thefootnote}{\fnsymbol{footnote}}

\begin{center}
{\large\bfseries
Anay Chauhan\footnote{Work does not relate to author's position at Synopsys}$^{1}$\quad
Gurucharan Marthi Krishna Kumar$^{2}$\quad
Arion Das$^{3}$\quad
Amit Dhanda\footnote{Work does not relate to author's position at Amazon}$^{4}$\\[4pt]
Vinija Jain\footnote{Work does not relate to author's position at Meta, USA}$^{5}$\quad
Aman Chadha\footnote{Work does not relate to author's position at Apple, USA}$^{6}$\quad
Amitava Das$^{7}$
}\\[10pt]

{\normalsize
$^{1}$Synopsys, India \quad
$^{2}$McGill University, Canada \quad
$^{3}$IIIT Ranchi, India \quad
$^{4}$Amazon, USA \\[2pt]
$^{5}$Meta, USA \quad
$^{6}$Apple, USA\quad
$^{7}$Pragya Lab, BITS Pilani Goa, India\\[4pt]
}
\end{center}

\renewcommand{\thefootnote}{\arabic{footnote}}
\setcounter{footnote}{0}

\vspace{-0.5em}

\begin{abstractbox}
  \AbstractTitle\\

Long-context inference is increasingly constrained by the \textbf{KV cache}: resident memory grows with context length, and decoding becomes limited by repeated \textbf{High Bandwidth Memory (HBM)} streaming rather than arithmetic. Existing methods such as eviction, windowing, quantization, and offloading reduce footprint, but often leave the critical-path bottleneck only partially addressed, especially when compressed states must still be reconstructed into dense vectors during decoding.

We present \textbf{Spherical KV}, a long-context inference method that treats KV allocation as a \textbf{rate--distortion} problem grounded in \textbf{attention geometry for efficient decoding}. The method is built on two ideas: \textbf{(i)} represent directional information cheaply in the decode hot loop, and \textbf{(ii)} allocate retention and precision according to estimated future utility. Its first component, \textbf{Angle-Domain Attention (ADA)}, stores keys in a spherical parameterization consisting of a scalar radius and compact angle codes, and computes attention logits directly from these codes without reconstructing dense keys. This preserves a paged, block-local, fusion-friendly decode path and directly targets HBM traffic in realistic serving settings. Its second component, \textbf{Rate--Distortion Retention (RDR)}, jointly chooses keep/drop decisions and precision tiers per token and head under a fixed budget, producing tier-homogeneous pages with lightweight metadata and coalesced reads. Together, ADA and RDR provide a deployment-oriented mechanism for reducing KV residency while preserving decode efficiency.

We evaluate Spherical KV on \textbf{Llama-3.1-8B-Instruct}, \textbf{Qwen2.5-14B-Instruct}, and \textbf{gpt-oss-20b} across three representative long-context regimes: language modeling on \textbf{PG-19}, retrieval-heavy question answering from \textbf{LongBench} using \textbf{HotpotQA} and \textbf{2WikiMultiHopQA}, and agentic tool rollouts on \textbf{AgentBench} and \textbf{ToolBench}. Across \textbf{8K}, \textbf{32K}, and \textbf{128K} contexts under a common paged/ragged serving substrate, Spherical KV improves decode throughput by \textbf{1.55$\times$--1.72$\times$} while reducing \textbf{resident KV bytes/token by 24--42\%} at matched quality, with the largest gains in the \textbf{128K} regime. \href{https://anonymous.4open.science/r/anonymous-submission-E510/README.md}{\textbf{Code \& evaluation artifacts}}.
\end{abstractbox}

\section{KV Cache and the Memory Bottleneck}
\label{sec:kv_cache_memory_bottleneck}

\paragraph{\textbf{KV growth is linear; \emph{HBM traffic} is the killer.}}
Autoregressive decoding caches per-layer \textbf{Keys} and \textbf{Values} to avoid recomputing attention over the full prefix. At long context, this becomes the dominant systems bottleneck: the cache grows \textbf{linearly} with context length \(T\) and multiplicatively with batch size \(B\), layers \(L\), and heads \(H\). A useful footprint estimate is
\[
\textbf{Mem}_{\text{KV}}
\approx
\underbrace{B}_{\textbf{batch}}
\cdot
\underbrace{L}_{\textbf{layers}}
\cdot
\underbrace{T}_{\textbf{context}}
\cdot
\underbrace{H}_{\textbf{heads}}
\cdot
\underbrace{\bigl(d_h^{(K)}+d_h^{(V)}\bigr)}_{\textbf{per-head dims}}
\cdot \text{bytes}.
\]
During decoding, however, the dominant cost is often not FLOPs but \textbf{HBM bytes streamed per generated token}: each new token attends over a long prefix, forcing repeated reads of historical KV pages from device memory. Past a few thousand tokens, decoding typically becomes \textbf{memory-limited} in three ways: \textbf{(i) capacity-limited}, \textbf{(ii) bandwidth/latency-limited}, and \textbf{(iii) orchestration-limited}, where paging and scheduling overheads dominate. 

\paragraph{\textbf{Deployment substrate: paging + kernel reality.}}
Long-context acceleration is therefore not primarily a FLOP-reduction problem; it is the problem of improving the \textbf{memory--throughput--quality frontier}: reduce \textbf{KV bytes}, reduce \textbf{bytes moved}, and preserve \textbf{behavioral stability} during long-horizon generation. In production, KV is rarely contiguous. Serving stacks use \textbf{paged / block-allocated KV} to reduce fragmentation, support batching, and enable preemption, making \textbf{PagedAttention}-style layouts the practical baseline abstraction \citep{kwon2023pagedattention}. Exact-attention kernels further show that end-to-end performance is governed by \textbf{HBM reads/writes} and \textbf{kernel fusion}, not only tensor arithmetic \citep{dao2022flashattention,dao2023flashattention2}. Modern serving libraries and runtimes treat paged/ragged KV and multiple cache formats as first-class objects \citep{ye2025flashinfer,nvidia_trtllm_kv_cache_2025,hf_kv_cache,vllm2026_kv_offloading}; consequently, methods that require \textbf{dense materialization}, \textbf{irregular gathers}, or heavy non-fusible transforms often lose much of their theoretical gain in deployment.

\subsection{Existing Solutions: \textbf{Three Levers} for KV Efficiency}
\label{sec:existing_kv_solutions}

\paragraph{\textbf{Three levers, plus the deciding constraint.}}
KV-efficiency methods mainly act through three levers: \textbf{(I) store fewer tokens} (retention / eviction / admission), \textbf{(II) store fewer bits} (quantization / compression), and \textbf{(III) store elsewhere} (offload / reuse). In practice, realized speedup is often decided by a fourth constraint: \textbf{kernel/serving compatibility}. If a method breaks paged layouts, requires irregular gathers, or inserts non-fusible transforms, nominal KV savings may not translate into wall-clock gains. The real target is therefore not just smaller \(\textbf{Mem}_{\text{KV}}\), but a better frontier of \textbf{peak capacity}, \textbf{decode IO}, and \textbf{behavioral stability} \citep{scbench2025,openreview2025_kvcache_reasoning,assessing_kv_reasoning_2025}.

\subsubsection{\textbf{Lever I: store fewer tokens}}
\label{sec:lever_tokens_compact}

A first family reduces KV residency by deciding \emph{which} states remain at decode time. Early heuristics rely on recency or attention surrogates: \textbf{StreamingLLM}, \textbf{H$_2$O}, and \textbf{Scissorhands} \citep{xiao2023streamingllm,zhang2023h2o,liu2023scissorhands}. A second pattern compresses once after prefill and keeps the reduced cache fixed during decoding, as in \textbf{SnapKV} and \textbf{SAGE-KV} \citep{li2024snapkv,wang2025sagekv}. More recent work exploits structural heterogeneity: \textbf{PyramidKV}, \textbf{CAKE}, \textbf{ChunkKV}, \textbf{GraphKV}, and \textbf{CurDKV} allocate memory using layer structure, semantic chunks, graph propagation, or value-guided preservation \citep{cai2024pyramidkv,qin2025cake,chunkkv2025,graphkv2025,curdkv2025}. In parallel, control is shifting from eviction to write-time policies: \textbf{TRIM-KV}, \textbf{Write-Gated KV}, and \textbf{Expected Attention} learn or predict future utility more explicitly \citep{bui2025trimkv,huang2025writegatedkv,devoto2025expectedattention}. The broader lesson is that token utility is \textbf{non-local}, \textbf{task-dependent}, and \textbf{head-/layer-dependent}, pushing retention toward explicit controllers.

\subsubsection{\textbf{Lever II: store fewer bits}}
\label{sec:lever_bits_compact}

A second family reduces the \emph{cost per retained state}. Representative examples include \textbf{KIVI}, \textbf{KVQuant}, \textbf{ZipCache}, and \textbf{GEAR} \citep{liu2024kivi,hooper2024kvquant,he2024zipcache,kang2024gear}. At aggressive compression levels, \textbf{outliers} become central: \textbf{OTT}, \textbf{RotateKV}, and \textbf{Kitty} explicitly protect them while using mixed or dynamic precision \citep{su2025ott,su2025rotatekv,xia2025kitty}. A parallel line explores \textbf{geometry-aware encodings}: polar/spherical parameterizations separate \textbf{direction} from \textbf{magnitude}, improving rate allocation and enabling compressed-domain similarity primitives \citep{han2025polarquant,wu2025polarquant,xiao2026spherical_similarity}. The main systems lesson is that low-bit KV is useful only with \textbf{outlier protection}, \textbf{adaptive precision}, and \textbf{kernel co-design}; otherwise decode still pays a hidden \emph{dequantize / reconstruct / dot-product} tax \citep{du2025bitdecoding}.

\subsubsection{\textbf{Lever III: store elsewhere}}
\label{sec:lever_elsewhere_compact}

A third family changes \emph{where} KV resides. \textbf{Offloading/tiering} systems move KV across GPU, CPU, or storage to support longer contexts and higher concurrency, as in the \textbf{vLLM KV offloading connector} and \textbf{FlexiCache} \citep{vllm2026_kv_offloading,takbir2025flexicache}. \textbf{Reuse-oriented} systems externalize KV across requests or shared prefixes, as in \textbf{LMCache} \citep{cheng2025lmcache}. These methods expand effective capacity, but their gains depend strongly on transfer bandwidth, page layout, prefetch quality, and scheduler behavior. The main systems lesson is that offloading can relieve residency pressure, but often at the cost of an \textbf{orchestration tax}. Extended method-level comparisons, runtime details, and a fuller related-work taxonomy are deferred to the Appendix ~\ref{app:kv-bottleneck}.

\begin{figure}[ht!]
\vspace{-1em}
  \centering
  \includegraphics[width=\linewidth]{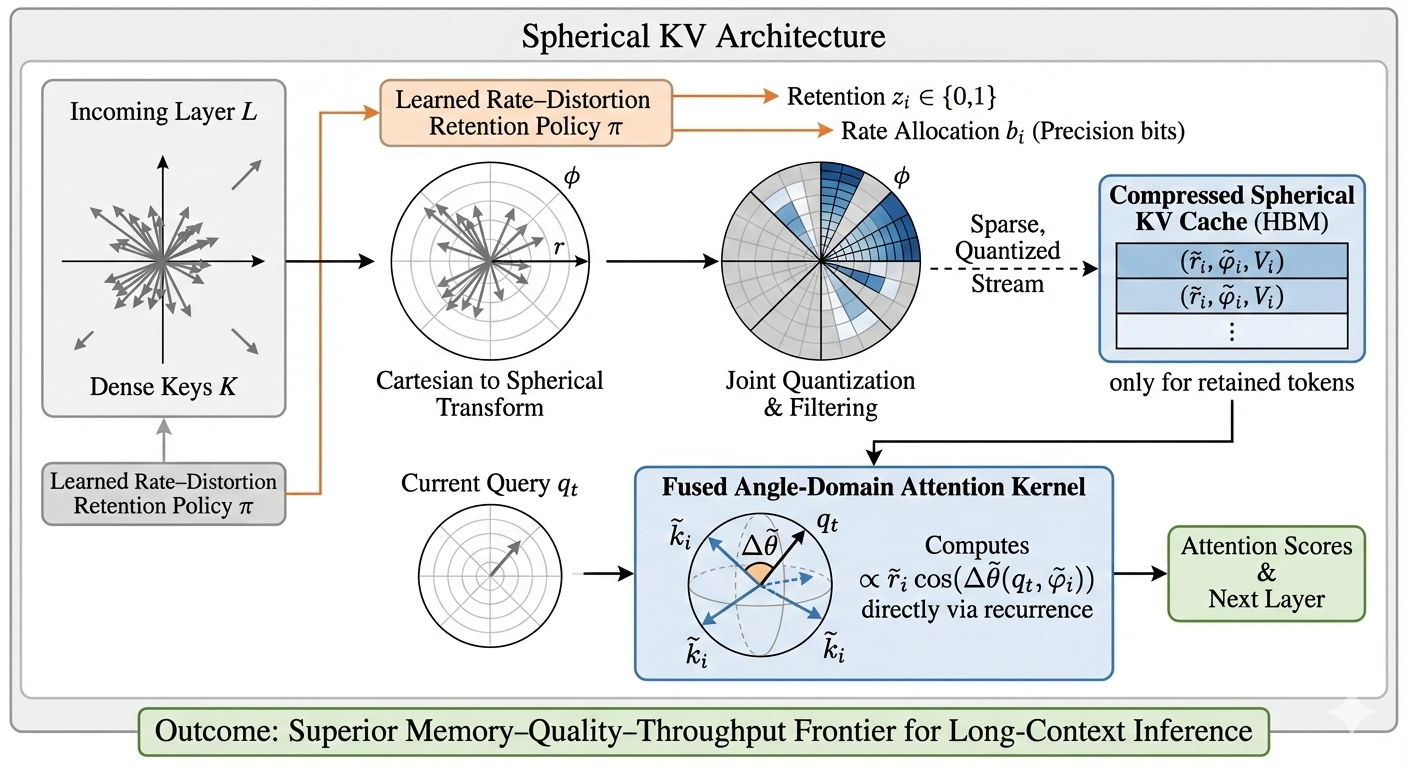}
  \caption{\textbf{Spherical KV.}
Dense keys from the incoming layer are mapped to a \textbf{\emph{spherical parameterization}} with radius $r$ and angles $\phi$. A learned \textbf{\emph{rate--distortion retention policy}} $\pi$ enforces a strict memory budget $B$ by predicting \textbf{\emph{(i) a keep/drop decision}} $z_i \in \{0,1\}$ and \textbf{\emph{(ii) a per-item precision tier}} $b_i$. Joint quantization and filtering produce a \textbf{\emph{sparse, quantized spherical stream}} that is stored in a compressed KV cache in \textbf{\emph{HBM}} for retained tokens only. During decoding, a fused \textbf{\emph{angle-domain attention kernel}} consumes the quantized $(\tilde r_i,\tilde\phi_i)$ directly, thereby \textbf{\emph{avoiding dense Cartesian reconstruction in the critical path}}, and computes attention scores for the next layer. The overall pipeline is designed to improve the \textbf{\emph{memory--quality--throughput trade-off}} for long-context inference.}
  \label{fig:spherical_kv_architecture}
  \vspace{-1em}
\end{figure}

\section{Spherical KV: \texorpdfstring{\textbf{Angle-Domain Attention}}{Angle-Domain Attention} and \texorpdfstring{\textbf{Rate--Distortion Retention}}{Rate--Distortion Retention}}
\vspace{-2pt}
Existing KV methods often optimize a single axis at a time, whereas \textbf{\emph{Spherical KV}} jointly addresses \textbf{representation}, \textbf{decode-time compute}, and \textbf{retention budgeting}. \textbf{Not ``just another quantizer'': two hard commitments.} \textbf{First}, replace ``\emph{decompress then dot-product}'' with \textbf{angle-domain compute} fused into IO-aware decoding kernels, so that reductions in stored KV bytes translate into lower \textbf{HBM traffic per token} under realistic paged serving \citep{dao2022flashattention,dao2023flashattention2,ye2025flashinfer}. \textbf{Second}, unify \textbf{keep/drop} with \textbf{precision allocation} by treating KV control as a \textbf{rate--distortion} problem over token/head/layer states, rather than as separate eviction and quantization stages \citep{bui2025trimkv,huang2025writegatedkv,xia2025kitty,su2025ott}. Concretely, Spherical KV consists of \textbf{two orthogonal components} together with a \textbf{serving-oriented KV layout}. \textbf{(i) Angle-Domain Attention (ADA)} is the \emph{compute primitive}: it evaluates attention logits directly from \emph{compact spherical key codes}, avoiding dense-key reconstruction in HBM. \textbf{(ii) Rate--Distortion Retention (RDR)} is the \emph{control primitive}: it allocates both \emph{residency} and \emph{precision} under a strict KV budget. Together, ADA and RDR yield \textbf{paged, tier-homogeneous KV pages} compatible with ragged batching and streaming decode kernels, consistent with modern fused-attention and paged-serving substrates used in practical LLM inference \citep{dao2022flashattention,dao2023flashattention2,kwon2023pagedattention}. Figure~\ref{fig:spherical_kv_architecture} summarizes the full pipeline: dense keys are mapped into a \textbf{\emph{spherical representation}}, \textbf{\emph{rate--distortion control}} governs \textbf{\emph{retention}} and \textbf{\emph{precision}}, and decoding proceeds through a fused \textbf{\emph{angle-domain kernel}} \textbf{\emph{without dense reconstruction}}. Extended details, and derivations are deferred to Appendix~\ref{app:spherical-kv}.

\begin{figure}[ht!]
\vspace{-0.5em}
  \centering
  \includegraphics[width=\linewidth]{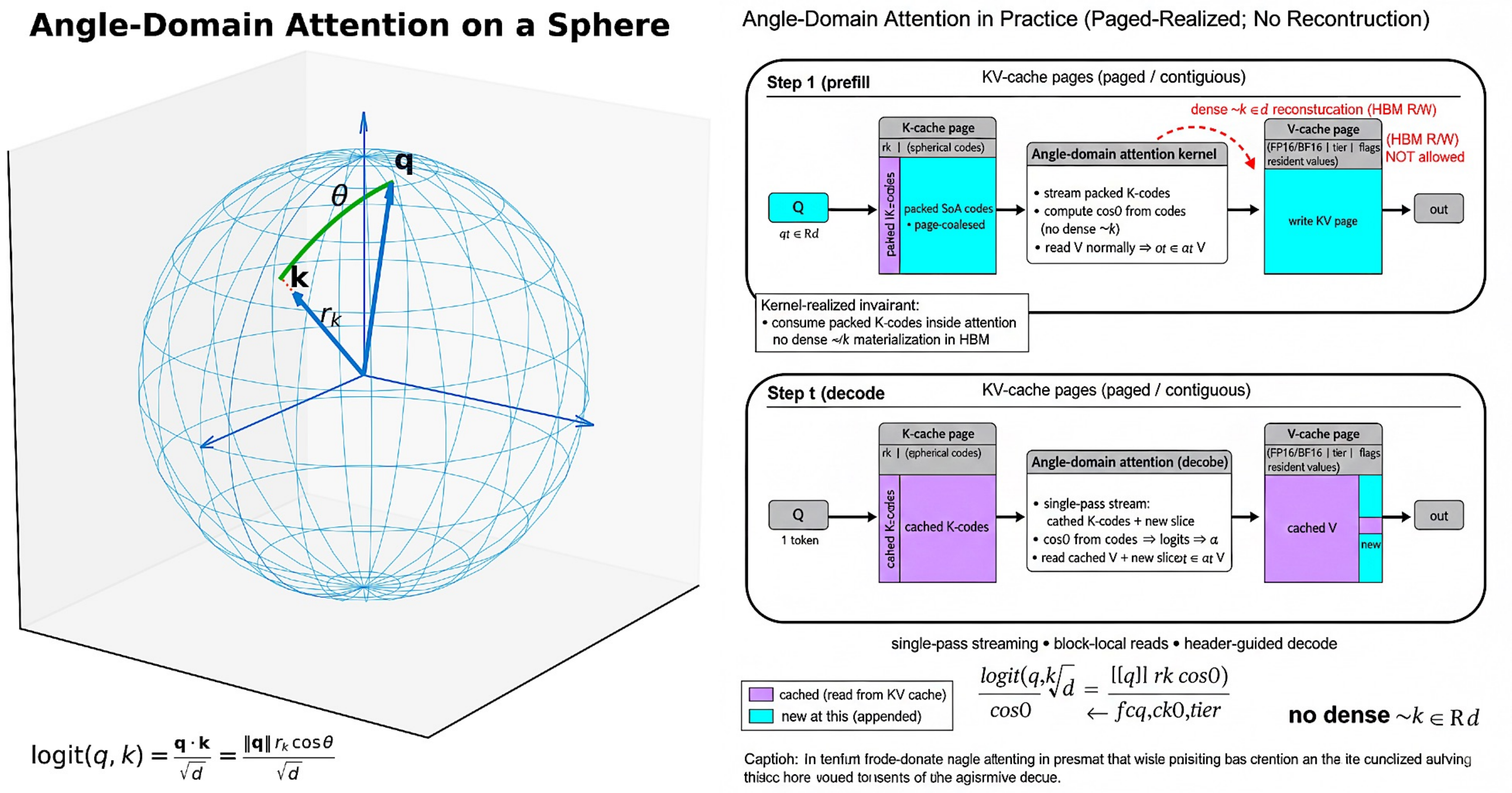}
  \caption{\textbf{Angle-Domain Attention in Practice (paged KV; kernel-realized, \emph{no reconstruction}).}
\textbf{ADA} integrates into a paged KV decode loop without densification.
\textbf{(i) Prefill:} per token, we write a \textbf{paged K-cache page} storing \textbf{compact spherical key codes}---radius \(r_k\), packed angular codes \(c_k^\theta\), and \textbf{tier/flags} metadata in a \textbf{header-guided}, kernel layout. In parallel, we write a \textbf{V-cache page} in FP16/BF16, preserving paged KV, pointer tables, and batching.
\textbf{(ii) Decode:} at step \(t\), cached pages are reused and a thin slice is appended for the generated token. Inside the attention kernel, instead of reconstructing a \(\tilde{\mathbf{k}}\in\mathbb{R}^d\), the kernel \textbf{consumes packed K-codes directly} and computes similarity via \(\cos\theta\), so that \( \mathbf{q}^{\top}\mathbf{k}\propto \|\mathbf{q}\|\,r_k\,\cos\theta\). The resulting logits produce \(\alpha_t=\mathrm{softmax}(\cdot)\), after which values are read to form \(o_t=\alpha_tV\); thus, \textbf{K is consumed in compressed domain, while \(V\) remains standard-domain read}. The central invariant is \textbf{no dense-key reconstruction} in the decode hot loop, preserving a \textbf{single-pass, block-local stream} over packed K-codes and coalesced reads.}
  \label{fig:angle_domain_attention_practice}
  \vspace{-1em}
\end{figure}

\subsection{Angle-Domain Attention}
\label{sec:angle_domain_attention}
\vspace{-2pt}

\paragraph{\textbf{Directional factorization of attention.}}
For a head of dimension \(d\), standard attention uses
\[
\ell(\mathbf q,\mathbf k)=\frac{\mathbf q^\top \mathbf k}{\sqrt d}.
\]
Write \(\mathbf q=\|\mathbf q\|\hat{\mathbf q}\) and \(\mathbf k=\|\mathbf k\|\hat{\mathbf k}\), where \(\hat{\mathbf q},\hat{\mathbf k}\in \mathbb S^{d-1}\). Then
\[
\ell(\mathbf q,\mathbf k)
=
\frac{\|\mathbf q\|\,\|\mathbf k\|}{\sqrt d}\,
\hat{\mathbf q}^\top \hat{\mathbf k}
=
\frac{\|\mathbf q\|\,\|\mathbf k\|}{\sqrt d}\cos\theta,
\qquad
\cos\theta \doteq \hat{\mathbf q}^\top \hat{\mathbf k}.
\]
This decomposition isolates \textbf{direction} and \textbf{magnitude}: the former determines angular alignment, while the latter acts as a scalar gain. ADA exploits this factorization by representing the directional term in compressed form and preserving only the radial correction needed for stable logits.

\paragraph{\textbf{Spherical coding and compressed-domain similarity.}}
Each key is mapped to a spherical tuple
\[
\mathbf k \;\mapsto\; \bigl(r_k,\phi_k\bigr),\qquad
r_k=\|\mathbf k\|,\ \ \phi_k=\Phi(\hat{\mathbf k}),
\]
where \(\Phi:\mathbb S^{d-1}\rightarrow \mathbb R^{d-1}\) denotes a spherical coordinate map. Rather than caching \((r_k,\phi_k)\) in full precision, ADA stores
\[
\mathcal C(\mathbf k)=\bigl(\tilde r_k,\;c_k^\theta;\;b,\mathrm{flags}\bigr),
\]
where \(\tilde r_k\) is a quantized radius, \(c_k^\theta\) is a packed angular code at tier \(b\in\mathcal B\), and \(\mathrm{flags}\) encode lightweight protection or segment metadata. At decode step \(t\), the current query is encoded on the fly as \((r_q,c_q^\theta)\). The kernel then evaluates a compressed-domain angular surrogate
\[
\widehat{\cos\theta}
\;\doteq\;
f_b(c_q^\theta,c_k^\theta),
\]
where \(f_b\) is implemented directly on packed code streams. Thus ADA replaces dense directional vectors \(\hat{\mathbf k}\in\mathbb R^d\) with a code-domain operator over \((d-1)\)-dimensional angular information.

\begin{figure}[ht!]
\vspace{-1em}
  \centering
  \includegraphics[width=\linewidth]{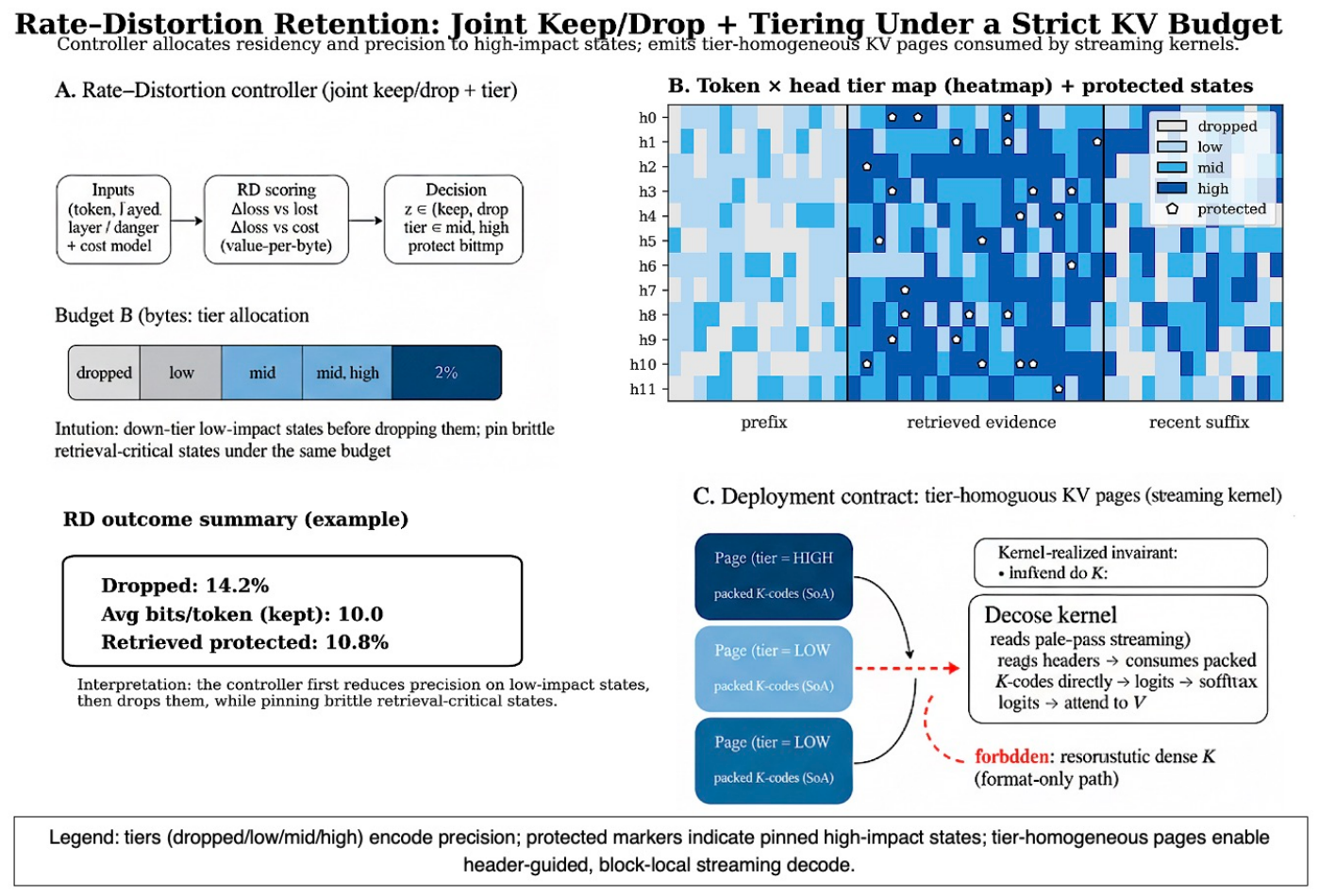}
  \vspace{-1.5em}
\caption{\textbf{Rate--Distortion Retention: joint keep/drop + tiering under a strict KV budget.}
KV residency is treated as a constrained \textbf{\emph{rate--distortion allocation}} problem: under a fixed budget, the controller spends precision on \textbf{\emph{high-impact states}} and removes \textbf{\emph{low-utility states}} first.
\textbf{(A) RD controller.} Given token/head/layer context and a cost model, the controller assigns each state to \emph{drop} or \emph{keep} with a precision tier (low/mid/high) and protection; the budget bar highlights \textbf{\emph{down-tier before drop}}.
\textbf{(B) Token $\times$ head tier map.} Rows are heads, columns are tokens, and color indicates the assigned tier; separators mark prompt segments and markers denote \textbf{\emph{protected states}}. The pattern reveals \textbf{\emph{segment-aware allocation}}, \textbf{\emph{head heterogeneity}}, and \textbf{\emph{precision reduction before dropping}}.
\textbf{(C) Deployment contract.} Retained states are written into \textbf{\emph{tier-homogeneous KV pages}} so the decode kernel can perform \textbf{\emph{single-pass, header-guided reads}}: packed $K$-codes are consumed, while $V$ is read normally. The red dashed path marks the forbidden \emph{format-only} alternative---reconstructing dense $K$---which would reintroduce HBM traffic and erase the kernel-realized gain.}
  \label{fig:rd_retention_neurips}
  \vspace{-1em}
\end{figure}

\paragraph{\textbf{Kernel-realized angle-domain logits.}}
The defining invariant is that the decode kernel never materializes a dense \(\tilde{\mathbf k}\in\mathbb R^d\) in HBM. Instead, it computes
\[
\hat{\ell}(\mathbf q,\mathbf k)
\;\doteq\;
\frac{r_q\,\tilde r_k}{\sqrt d}\,f_b(c_q^\theta,c_k^\theta),
\qquad r_q=\|\mathbf q\|,
\]
using only the cached radius/code pair and the current query code. Softmax and value mixing remain standard, \(\alpha_i=\exp(\hat{\ell}_i)/\sum_j \exp(\hat{\ell}_j)\) and \(\mathbf{o}=\sum_i \alpha_i \mathbf{v}_i\), with values \(\mathbf{v}_i\) stored and read in a serving-native format. The asymmetry is intentional: \textbf{keys are consumed in compressed domain}, while \textbf{values remain in standard domain}, preserving the critical-path reduction where long-context decoding is most bandwidth-sensitive.

\paragraph{\textbf{Distortion enters through angular approximation.}}
Let
\[
f_b(c_q^\theta,c_k^\theta)=\cos\theta+\varepsilon_\theta,
\qquad
\tilde r_k=\|\mathbf k\|+\varepsilon_r.
\]
Then the induced logit error is
\[
\Delta \ell
\;\doteq\;
\hat{\ell}-\ell
=
\frac{r_q}{\sqrt d}
\Big[
(\|\mathbf k\|+\varepsilon_r)(\cos\theta+\varepsilon_\theta)
-\|\mathbf k\|\cos\theta
\Big].
\]

This makes the design target explicit: the dominant error term is typically angular, and finer tiers reduce it directly. ADA therefore exposes a concrete rate--distortion knob through the tier \(b\), which RDR later allocates under budget.

\paragraph{\textbf{Why ADA is not standard KV quantization.}}
Unlike KV quantization schemes that still reconstruct dense keys before the dot-product, ADA computes similarity \textbf{in compressed domain}, avoiding reconstruction-induced HBM traffic. ADA is therefore a \textbf{kernel-level attention primitive}, aligned with fused attention kernels \citep{dao2022flashattention,dao2023flashattention2} and paged serving stacks \citep{kwon2023pagedattention}; Fig.~\ref{fig:angle_domain_attention_practice} shows this \textbf{no-reconstruction} path.

\subsection{Rate--Distortion Retention}
\label{sec:rate_distortion_retention}
\vspace{-2pt}

\paragraph{\textbf{Joint control of residency and precision.}}
Under a strict memory budget, selecting \textbf{\emph{which states to keep}} is insufficient; one must also choose \textbf{\emph{at what precision}} they are stored. We index cache states by \(i\) (token \(\times\) head, optionally layer), choose a binary residency variable \(z_i\in\{0,1\}\) and a tier assignment \(b_i\in\mathcal B\), let \(\mathrm{cost}(b)\) denote bytes per state under tier \(b\) (including amortized metadata), and let \(\mathcal D_i(b)\) denote the expected distortion from coding state \(i\) at tier \(b\). The controller solves
\[
\min_{\{z_i,b_i\}}
\ \mathbb{E}\!\left[\mathcal L\!\left(\mathrm{Decode}_{\text{SphKV}}(\{z_i,b_i\})\right)\right]
\qquad
\text{s.t.}\qquad
\sum_i z_i\,\mathrm{cost}(b_i)\le B,
\]
a discrete \textbf{\emph{rate--distortion allocation}} problem \citep{shannon1948} in which the budget \(B\) must be distributed jointly across \textbf{\emph{which}} states are retained and \textbf{\emph{how accurately}} each retained state is represented.

\paragraph{\textbf{Lagrangian form and state-wise value.}}
Introducing a multiplier \(\lambda\ge 0\) yields the relaxed objective
\[
\min_{\{z_i,b_i\}}
\ \mathbb{E}\!\left[\mathcal L(\cdot)\right]
+\lambda\!\left(\sum_i z_i\,\mathrm{cost}(b_i)-B\right).
\]
A useful surrogate is obtained by comparing each action against a baseline \(a_0\) (e.g., \emph{drop} or \emph{lowest tier}). Define the incremental utility of assigning state \(i\) to tier \(b\) as
\[
\Delta_i(b)
\ \doteq\
\mathbb{E}\!\left[\mathcal L_i(a_0)\right]
-
\mathbb{E}\!\left[\mathcal L_i(b)\right],
\qquad
\mathcal U_i(b;\lambda)\doteq \Delta_i(b)-\lambda\,\mathrm{cost}(b),
\]
and the corresponding \textbf{value per byte}
\[
\mathrm{vpB}_i(b)
\ \doteq\
\frac{\Delta_i(b)}
{\mathrm{cost}(b)-\mathrm{cost}(a_0)}.
\]
In \textbf{\emph{discrete-tier form}}, this yields a \textbf{\emph{multiple-choice knapsack}} over \((i,b)\) pairs. Exact optimization is generally intractable, but maximizing \(\mathcal U_i(b;\lambda)\) or greedily allocating by \(\mathrm{vpB}\) gives a strong controller and typically traces a near-Pareto frontier in serving settings \citep{martello1990knapsack}.

\paragraph{\textbf{Mechanistic distortion surrogate from ADA.}}
The distortion \(\mathcal D_i(b)\) is not abstract: it arises from the angular and radial approximation errors induced by the coded key representation. If \(\varepsilon_{\theta,i}(b)\) and \(\varepsilon_{r,i}(b)\) denote the angular and radius errors for state \(i\) at tier \(b\), the ADA analysis gives
\[
|\Delta \ell_i(b)|
\ \lesssim\
\frac{r_q\,\|\mathbf k_i\|}{\sqrt d}\,|\varepsilon_{\theta,i}(b)|
\;+\;
\frac{r_q}{\sqrt d}\,|\varepsilon_{r,i}(b)|.
\]
This suggests a state-wise distortion surrogate of the form
\[
\mathcal D_i(b)
\ \propto\
w_i^\theta\,|\varepsilon_{\theta,i}(b)|
+
w_i^r\,|\varepsilon_{r,i}(b)|,
\qquad
w_i^\theta \asymp \frac{r_q\|\mathbf k_i\|}{\sqrt d},
\]
where the weights capture error amplification through the logit map. Thus, tier selection can be interpreted as allocating bits to states whose approximation error would most strongly perturb attention. RDR is therefore \textbf{mechanistically coupled} to ADA: it spends bytes where angular/radial distortion is amplified and saves bytes where the model is comparatively insensitive.

\paragraph{\textbf{Protection constraints and deployable control.}}
Some states are too brittle for opportunistic compression---for example \textbf{\emph{retrieved evidence}}, \textbf{\emph{safety-critical spans}}, \textbf{\emph{outliers}}, or \textbf{\emph{low-margin decisions}}. We therefore introduce a protection indicator \(\mathbb I_{\mathrm{prot}}(i)\in\{0,1\}\) and enforce \(\mathbb I_{\mathrm{prot}}(i)=1 \Rightarrow z_i=1,\; b_i=b_{\max}\). The controller then follows a simple deployment rule: \textbf{\emph{score}} states using \(\mathcal U_i(b;\lambda)\) or \(\mathrm{vpB}_i(b)\), \textbf{\emph{down-tier before drop}}, and \textbf{\emph{write}} retained states into fixed-header, tier-homogeneous pages. This policy avoids \textbf{\emph{hard truncation}}, yields \textbf{\emph{smoother degradation}} as budgets tighten, and preserves layouts compatible with \textbf{\emph{specialized kernels}} and \textbf{\emph{coalesced reads}}.

\subsection{Working Together: One Contract, Two Mechanisms}
\label{sec:both_together}
\vspace{-2pt}

\textbf{Angle-Domain Attention (ADA)} and \textbf{Rate--Distortion Retention (RDR)} operate on complementary axes: ADA determines \emph{how} logits are computed from codes, while RDR determines \emph{which} states are retained and \emph{at what precision tier}. The deployment contract is a \textbf{paged, tier-homogeneous KV layout} with pointer tables for ragged batching \citep{kwon2023pagedattention}, compact headers, SoA-packed key payloads \((r_k,c_k^\theta)\), and standard FP16/BF16 values; together, these preserve compression in the actual decode path.In KV-dominated regimes, this joint design shifts the \textbf{memory--quality--throughput frontier}: \textbf{RDR} lowers effective KV bytes/token, \textbf{ADA} reduces hot-path bandwidth, and the shared layout realizes these savings under practical paged serving in end-to-end decoding while keeping the distortion knob explicit through tier \(b\) \citep{dao2022flashattention,dao2023flashattention2,kwon2023pagedattention,shannon1948}.

\section{Experiments and Results}
\label{sec:experiments_results}
\vspace{-2pt}

We evaluate \textbf{\emph{Spherical KV}} in the regime where long-context inference is \textbf{\emph{memory-}} and \textbf{\emph{bandwidth-bound}}: long prompts, ragged retrieval blocks, and long-horizon decode rollouts. Our central claim is a \textbf{\emph{Pareto improvement}} in deployment-relevant operating points: at matched task quality, Spherical KV reduces \textbf{\emph{resident KV bytes/token}} and \textbf{\emph{HBM bytes/generated-token}} while increasing \textbf{\emph{steady-state decode throughput}}. All results are measured under a common \textbf{\emph{paged/ragged serving substrate}}, so gains must survive realistic cache layouts, metadata overheads, and kernel constraints rather than contiguous-cache toy settings \citep{kwon2023efficient}. cf. Appendix ~\ref{app:exp-protocol} and ~\ref{app:extended_results}.

\begin{figure}[ht!]
\vspace{-1em}
  \centering
  \includegraphics[width=\textwidth]{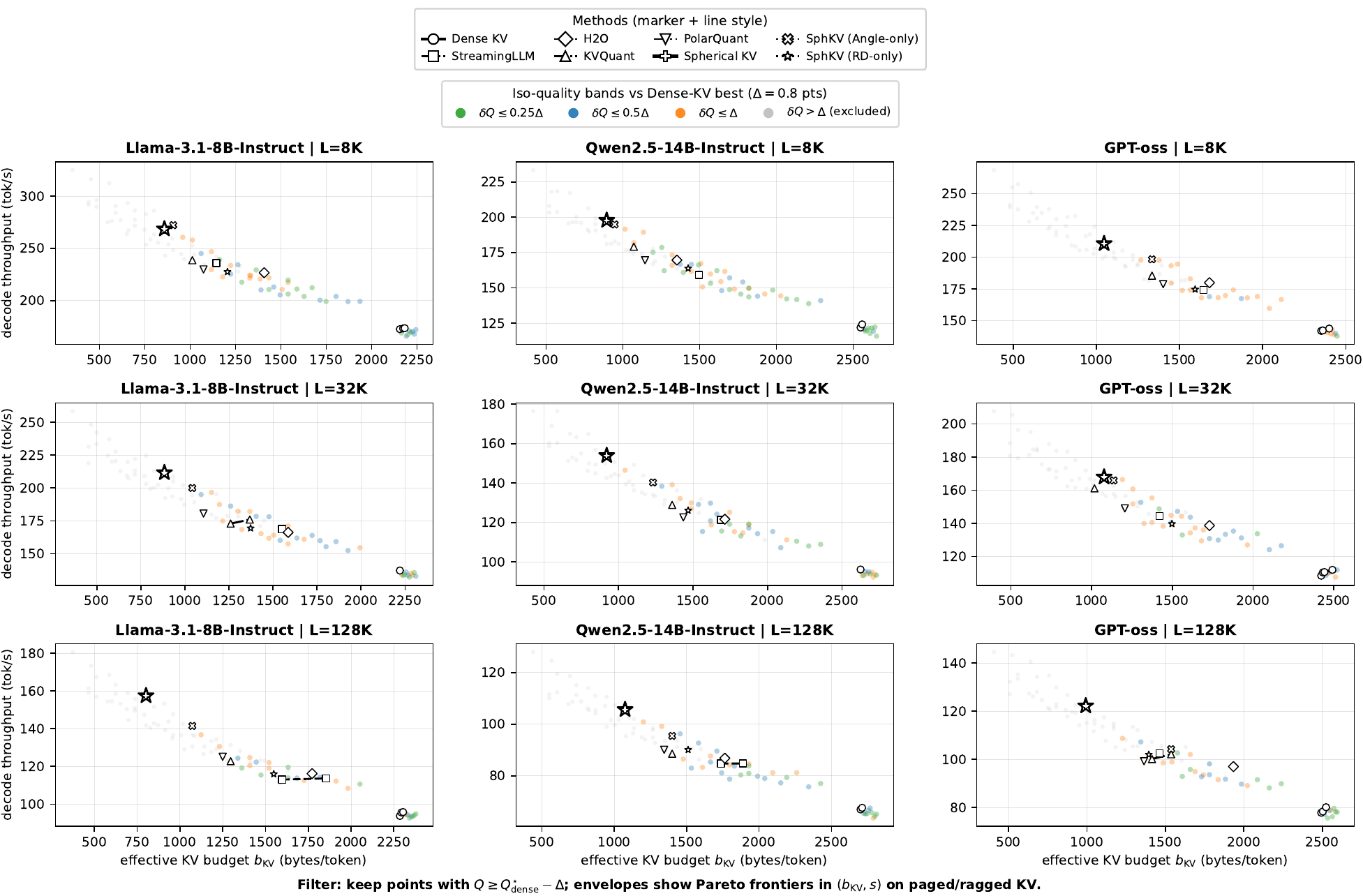}
  \caption{\textbf{Iso-quality Pareto frontiers for \emph{memory-bounded} decoding.}
Each panel plots \textbf{decode throughput} (tok/s; \textbf{higher is better}) versus \textbf{effective KV budget} $b_{\mathrm{KV}}$ (bytes/token; \textbf{lower is better}) under \textbf{paged/ragged serving}, across three models and context lengths $L\in\{8\mathrm{K},32\mathrm{K},128\mathrm{K}\}$.
Let $Q$ be the \textbf{quality score} (\textbf{higher is better}; defined in \S\,[X]) and let $Q^{\star}_{\mathrm{dense}}$ denote the \textbf{best Dense-KV quality} in the panel.
We enforce an \textbf{iso-quality constraint} by retaining only configurations with \textbf{$Q \geq Q^{\star}_{\mathrm{dense}}-\Delta$} (\textbf{$\Delta=0.8$} points).
Background dots are colored by the \textbf{quality gap} $\delta Q = Q^{\star}_{\mathrm{dense}}-Q$ (\textbf{gray}: excluded, $\delta Q>\Delta$).
For each method, the black polyline traces the \textbf{Pareto envelope} over retained points in $(b_{\mathrm{KV}},\text{tok/s})$.
The \textbf{$\star$} marks the \textbf{best throughput-per-byte} operating point for \textbf{Spherical KV} among retained configurations, highlighting an \textbf{up-left} shift of the \textbf{iso-quality frontier}.}
  \label{fig:iso_quality_frontier}
  \vspace{-1.5em}
\end{figure}

\subsection{Experimental Setup and Evaluation Protocol}
\label{sec:exp_setup_protocol}
\vspace{-2pt}

We evaluate three open-weight instruct-tuned LLMs---\textbf{\emph{Llama-3.1-8B-Instruct}}, \textbf{\emph{Qwen2.5-14B-Instruct}}, and \textbf{\emph{gpt-oss-20b}}---across context lengths \(L\in\{8\text{K},32\text{K},128\text{K}\}\). Workloads cover \textbf{\emph{(i)}} long-context language modeling on \textbf{\emph{PG-19}} \citep{rae2020compressive,dai2019transformerxl}, \textbf{\emph{(ii)}} retrieval-heavy QA from \textbf{\emph{LongBench}} using \textbf{\emph{HotpotQA}} and \textbf{\emph{2WikiMultiHopQA}} \citep{bai2024longbench,ho2020constructing}, and \textbf{\emph{(iii)}} agentic/tool rollouts on \textbf{\emph{AgentBench}} and \textbf{\emph{ToolBench}} \citep{liu2023agentbench,qin2023toolbench}. All methods run on the \textbf{\emph{same paged/ragged KV substrate}} with identical prompt templates, tokenization, and decoding settings \citep{kwon2023pagedattention}.

Our evaluation is deployment-true. We define the effective resident KV budget as
\[
b_{\mathrm{KV}}
\;\doteq\;
\frac{M_{\mathrm{KV}}^{\mathrm{resident}}}{T_{\mathrm{active}}},
\qquad
b_{\mathrm{HBM}}
\;\doteq\;
\frac{\mathrm{HBM}_{\mathrm{read}}+\mathrm{HBM}_{\mathrm{write}}}{\#\text{ decode tokens}},
\]
where \(M_{\mathrm{KV}}^{\mathrm{resident}}\) includes packed codes, page headers, pointer tables, protect bitmaps, fragmentation, and allocator overhead \citep{kwon2023efficient}. We report four axes: \textbf{\emph{memory}} (peak KV footprint, \(b_{\mathrm{KV}}\)), \textbf{\emph{throughput}} (steady-state tok/s), \textbf{\emph{traffic}} (\(b_{\mathrm{HBM}}\)), and \textbf{\emph{quality/stability}} (NLL/PPL, EM/F1, success, seed sensitivity, and trajectory disagreement). Because long-context decoding is governed by \textbf{\emph{memory traffic}} rather than arithmetic alone, \(b_{\mathrm{HBM}}\) is a primary systems KPI \citep{dao2022flashattention,dao2023flashattention2,kwon2023efficient}.

To ensure \textbf{\emph{deployment-faithful}} comparisons, we enforce an \textbf{\emph{iso-quality contract}}: if \(Q_{\mathrm{dense}}^\star\) is the best dense-KV quality under the same decoding settings, an operating point \(p\) is retained only when \(Q(p)\ge Q_{\mathrm{dense}}^\star-\Delta\). Any baseline that requires dense reconstruction or decode-time staging must pay that cost end-to-end \citep{lin2024kvquant,kwon2023efficient}; full runtime details, profiler counters, hardware configuration, and measurement windows are deferred to the appendix.

\subsection{Iso-Quality Frontier Results}
\label{sec:isoq_frontier_results}
\vspace{-2pt}

Our main result is the \textbf{\emph{iso-quality Pareto frontier}} in Fig.~\ref{fig:iso_quality_frontier}. For each \textbf{\emph{method}}, \textbf{\emph{model}}, and \textbf{\emph{context length}}, we sweep operating points and retain only configurations satisfying the iso-quality constraint above. The frontier is then the Pareto envelope
\[
\mathrm{Env}^{\Delta}
\;\doteq\;
\mathrm{Pareto}\!\left(\{(b_{\mathrm{KV}}(p),\,s(p)):\ p\in \mathcal P^{\Delta}\}\right),
\]
where \(s(p)\) is decode throughput and \(\mathcal P^{\Delta}\) is the quality-matched feasible set. Thus, each panel is a \textbf{\emph{frontier-at-matched-quality}} comparison, not a generic speed--quality trade \citep{lin2024kvquant,zhang2023h2o,xiao2023streamingllm}.

Across models and context lengths, \textbf{\emph{Spherical KV}} yields a consistent \textbf{\emph{up-left shift}} of the retained envelope relative to Dense KV and strong baselines: it achieves \textbf{\emph{higher throughput at lower effective KV residency}} under the same quality constraint. The effect strengthens toward the \textbf{\emph{128K stress regime}}, where KV residency dominates cost and paging/fragmentation effects are most severe. In this regime, methods that reduce nominal bytes but still rely on reconstruction often fall off the iso-quality set or fail to deliver comparable throughput gains \citep{liu2023lost,kwon2023efficient}.

We summarize the frontier with two deployment-relevant gains:
\[
\Gamma_s
\;\doteq\;
\max_{p\in\mathcal P^{\Delta}}
\frac{s(p)}{s_{\mathrm{dense}}},
\qquad
\Gamma_m
\;\doteq\;
\min_{p:\,s(p)\ge s_{\mathrm{dense}}}
\frac{b_{\mathrm{KV}}(p)}{b_{\mathrm{KV},\mathrm{dense}}}.
\]
Here \(\Gamma_s\) is the \textbf{\emph{iso-quality speedup}}, and \(\Gamma_m\) is the \textbf{\emph{iso-throughput memory ratio}}; lower is better for \(\Gamma_m\). We also report the corresponding \(b_{\mathrm{HBM}}\) at the chosen points, since any meaningful long-context speedup must be explained by reduced memory traffic \citep{dao2022flashattention,kwon2023efficient}.

\begin{figure}[t]
  \centering
  \includegraphics[width=\textwidth]{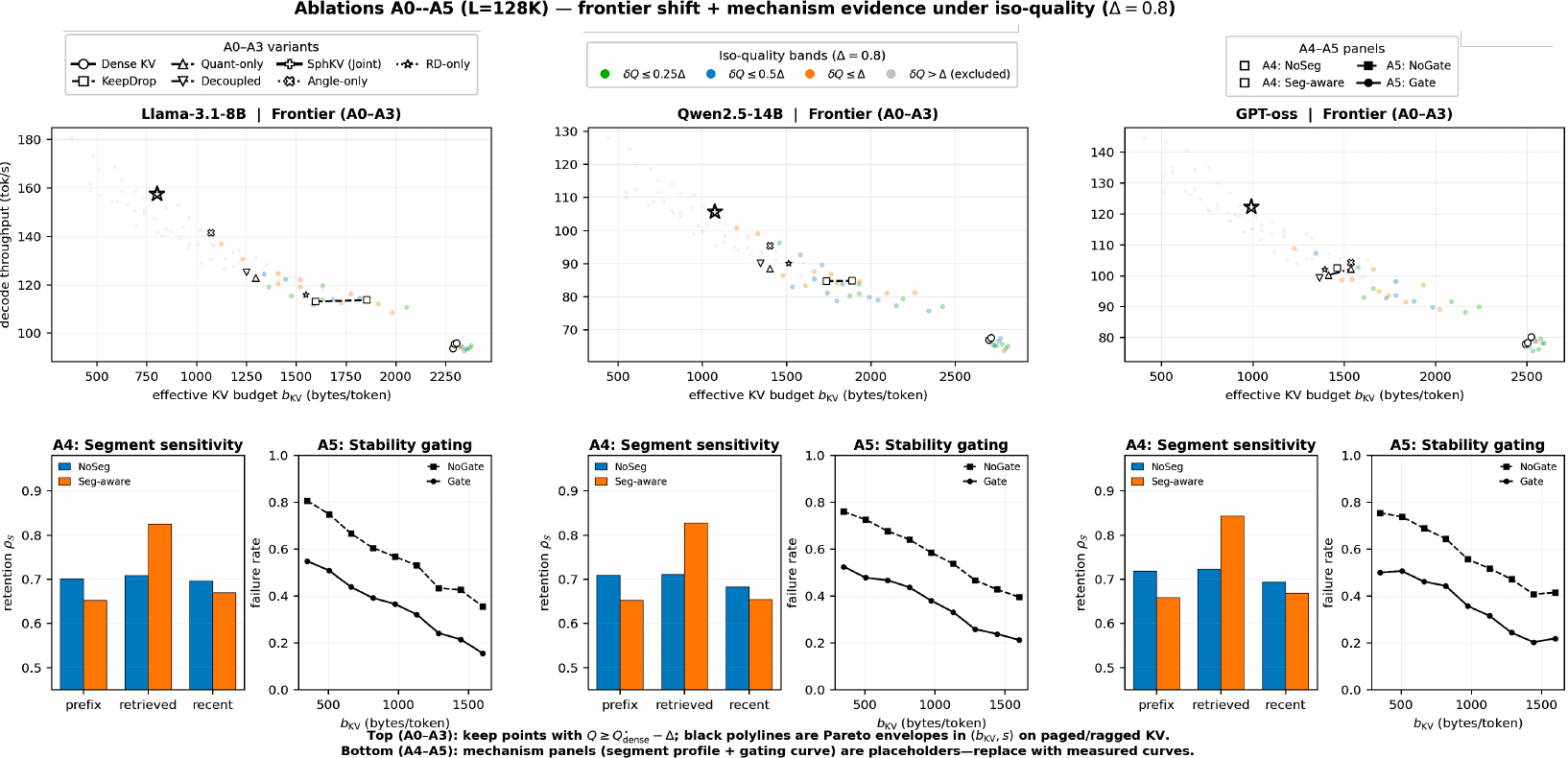}
  \caption{\textbf{Ablations A0--A5 at extreme context (L=128K): frontier shift + mechanism evidence under iso-quality ($\Delta=0.8$).}
\textbf{Top row (A0--A3, three LLMs):} For each model (columns), we plot decode throughput $s$ (tok/s; \textbf{higher is better}) versus effective resident KV budget $b_{\mathrm{KV}}$ (bytes/token; \textbf{lower is better}) on a \textbf{paged/ragged} KV substrate. Let $Q^{\star}_{\mathrm{dense}}$ be the best \textbf{Dense KV} quality for that model; we retain points satisfying $Q \ge Q^{\star}_{\mathrm{dense}}-\Delta$ and color them by the quality gap $\delta Q = Q^{\star}_{\mathrm{dense}}-Q$ (green/blue/orange: within-band; gray: excluded). For each variant, the \textbf{black polyline} traces the \textbf{Pareto envelope} in $(b_{\mathrm{KV}}, s)$, showing whether the ablation \emph{shifts} the \textbf{iso-quality frontier}. Variants instantiate the A0--A3 story: \textbf{KeepDrop} (retention-only), \textbf{Quant-only}, \textbf{Decoupled} (two-stage), and \textbf{SphKV (Joint)} plus \textbf{Angle-only} and \textbf{RD-only}; the $\star$ marks the best throughput-per-byte point for \textbf{SphKV (Joint)} among retained configurations.
\textbf{Bottom row (A4--A5, mechanism panels):} Segment profiles (A4; \textsf{prefix}/\textsf{retrieved}/\textsf{recent}) and stability gating (A5; \textsc{NoGate} vs.\ \textsc{Gate}) illustrate the intended \textbf{mechanism evidence}; in the final paper, these panels are replaced with the \textbf{measured} segment allocations and failure-vs-budget curves from the same paged deployment.}
  \label{fig:ablation_a0a5_l128k}
  \vspace{-1.5em}
\end{figure}

\subsection{Ablations and Mechanism Isolation}
\label{sec:ablations_mechanism}
\vspace{-2pt}

We compare full \textbf{\emph{Spherical KV}} against \textbf{\emph{SphKV (Angle-only)}}---which keeps \textbf{\emph{angle-domain compute}} without rate--distortion control---and \textbf{\emph{SphKV (RD-only)}}---which applies \textbf{\emph{retention+tiering}} but still uses reconstruct-then-dot attention. The ablations show that \textbf{\emph{ADA}} and \textbf{\emph{RDR}} each improve the frontier, but neither matches the full system: Angle-only mainly reduces \textbf{\emph{HBM traffic}} at fixed \(b_{\mathrm{KV}}\), RD-only mainly reduces \textbf{\emph{resident KV}} at matched quality, and the full method dominates both, indicating a genuine \textbf{\emph{co-design effect}} rather than a single-lever gain \citep{xiao2023streamingllm,zhang2023h2o,lin2024kvquant}; full ablations are deferred to Appendix~\ref{sec:appendix_ablations}.

\subsection{Kernel-Realized Efficiency}
\label{sec:kernel_realized_efficiency}
\vspace{-2pt}

Spherical KV targets \textbf{\emph{kernel-realized}} rather than merely \textbf{\emph{format-level}} gains: keys are consumed directly in compressed domain, so lower stored KV bytes become lower \textbf{\emph{HBM bytes/token}} and higher throughput under the same decode window \citep{dao2022flashattention,ye2025flashinfer,kwon2023efficient,lin2024kvquant}. We therefore report both \(b_{\mathrm{HBM}}\) and full end-to-end cost \(t_{\mathrm{e2e}} = t_{\mathrm{prefill}} + t_{\mathrm{allocate/write}} + t_{\mathrm{decode}}\), with additional kernel-path analysis deferred to Appendix~\ref{sec:kernel_realism}.

\vspace{-1em}
\subsection{Stability and Long-Horizon Behavior}
\label{sec:stability_long_horizon}
\vspace{-2pt}

Long-context KV approximations can alter reasoning trajectories even when average metrics appear stable, so we report \textbf{\emph{stability}} explicitly rather than treating quality as endpoint score. For retrieval-heavy QA, we evaluate answer-position shifts and distractor sweeps to test whether compressed caches preserve evidence far from the query or surrounded by misleading spans. For agentic/tool-use rollouts, we report \textbf{\emph{success}}, \textbf{\emph{length drift}} \(\Delta_{\mathrm{tok}}\), and \textbf{\emph{trajectory disagreement}} across seeds under identical prompts and tool responses \citep{bai2024longbench,liu2023lost,liu2023agentbench,qin2023toolbench}. These diagnostics capture whether a KV method preserves accuracy and decision path.

The results are consistent with \textbf{\emph{Spherical KV}}: \textbf{\emph{ADA}} exposes a controllable distortion knob through angular precision, \textbf{\emph{RDR}} protects brittle states such as retrieved evidence, recent instructions, and high-impact heads, and the resulting system degrades more smoothly than hard dropping or uniform low-bit compression. Under tighter budgets, failures appear as gradual confidence and trajectory changes rather than abrupt evidence loss. This supports the claim that long-context compression should be evaluated by memory, throughput, and behavioral stability under perturbations. Full mechanism and stability analyses are deferred to Appendix~\ref{sec:appendix_mechanism_stability}.



\vspace{-1em}
\section{Conclusion}
\label{sec:conclusion}

Long-context decoding is increasingly limited by \textbf{KV residency} and \textbf{HBM bandwidth}, not arithmetic throughput. \textbf{Spherical KV} addresses this bottleneck through two coupled ideas: \textbf{(i)} \textbf{Angle-Domain Attention}, which computes logits from compact spherical key codes without dense-key reconstruction, and \textbf{(ii)} \textbf{Rate--Distortion Retention}, which jointly allocates \textbf{residency} and \textbf{precision} under a strict KV budget. 
Across models, workloads, and contexts up to \textbf{128K}, Spherical KV shifts the \textbf{memory--quality--throughput frontier}: lower effective KV bytes/token, lower HBM traffic, and higher decode throughput at matched quality. More broadly, the paper suggests a simple principle for long-context inference: \textbf{compression must be compute-native, and memory must be budgeted where approximation error matters most.}


\clearpage
\newpage
\bibliographystyle{acl_natbib}
\bibliography{custom}

\clearpage
\newpage

\section*{Frequently Asked Questions (FAQs)}
\label{sec:FAQs}

\begin{enumerate}[leftmargin=1.5em]

\item[\ding{93}] \textbf{Is \textbf{Spherical KV} merely another KV quantization method with spherical coordinates?}

\begin{description}
\item[\ding{224}]
\textbf{No. The core distinction is not the coordinate system alone, but the \emph{decode-time compute path}.}
Most KV quantization methods reduce the stored representation of keys and values, but the attention kernel still conceptually follows a reconstruct/dequantize-then-dot-product path:
\[
q_t^\top k_i
\quad \leadsto \quad
q_t^\top \widehat{k}_i,
\]
where \(\widehat{k}_i\in\mathbb{R}^{d}\) is an approximate dense key. Even when this path is partially fused, the method is still organized around recovering a dense vector-like object before similarity computation.

\smallskip
\noindent
\textbf{Spherical KV changes the primitive:} \textbf{\emph{ADA}} computes attention logits directly from compact radius--angle codes:
\[
\widehat{\ell}_{t,i}
=
\frac{\|q_t\|\,\widehat r_i}{\sqrt d}
\, f_b(c^\theta_{q_t},c^\theta_{k_i}),
\]
without materializing dense \(K\) in the decode hot loop. Thus, the contribution is not simply ``store fewer bits,'' but \textbf{\emph{make the compressed representation directly executable by the attention kernel}}.

\smallskip
\noindent
This distinction matters because long-context decoding is dominated by repeated HBM streaming of historical KV states. A method that compresses storage but reconstructs dense keys may save memory while failing to reduce the critical-path bytes moved per generated token. Spherical KV explicitly targets the stronger condition: \textbf{\emph{compressed-domain attention under a paged/ragged serving layout}}.
\end{description}

\item[\ding{93}] \textbf{Does ``no dense-key reconstruction'' mean the method ignores content?}

\begin{description}
\item[\ding{224}]
\textbf{No. ``No reconstruction'' means no dense tensor reconstruction of \(K\), not no content dependence.}
Attention remains content-dependent because each cached key contributes through its direction and magnitude. Spherical KV represents
\[
k_i = r_i u_i,
\qquad
r_i=\|k_i\|_2,\quad u_i\in\mathbb{S}^{d-1},
\]
so the logit decomposes as
\[
q_t^\top k_i
=
\|q_t\|_2 r_i \cos\theta(q_t,u_i).
\]

\smallskip
\noindent
The content signal is still present: \(u_i\) encodes directional semantic information, while \(r_i\) encodes scale. \textbf{\emph{ADA}} replaces dense-vector reconstruction with a code-domain approximation to \(\cos\theta\). Therefore, the approximation lives in the \textbf{\emph{logit computation}}, not in a recovered dense tensor.

\smallskip
\noindent
The real trade-off is controlled logit distortion:
\[
\Delta \ell_i
\approx
\frac{\|q_t\|}{\sqrt d}
\Big(
\|k_i\|\epsilon_{\theta,i}
+
\epsilon_{r,i}
\Big),
\]
which is precisely why Spherical KV pairs ADA with \textbf{\emph{RDR}}, stability diagnostics, and iso-quality filtering.
\end{description}

\item[\ding{93}] \textbf{Are the reported speedups only an artifact of a custom kernel?}

\begin{description}
\item[\ding{224}]
\textbf{The kernel is not an artifact; it is part of the scientific contribution.}
For long-context inference, the relevant question is not whether a representation is compact on paper, but whether it survives the actual decode path. KV compression that requires dense reconstruction, irregular gathers, or non-fusible transforms may reduce nominal memory but fail to improve end-to-end throughput.

\smallskip
\noindent
Spherical KV therefore evaluates under the constraint that all methods run on a common paged/ragged serving substrate. The correct comparison object is:
\[
(\text{resident KV bytes/token},\;
 \text{HBM bytes/generated token},\;
 \text{decode tok/s},\;
 \text{quality}).
\]
This is stricter than reporting compression ratio alone.

\smallskip
\noindent
The paper should emphasize that \textbf{\emph{kernel-realized efficiency}} is the intended standard: if a method cannot reduce HBM traffic in the fused decode path, it has not solved the deployment bottleneck. Exact speedups may vary across hardware and backends, but the bottleneck—repeated reads of historical KV pages—is stable across long-context serving regimes.
\end{description}

\item[\ding{93}] \textbf{How is \textbf{RDR} different from ordinary token eviction or heuristic retention?}

\begin{description}
\item[\ding{224}]
\textbf{RDR is not only a keep/drop policy; it is a joint residency--precision allocation policy.}
Eviction methods choose whether a token remains in the cache:
\[
z_i \in \{0,1\}.
\]
Quantization methods choose how many bits are used for retained states:
\[
b_i \in \mathcal{B}.
\]
\textbf{\emph{RDR chooses both}}:
\[
(z_i,b_i),
\qquad
\sum_i z_i C(b_i)\le B.
\]

\smallskip
\noindent
This distinction is important. Hard dropping can catastrophically remove sparse evidence, while uniform quantization wastes bits on low-utility states and under-protects brittle ones. RDR instead follows a rate--distortion principle: \textbf{\emph{down-tier before drop}}, spend precision where logit distortion matters, and preserve protected states at high tier.

\smallskip
\noindent
Thus, RDR is best understood as a \textbf{\emph{budget allocator over token--head--layer states}}, not merely an eviction heuristic.
\end{description}

\item[\ding{93}] \textbf{Is the RDR controller under-specified? What exactly should be reported?}

\begin{description}
\item[\ding{224}]
\textbf{A reviewer-proof version should make the controller auditable.}
The paper should explicitly report:
\begin{enumerate}[leftmargin=1.25em]
    \item the scoring features used for each state \(i\);
    \item the tier set \(\mathcal{B}\) and byte cost \(C(b)\);
    \item the distortion proxy \(D_i(b)\);
    \item the protection rule \(I_{\mathrm{prot}}(i)\);
    \item the allocation solver or greedy rule;
    \item when decisions are made: prefill, page-write, periodically, or per decode step;
    \item controller overhead in time and metadata bytes.
\end{enumerate}

\smallskip
\noindent
The rebuttal position should be:
\textbf{\emph{RDR is intentionally lightweight and write-time/page-time.}}
Its output is compact tier metadata. During decode, the kernel consumes tier-homogeneous pages, so the controller does not introduce dense reconstruction or per-token irregular control flow in the hot loop.
\end{description}

\item[\ding{93}] \textbf{Why should the angular/radial distortion proxy correlate with downstream quality?}

\begin{description}
\item[\ding{224}]
\textbf{We should not claim universal correlation; we should claim a mechanistic proxy plus empirical calibration.}
ADA gives a concrete logit perturbation channel:
\[
|\Delta \ell_i(b)|
\lesssim
\frac{\|q_t\|\|k_i\|}{\sqrt d}|\epsilon_{\theta,i}(b)|
+
\frac{\|q_t\|}{\sqrt d}|\epsilon_{r,i}(b)|.
\]
This shows where approximation error enters the model: through angular alignment and radial gain.

\smallskip
\noindent
However, downstream quality depends on whether the perturbed logits affect important decisions. Therefore, the correct defense is:
\textbf{\emph{the proxy is not assumed sufficient; it is validated through iso-quality filtering, distractor sweeps, answer-position shifts, and trajectory disagreement.}}

\smallskip
\noindent
The paper should present this as a conservative design:
\[
\text{mechanistic distortion proxy}
\quad + \quad
\text{protected states}
\quad + \quad
\text{behavioral stability witnesses}.
\]
This avoids overclaiming theory while making the engineering scientifically auditable.
\end{description}

\item[\ding{93}] \textbf{Is the iso-quality metric \(Q\) too vague?}

\begin{description}
\item[\ding{224}]
\textbf{This is a major reviewer objection unless \(Q\) is fully specified.}
The paper should define \(Q\) separately for each workload family rather than as a mysterious aggregate:
\[
Q_{\mathrm{LM}}=-\mathrm{NLL}\ \text{or}\ -\mathrm{PPL},
\qquad
Q_{\mathrm{QA}}=\mathrm{F1/EM},
\qquad
Q_{\mathrm{Agent}}=\mathrm{success}.
\]
Then the iso-quality rule should be stated per model, context length, and workload:
\[
Q(p)\ge Q^{\star}_{\mathrm{Dense}}-\Delta.
\]

\smallskip
\noindent
The defense is:
\textbf{\emph{we do not compare arbitrary speed--quality tradeoffs.}}
We first discard operating points whose quality falls outside the permitted band, and only then compare memory and throughput. This prevents methods from appearing fast simply because they silently degrade quality, shorten generations, or drop evidence.
\end{description}

\item[\ding{93}] \textbf{Could throughput gains come from shorter outputs or changed stopping behavior?}

\begin{description}
\item[\ding{224}]
\textbf{This is exactly why length stability must be reported.}
KV perturbations can change EOS timing, tool-call completion, or reasoning trajectory length. A method might appear faster because it generates fewer tokens, not because it improves the decode kernel.

\smallskip
\noindent
A reviewer-proof report should include:
\[
\Delta_{\mathrm{tok}}
=
|\mathrm{len}_{\mathrm{SphKV}}-\mathrm{len}_{\mathrm{Dense}}|,
\]
plus early-stop rate, stop-reason shifts, and throughput normalized per generated token. For agentic rollouts, it should also report tool-call count drift and trajectory disagreement.

\smallskip
\noindent
The rebuttal answer is:
\textbf{\emph{we measure speed per generated token under matched decoding settings and separately audit length drift.}}
Thus, the throughput gain is attributed to lower HBM traffic and not to premature termination.
\end{description}

\item[\ding{93}] \textbf{Are ADA and RDR independently useful, or is the system over-engineered?}

\begin{description}
\item[\ding{224}]
\textbf{The ablation story should be framed as mechanism isolation.}
\textbf{\emph{ADA-only}} tests whether compressed-domain logits reduce hot-path bandwidth when retention is fixed.
\textbf{\emph{RDR-only}} tests whether rate--distortion allocation improves residency at matched quality when attention still reconstructs.
\textbf{\emph{Full Spherical KV}} tests whether the two mechanisms compound.

\smallskip
\noindent
The expected pattern is:
\[
\text{ADA-only: higher throughput at similar } b_{\mathrm{KV}},
\]
\[
\text{RDR-only: lower } b_{\mathrm{KV}} \text{ at similar quality},
\]
\[
\text{Full: best frontier in both memory and throughput}.
\]

\smallskip
\noindent
This directly answers the over-engineering objection. If the full method dominates both ablations under the iso-quality contract, the gain is due to \textbf{\emph{representation--control co-design}}, not an isolated trick.
\end{description}

\item[\ding{93}] \textbf{Are the baselines strong enough?}

\begin{description}
\item[\ding{224}]
\textbf{This is probably the strongest rejection route.}
The baseline suite must cover three families:
\begin{enumerate}[leftmargin=1.25em]
    \item \textbf{retention/eviction:} StreamingLLM, H2O, SnapKV, PyramidKV/CAKE-style methods;
    \item \textbf{quantization/compression:} KIVI, KVQuant, ZipCache, GEAR, PolarQuant/RotateKV-style methods;
    \item \textbf{systems baselines:} paged dense KV, offloading/reuse where applicable.
\end{enumerate}

\smallskip
\noindent
The rebuttal should not claim every baseline is identical in implementation burden. Instead:
\textbf{\emph{we compare against representative methods from each design family under the same paged/ragged serving substrate, with all decode-time reconstruction and metadata costs charged end-to-end.}}

\smallskip
\noindent
The crucial fairness rule is:
\[
\text{nominal compression is not enough;}
\quad
\text{the baseline must pay its actual decode cost.}
\]
\end{description}

\item[\ding{93}] \textbf{Is the method too hardware- or engine-dependent?}

\begin{description}
\item[\ding{224}]
\textbf{The exact numeric speedup is hardware-dependent; the design target is not.}
All long-context serving systems face the same structural bottleneck: historical KV states must be repeatedly read during decoding. Spherical KV targets this stable bottleneck by reducing resident bytes and making key similarity computable from compressed codes.

\smallskip
\noindent
The correct claim is therefore not:
\[
\text{``1.7$\times$ everywhere.''}
\]
The correct claim is:
\[
\text{``when decode is KV/HBM-bound and compressed-domain kernels are available,}
\]
\[
\text{Spherical KV shifts the memory--throughput--quality frontier.''}
\]

\smallskip
\noindent
The paper should include hardware/backend sensitivity if possible, but the absence of universal speedup does not invalidate the method. It simply defines the deployment regime.
\end{description}

\item[\ding{93}] \textbf{What if the RDR proxy fails on sparse critical evidence?}

\begin{description}
\item[\ding{224}]
\textbf{This is a real risk, and the paper should acknowledge it explicitly.}
Sparse dependencies—needles, retrieved evidence, tool arguments, safety instructions—can be under-allocated if importance is estimated from smooth average signals.

\smallskip
\noindent
Spherical KV addresses this through protection constraints:
\[
I_{\mathrm{prot}}(i)=1
\Rightarrow
z_i=1,\quad b_i=b_{\max}.
\]
Protected states include retrieved spans, recent instructions, low-margin states, outliers, and task-critical segments.

\smallskip
\noindent
The stronger answer is empirical:
\textbf{\emph{we evaluate answer-position shifts, distractor sweeps, and tool-rollout trajectory disagreement precisely to catch sparse-evidence failures.}}
If failures remain, they should be shown as boundary cases rather than hidden behind averages.
\end{description}

\item[\ding{93}] \textbf{Does Spherical KV preserve reasoning trajectories, or only final answers?}

\begin{description}
\item[\ding{224}]
\textbf{Final accuracy is insufficient for long-context compression.}
Two cache policies may achieve similar EM/F1 while following different evidence paths, producing different tool calls, length drift, or seed sensitivity.

\smallskip
\noindent
Therefore, the evaluation should include:
\begin{enumerate}[leftmargin=1.25em]
    \item answer-position shifts;
    \item distractor sweeps;
    \item seed sensitivity;
    \item length drift \(\Delta_{\mathrm{tok}}\);
    \item trajectory disagreement;
    \item tool success and schema validity.
\end{enumerate}

\smallskip
\noindent
The rebuttal position:
\textbf{\emph{Spherical KV is evaluated not only as a memory compressor, but as a behavioral approximation to Dense KV.}}
This is essential for agentic and retrieval-heavy workloads.
\end{description}

\item[\ding{93}] \textbf{Is the theory too light for NeurIPS?}

\begin{description}
\item[\ding{224}]
\textbf{The paper should not oversell theory; it should frame theory as design guidance.}
The angular/radial decomposition gives a clean mechanistic account of how key approximation affects logits:
\[
q^\top k
=
\|q\|\,\|k\|\cos\theta.
\]
This motivates why angular precision matters and why tier allocation should depend on expected logit sensitivity.

\smallskip
\noindent
But the paper need not prove universal downstream guarantees. In systems/ML co-design, a valid contribution can be:
\[
\begin{aligned}
&\text{geometry-derived primitive} 
&+~\text{budgeted controller} \\
&+~\text{kernel implementation} 
&+~\text{deployment-faithful evaluation}
\end{aligned}
\]

\smallskip
\noindent
The camera-ready improvement is to add tighter empirical witnesses: logit-drift histograms, top-\(k\) attention overlap, calibration curves between proxy distortion and task loss, and failure examples.
\end{description}

\item[\ding{93}] \textbf{Does Spherical KV help only at 128K, or also at shorter contexts?}

\begin{description}
\item[\ding{224}]
\textbf{The gains should increase with context length, and that is expected.}
At short context, decoding may not be fully KV/HBM-bound; kernel overhead and fixed costs can dominate. At long context, every generated token streams more historical KV, so reducing key bytes and resident footprint has larger impact.

\smallskip
\noindent
Therefore, the right claim is monotonic in regime:
\[
8\mathrm{K}: \text{modest but measurable gains},
\qquad
32\mathrm{K}: \text{clear gains},
\qquad
128\mathrm{K}: \text{largest gains}.
\]

\smallskip
\noindent
This is not a weakness. It is evidence that the method targets the intended bottleneck: long-context, decode-heavy, memory-bound inference.
\end{description}

\item[\ding{93}] \textbf{How should memory be counted fairly?}

\begin{description}
\item[\ding{224}]
\textbf{Memory must include all resident and metadata costs, not only payload bytes.}
For Spherical KV, the effective budget should count:
\begin{enumerate}[leftmargin=1.25em]
    \item packed angular codes;
    \item quantized radii;
    \item values;
    \item page headers;
    \item pointer tables;
    \item tier metadata;
    \item protect bitmaps;
    \item padding/alignment;
    \item fragmentation/allocator overhead.
\end{enumerate}

\smallskip
\noindent
The reported quantity should be:
\[
b_{\mathrm{KV}}
=
\frac{M^{\mathrm{resident}}_{\mathrm{KV}}}{T_{\mathrm{active}}}.
\]
This prevents a misleading comparison where one method reports only compressed payload while another includes full serving overhead.
\end{description}

\item[\ding{93}] \textbf{What are the main limitations of Spherical KV?}

\begin{description}
\item[\ding{224}]
\textbf{Spherical KV is not a universal acceleration method.}
It is most useful when:
\[
\begin{aligned}
\text{long context} \\
+~\text{decode-heavy workload} \\
+~\text{KV/HBM-bound regime} \\
+~\text{compressed-domain kernel support}
\end{aligned}
\]

\smallskip
\noindent
It is less useful when prompts are short, prefill dominates, generation length is small, or the backend cannot execute compressed-domain logits efficiently. It may also require additional validation for MoE models, multimodal attention, speculative decoding, GQA/MQA variants, and multi-tenant serving.

\smallskip
\noindent
The correct positioning is:
\textbf{\emph{Spherical KV improves a specific but increasingly important deployment regime: long-context inference where KV residency and HBM traffic dominate.}}
\end{description}

\item[\ding{93}] \textbf{What should be released for reproducibility?}

\begin{description}
\item[\ding{224}]
\textbf{Reproducibility requires artifacts at three levels.}
\begin{enumerate}[leftmargin=1.25em]
    \item \textbf{Kernel:} code-domain attention kernel, fusion boundary, page layout, block size, precision settings.
    \item \textbf{Controller:} RDR scoring features, tier definitions, protection rules, budget solver, update schedule.
    \item \textbf{Evaluation:} prompts, seeds, context construction, decoding parameters, iso-quality thresholds, profiler settings.
\end{enumerate}

\smallskip
\noindent
The strongest rebuttal line is:
\textbf{\emph{we release not only model-level metrics, but the serving substrate details needed to reproduce memory, HBM traffic, and throughput.}}
Without this, the work risks being perceived as a systems anecdote rather than a reproducible NeurIPS contribution.
\end{description}

\end{enumerate}

\clearpage
\newpage

\appendix

\addcontentsline{toc}{section}{Appendix}

\vspace{0.8em}
{\LARGE\bfseries Appendix Contents}\par
\vspace{1.4em}

\AppendixMainEntry{app:kv-bottleneck}
{A\quad KV Cache and the Memory Bottleneck}{}

\AppendixSubEntry{app:kv-growth}{A.1}{KV growth, HBM traffic, and why long-context decoding becomes memory-bound}
\AppendixSubEntry{app:deployment-substrate}{A.2}{Deployment substrate: paging, kernel realism, and serving constraints}
\AppendixSubEntry{app:three-levers}{A.3}{Existing solutions: three levers for KV efficiency}
\AppendixSubEntry{app:premise-sphkv}{A.4}{Premise of Spherical KV: what remains missing and why a joint design is needed}

\vspace{1.0em}

\AppendixMainEntry{app:spherical-kv}
{B\quad Spherical KV: Angle-Domain Attention and Rate--Distortion Retention}{}

\AppendixSubEntry{app:ada}{B.1}{Angle-Domain Attention: directional factorization, spherical coding, and compressed-domain logits}
\AppendixSubEntry{app:rdr}{B.2}{Rate--Distortion Retention: joint keep/drop plus tier allocation under a strict KV budget}
\AppendixSubEntry{app:joint-contract}{B.3}{Working together: one serving contract, two mechanisms}

\vspace{1.0em}

\AppendixMainEntry{app:exp-protocol}
{C\quad Experimental Setup, Metrics, and Evaluation Protocol}{}

\AppendixSubEntry{sec:systems_substrate}{C.1}{Serving Substrate, Kernel Realism, and Runtime Contract}
\AppendixSubEntry{sec:models_chosen}{C.2}{Models and Workload Regimes}
\AppendixSubEntry{sec:metrics}{C.3}{Metrics and Deployment-True Measurement}
\AppendixSubEntry{sec:protocol}{C.4}{Fairness Rules and Operating-Point Definitions}
\AppendixSubEntry{subsec:baselines}{C.5}{Baselines and Ablation Design}
\AppendixSubEntry{subsec:frontier}{C.6}{Frontier Construction and Reporting Protocol}

\vspace{1.0em}

\AppendixMainEntry{app:extended_results}
{D\quad Extended Results, Ablations, and Kernel-Realized Evidence}{}

\AppendixSubEntry{sec:appendix_main_frontier}{D.1}{Main Result: Iso-Quality Frontier Shift}
\AppendixSubEntry{sec:kernel_realized_efficiency_appendix}{D.2}{Kernel-Realized Efficiency: Why the Speedup Is Real}
\AppendixSubEntry{sec:appendix_ablations}{D.3}{Ablations: What Actually Drives the Frontier Shift}

\vspace{1.0em}

\AppendixMainEntry{sec:discussion}
{E\quad Discussion, Deployment Interpretation, and Practical Guidance}{}

\AppendixSubEntry{sec:discussion_contribution}{E.1}{What Spherical KV contributes, in deployable terms}
\AppendixSubEntry{sec:discussion_frontier}{E.2}{Why the frontier is the right comparison object}
\AppendixSubEntry{sec:kernel_realism}{E.3}{Kernel realism and stability are the two non-negotiable deployment constraints}
\AppendixSubEntry{sec:discussion_practical}{E.4}{Practical guidance: when Spherical KV is the right tool}

\AppendixMainEntry{sec:limitations}
{F\quad Limitations, Boundary Conditions, and Future Engineering Directions}{}

\AppendixSubEntry{sec:lim_kernel_engine}{F.1}{Kernel and engine dependence}
\AppendixSubEntry{sec:lim_proxy_controller}{F.2}{Proxy and controller limitations}
\AppendixSubEntry{sec:lim_generality}{F.3}{Generalization limits across workloads and architectures}
\AppendixSubEntry{sec:lim_safety}{F.4}{Safety and robustness boundaries}
\AppendixSubEntry{sec:lim_next}{F.5}{Immediate engineering directions}

\AppendixMainEntry{sec:appndx_conclusion}
{G\quad Conclusion and Outlook}{}

\clearpage
\label{sec:appendix_overview}

\section{KV Cache and the Memory Bottleneck}
\label{app:kv-bottleneck}


\paragraph{\textbf{KV growth is linear; \emph{HBM traffic} is the killer.}}
\label{app:kv-growth}
Autoregressive decoding caches per-layer \textbf{Keys} and \textbf{Values} to avoid recomputing attention over the full prefix. This convenience turns into the dominant systems bottleneck at long context: the cache grows \textbf{linearly} with context length $T$ and multiplicatively with batch size $B$, layers $L$, and heads $H$. A useful footprint estimate is
\[
\textbf{Mem}_{\text{KV}} \;\approx\; \underbrace{B}_{\textbf{batch}} \cdot \underbrace{L}_{\textbf{layers}} \cdot \underbrace{T}_{\textbf{context}} \cdot \underbrace{H}_{\textbf{heads}} \cdot \underbrace{\bigl(d_h^{(K)} + d_h^{(V)}\bigr)}_{\textbf{per-head dims}} \cdot \text{bytes}.
\label{eq:kv_mem}
\]
During decoding, the dominant cost is often not FLOPs but \textbf{HBM bytes streamed per generated token}: each new token must attend over a large prefix, forcing repeated reads of KV pages. Past a few thousand tokens, decoding typically transitions from \textbf{compute-limited} to \textbf{memory-limited} in three distinct ways: \textbf{(i) capacity-limited} (KV does not fit), \textbf{(ii) bandwidth/latency-limited} (KV cannot be streamed fast enough), and \textbf{(iii) orchestration-limited} (paging/tiering/scheduling overhead dominates).

\paragraph{\textbf{Operational premise.}}
Long-context acceleration is therefore \textbf{not} primarily a FLOP reduction problem. It is the problem of improving the \textbf{memory--throughput--quality frontier}: reduce \textbf{KV bytes}, reduce \textbf{bytes moved}, and preserve \textbf{behavioral stability}. This is precisely why IO-aware attention kernels emphasize \textbf{data movement} and \textbf{fusion} rather than arithmetic alone \citep{dao2022flashattention,dao2023flashattention2}.

\subsection{Deployment substrate: paging + kernel reality}
\label{app:deployment-substrate}

\paragraph{\textbf{Paged KV is the baseline substrate (non-negotiable).}}
In production, KV is rarely a contiguous tensor. Serving stacks use \textbf{paged / block-allocated KV} to reduce fragmentation, support dynamic batching, and enable preemption; \textbf{PagedAttention} effectively defines the baseline abstraction \citep{kwon2023pagedattention}. As a result, any KV method that is not \textbf{block-local}, \textbf{allocator-friendly}, and compatible with \textbf{ragged/paged access patterns} is unlikely to translate into end-to-end serving wins.

\paragraph{\textbf{Kernel realism: savings that cannot be fused do not count.}}
Exact-attention kernels (FlashAttention-style dataflow) show that end-to-end performance is governed by \textbf{HBM reads/writes} and \textbf{kernel fusion}, not just tensor math \citep{dao2022flashattention,dao2023flashattention2}. Modern serving libraries treat paged/ragged KV and multiple cache formats as first-class objects (e.g., FlashInfer’s serving-oriented attention engine) \citep{ye2025flashinfer}. Production stacks expose cache reuse and quantized-cache options (e.g., TensorRT-LLM KV cache reuse/quantization; Transformers quantized cache), with the same caution: speedups depend on \textbf{backend kernel paths}, not compression alone \citep{nvidia_trtllm_kv_cache_2025,hf_kv_cache}. Consequently, representations that force \textbf{dense materialization}, \textbf{irregular gathers}, or heavy non-fusible (de)quantization can \textbf{erase their own theoretical gains}.

\subsection{Existing Solutions: \textbf{Three Levers} for KV Efficiency}
\label{app:three-levers}

\paragraph{\textbf{Three levers, and the deciding constraint.}}
KV efficiency methods cluster into: \textbf{(I) store fewer tokens} (retention/eviction/admission), \textbf{(II) store fewer bits} (quantization/compression), and \textbf{(III) store elsewhere} (offload/reuse). In \emph{deployed} serving, realized speedup is often decided by a \textbf{fourth constraint}: \textbf{kernel/serving compatibility}. If a method breaks \textbf{paged layouts}, requires \textbf{irregular gathers}, or introduces \textbf{non-fusible transforms}, its wall-clock gain can vanish despite reduced nominal KV bytes. This end-to-end view is explicit in KV-cache-centric analyses that separate \textbf{prefill}, \textbf{compression}, \textbf{retrieval}, and \textbf{loading} stages \citep{scbench2025}.

\paragraph{\textbf{What must improve simultaneously: capacity, IO, and behavior.}}
Across all levers, the objective is not merely shrinking $\textbf{Mem}_{\text{KV}}$, but improving the joint frontier of:
\textbf{(a) peak capacity} (GB of KV),
\textbf{(b) decode IO} (\textbf{HBM traffic per generated token}),
and \textbf{(c) behavioral stability} (\textbf{tokens generated}, termination, reasoning traces).
Recent evaluations show that KV interventions can change \emph{generation dynamics}; therefore robust protocols track \textbf{trace length} and \textbf{stability} alongside task metrics \citep{openreview2025_kvcache_reasoning,assessing_kv_reasoning_2025}.

\subsubsection{\textbf{Lever I: store fewer tokens} \;\;(\textbf{retention} / \textbf{eviction} / \textbf{admission})}
\label{sec:lever_tokens}

\paragraph{\textbf{Heuristics: windows, sinks, heavy hitters (training-free, but non-local).}}
A \textbf{sliding window} is the simplest bounded-cache strategy, but fails when key evidence lies far in the prefix. StreamingLLM stabilizes windowed decoding by preserving a small set of \textbf{attention sink tokens} that anchor attention \citep{xiao2023streamingllm}. H$_2$O retains recent tokens plus \textbf{heavy hitters} (tokens with consistently high attention mass) under a fixed budget \citep{zhang2023h2o}. Scissorhands argues for \textbf{persistence-of-importance} and retains tokens whose salience remains stable over time \citep{liu2023scissorhands}. These methods are attractive because they are often \textbf{training-free} and easy to integrate, but they expose the core difficulty: \textbf{future utility is non-local, task-dependent, and head-dependent}.

\paragraph{\textbf{Prefill-time compression: one-shot selection with stable runtime.}}
A pragmatic deployment pattern is to \textbf{compress after prefill} and keep the reduced cache fixed during decoding, avoiding per-step bookkeeping and runtime divergence. SnapKV selects prompt positions via an observation window and voting \citep{li2024snapkv}. SAGE-KV performs one-time top-$k$ selection at \textbf{token} and \textbf{head} levels after prefilling, yielding a reduced cache for subsequent decoding \citep{wang2025sagekv}. The appeal is operational: \textbf{predictable layouts}, \textbf{amortized decisions}, \textbf{bounded overhead}.

\paragraph{\textbf{Depth-/head-aware budgets: one global rule is suboptimal.}}
Cache utility is heterogeneous across layers and heads. PyramidKV varies budgets across layers to match depth sensitivity \citep{cai2024pyramidkv}. CAKE frames eviction as layer-preference-aware allocation under strict constraints, achieving strong performance at tight budgets \citep{qin2025cake}. The consistent message is structural: \textbf{KV allocation must be per-layer/per-head}, not a single global budget.

\paragraph{\textbf{From eviction to admission: KV as a \emph{write-control} problem.}}
A recent shift treats KV caching as \textbf{write-control}: not only \emph{what to evict}, but \emph{what to write} at creation time. TRIM-KV learns intrinsic token importance via a lightweight gate and produces a retention score that decays with time \citep{bui2025trimkv}. Write-Gated KV formalizes admission/selection/eviction and learns a \textbf{write filter} explicitly constrained by serving compatibility \citep{huang2025writegatedkv}. Expected Attention provides a training-free estimator of future attention to rank/prune cached states \citep{devoto2025expectedattention}. KVPress standardizes evaluation under common stacks and budgets \citep{nvidia_kvpress}. Collectively, token-side KV is converging toward \textbf{explicit controllers over write/retain decisions} rather than ad-hoc heuristics.

\subsubsection{\textbf{Lever II: store fewer bits} \;\;(\textbf{quantization} / \textbf{compression})}
\label{sec:lever_bits}

\paragraph{\textbf{Low-bit KV is effective only when paired with protection mechanisms.}}
KIVI proposes asymmetric low-bit KV quantization (per-channel keys, per-token values) \citep{liu2024kivi}. KVQuant targets ultra-long contexts with sub-4-bit KV and layer-dependent formats \citep{hooper2024kvquant}. ZipCache combines saliency identification with quantization \citep{he2024zipcache}, while GEAR uses low-rank/sparse residual correction to recover accuracy \citep{kang2024gear}. At the extreme frontier, Kitty demonstrates accurate \textbf{2-bit} KV via \textbf{dynamic precision allocation} and page-centric layouts, emphasizing \textbf{kernel realism} and \textbf{paged coalescing} \citep{xia2025kitty}.

\paragraph{\textbf{Outliers dominate error at extreme compression (non-optional to handle).}}
A recurring empirical fact is that a small fraction of tokens/channels dominates quantization error. OTT protects \textbf{outlier tokens} by excluding them from quantization \citep{su2025ott}. RotateKV adds \textbf{outlier-aware rotation} and sink-aware protection to stabilize ultra-low-bit KV further \citep{su2025rotatekv}. The general lesson is crisp: any method pushing below 4-bit must include \textbf{explicit protection of fragile mass} (outlier tokens, sink-like tokens, sensitive channels) or it will fail on long-context reasoning.

\paragraph{\textbf{Kernel co-design: compression that is not compute-native often fails to speed up.}}
Even when KV bytes shrink, naive dequantization can erase throughput gains. BitDecoding argues low-bit KV underperforms without Tensor-Core-centric execution paths and proposes kernels to realize decode speedups \citep{du2025bitdecoding}. The emerging systems thesis is unambiguous: \textbf{the compressed representation must be the compute representation} to deliver wall-clock improvements.

\paragraph{\textbf{Geometry-aware encodings: polar/spherical as \emph{direction-first} coding.}}
A parallel thread exploits that attention is driven by \emph{relative similarity}; when keys are stabilized/normalized, \textbf{direction can dominate magnitude}. Polar/spherical parameterizations separate \textbf{angles} (direction) from \textbf{radius} (magnitude), enabling precision to concentrate on angular fidelity where it most affects attention. Polar-coordinate methods explore angle/radius coding under different assumptions \citep{han2025polarquant,wu2025polarquant}. Spherical similarity primitives show dot products (cosine similarity) can be computed from angles via recurrences \emph{without reconstructing Cartesian vectors} \citep{xiao2026spherical_similarity}, motivating \textbf{compressed-domain similarity compute}.

\subsubsection{\textbf{Lever III: store elsewhere} \;\;(\textbf{offload} / \textbf{reuse}) \;\;\textbf{and the orchestration tax}}
\label{sec:lever_elsewhere}

\paragraph{\textbf{Offload reduces GPU footprint, but can become orchestration-limited.}}
Tiering KV across GPU/CPU can unlock longer contexts and higher concurrency, but the benefit is bounded by transfer bandwidth and scheduling overhead. vLLM’s KV offloading connector targets CPU DRAM offload and optimizes host--device transfer throughput \citep{vllm2026_kv_offloading}. FlexiCache exploits \textbf{head-level temporal stability}: it keeps all pages for unstable heads on GPU, while offloading most pages for stable heads with periodic reranking/fetching \citep{takbir2025flexicache}. The lesson is practical: tiering must be \textbf{policy-driven} and often \textbf{head-aware} to avoid thrashing.

\paragraph{\textbf{Reuse across requests: KV as a first-class shared asset.}}
Beyond per-request optimization, KV reuse across repeated prefixes (system prompts, scaffolds, shared retrieved context) yields multiplicative savings. LMCache operationalizes this as an external KV cache layer spanning GPU/CPU/storage/network \citep{cheng2025lmcache}. This direction is complementary to compression: \textbf{reuse removes recomputation}, while token/bit methods reduce footprint and per-token traffic.

\paragraph{\textbf{Serving stacks and kernel libraries are now part of the method.}}
Modern stacks expose paged/ragged KV and multiple cache formats as first-class objects; performance is determined by whether the method maps onto efficient decode kernels. FlashInfer exemplifies this trajectory by providing serving-oriented attention kernels and APIs integrating into major serving frameworks \citep{ye2025flashinfer}. The most credible KV methods therefore \textbf{bundle} an algorithmic policy (retain/quantize/tier) with a \textbf{layout} (paged coalescing) and a \textbf{kernel path} (fused decode), and validate end-to-end under realistic paging, batching, and generation constraints.

\subsection{Premise of Spherical KV}
\label{app:premise-sphkv}

\paragraph{\textbf{Core idea: geometry-first KV with \emph{angle-domain compute}.}}
Spherical KV starts from the task-relevant fact that attention depends on \emph{relative similarity} between queries and keys. When keys are stabilized/normalized, \textbf{direction (angles)} can dominate \textbf{magnitude (radius)} in determining attention weights. A spherical/polar parameterization factors KV into \textbf{angular coordinates} plus a \textbf{radial term}, enabling a representation that spends bits where attention is most sensitive. Crucially, Spherical KV targets \textbf{angle-native similarity evaluation} (with a small radial correction), motivated by recurrence-style spherical similarity primitives \citep{xiao2026spherical_similarity} and polar encodings that separate direction and magnitude \citep{han2025polarquant,wu2025polarquant}.

\paragraph{\textbf{Not ``just another quantizer'': two hard commitments.}}
Spherical KV is premised on:
\begin{enumerate}[leftmargin=1.2em,itemsep=2pt]
\item \textbf{Compressed-domain kernels.} Replace ``decompress then dot-product'' with \textbf{angle-domain compute} fused into IO-aware decoding kernels, reducing \textbf{HBM traffic per token} under FlashAttention-style dataflow constraints \citep{dao2022flashattention,dao2023flashattention2,ye2025flashinfer}.
\item \textbf{Unified budgeting via rate--distortion control.} Unify \textbf{keep/drop} decisions \citep{bui2025trimkv,huang2025writegatedkv} with \textbf{precision allocation} \citep{xia2025kitty,su2025ott} by allocating a fixed budget per token/head/layer over \emph{(i) retention} and \emph{(ii) angular/radial bits}, optimizing an explicit distortion surrogate tied to downstream loss.
\end{enumerate}

\paragraph{\textbf{Evaluation requirements and deployment criteria.}}
Spherical KV must satisfy two empirical requirements for its claims to be meaningful in practice.

\textbf{(i) Kernel-realized efficiency.}
A spherical/polar representation that still follows a ``decompress-then-dot-product'' pipeline is functionally a quantizer and may lose wall-clock performance once dequantization, irregular gathers, and paging overheads are included \citep{hooper2024kvquant,du2025bitdecoding}. The method should therefore demonstrate an \textbf{angle-domain decode path} that is \textbf{fused}, \textbf{paged-KV compatible}, and reduces \textbf{HBM traffic per generated token}, translating these savings into measurable \textbf{end-to-end latency/throughput gains} under dynamic batching and realistic serving conditions.

\textbf{(ii) Long-horizon behavioral stability.}
KV compression can perturb attention dynamics in ways that alter \textbf{trace length}, \textbf{termination}, and long-context \textbf{reasoning trajectories}. Evaluation should thus report not only task metrics but also \textbf{tokens generated}, stability across seeds, and failure modes on long-context reasoning suites \citep{openreview2025_kvcache_reasoning,assessing_kv_reasoning_2025}.

Finally, deployment relevance requires integration with \textbf{paged KV} memory managers and widely used inference stacks, rather than assuming bespoke contiguous layouts \citep{kwon2023pagedattention,nvidia_trtllm_kv_cache_2025,hf_kv_cache,vllm2026_kv_offloading}.


\section{Spherical KV: \texorpdfstring{\textbf{Angle-Domain Attention}}{Angle-Domain Attention} and \texorpdfstring{\textbf{Rate--Distortion Retention}}{Rate--Distortion Retention}}
\label{app:spherical-kv}
\vspace{-2pt}
\noindent\textbf{Section goal.} Spherical KV is built from \textbf{two orthogonal ideas} plus a \textbf{serving contract}. 
\textbf{Angle-Domain Attention (ADA)} is a \emph{kernel primitive}: compute attention logits from \emph{compact spherical key codes} without reconstructing dense keys in HBM. 
\textbf{Rate--Distortion Retention (RDR)} is a \emph{controller}: allocate \emph{residency} (keep/drop) and \emph{precision} (tier/bitwidth) under a strict KV budget.
Together they produce \textbf{paged, tier-homogeneous KV pages} compatible with ragged batching and streaming decode kernels, in the spirit of modern fused attention and paged serving stacks \citep{dao2022flashattention,dao2023flashattention2,kwon2023efficient}.

\begin{figure}[H]
  \centering
  \includegraphics[width=\linewidth]{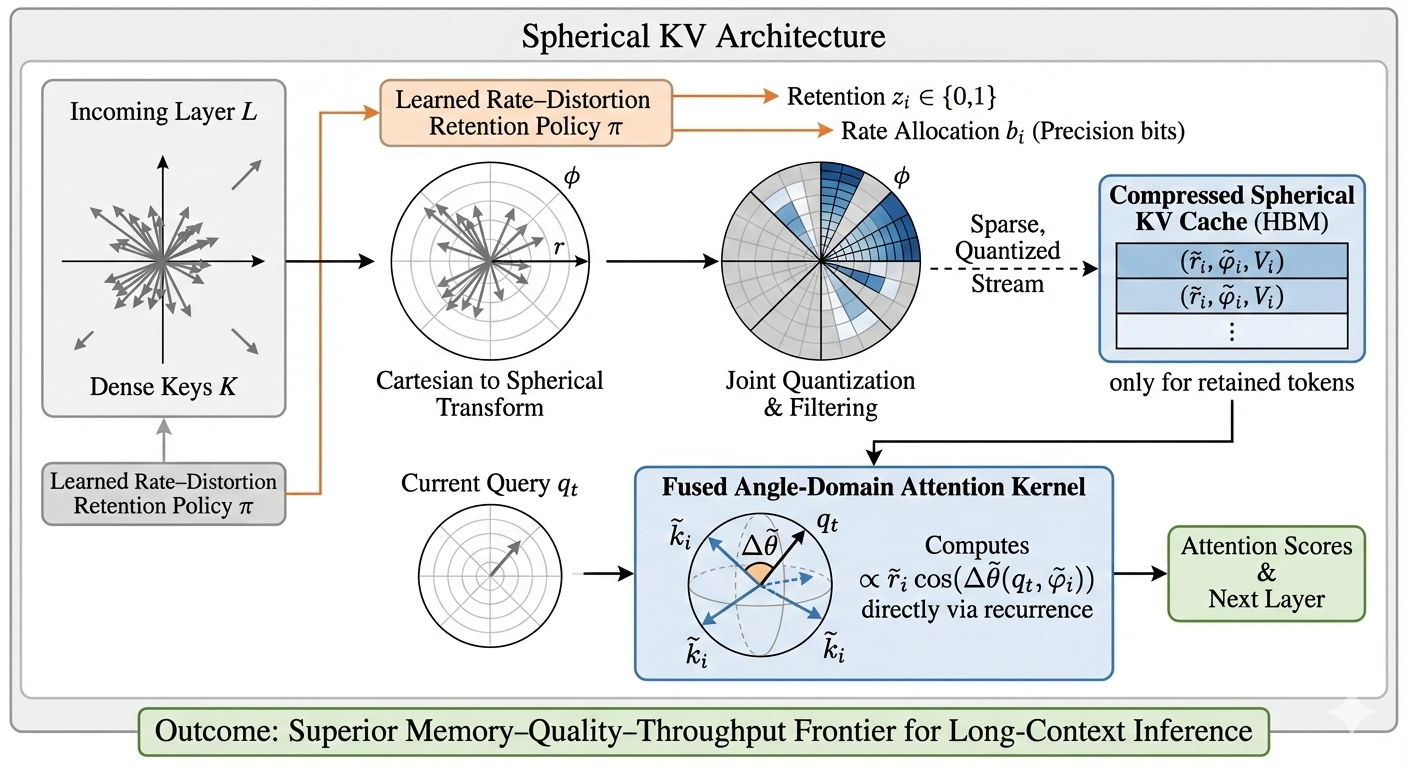}
  \caption{\textbf{Spherical KV Architecture.}
  Dense keys from the incoming layer are mapped to a spherical parameterization (radius $r$ and angles $\phi$).
  A learned rate--distortion retention policy $\pi$ enforces a strict memory budget $B$ by predicting (i) a keep/drop decision $z_i\in\{0,1\}$ and (ii) a per-item precision (rate allocation) $b_i$.
  Joint quantization and filtering produce a sparse, quantized spherical stream that is stored in a compressed KV cache (HBM) for retained tokens only.
  During decoding, a fused angle-domain attention kernel consumes the quantized $(\tilde r_i,\tilde\phi_i)$ directly—avoiding dense Cartesian reconstruction in the critical path—to produce attention scores for the next layer.
  Overall, the pipeline targets a better memory--quality--throughput frontier for long-context inference.}
  \label{fig:spherical_kv_architecture}
\end{figure}
\subsection{Angle-Domain Attention}
\label{app:ada}
\vspace{-2pt}
\paragraph{\textbf{Directional factorization of attention logits.}}
For a head with dimension \(d\), standard attention uses
\[
\ell(\mathbf{q},\mathbf{k})
=\frac{\mathbf{q}^\top\mathbf{k}}{\sqrt{d}}.
\]
Write \(\mathbf{q}=\|\mathbf{q}\|\hat{\mathbf{q}}\) and \(\mathbf{k}=\|\mathbf{k}\|\hat{\mathbf{k}}\), where \(\hat{\mathbf{q}},\hat{\mathbf{k}}\in\mathbb{S}^{d-1}\). Then
\[
\mathbf{q}^\top\mathbf{k}
=\|\mathbf{q}\|\,\|\mathbf{k}\|\,\hat{\mathbf{q}}^\top\hat{\mathbf{k}}
=\|\mathbf{q}\|\,\|\mathbf{k}\|\,\cos\theta,
\qquad \cos\theta \doteq \hat{\mathbf{q}}^\top\hat{\mathbf{k}}.
\]
Hence the logit decomposes as
\[
\ell(\mathbf{q},\mathbf{k})
=\frac{\|\mathbf{q}\|\,\|\mathbf{k}\|}{\sqrt{d}}\,\cos\theta.
\]
This identity is exact and holds regardless of whether the model applies RMSNorm/LayerNorm upstream; in practice norms \(\|\mathbf{q}\|\) and \(\|\mathbf{k}\|\) remain well-behaved due to normalization and training dynamics, while \(\cos\theta\) captures the \emph{directional match} that dominates the ranking of keys.

\begin{figure}[t]
  \centering
  \includegraphics[width=\linewidth]{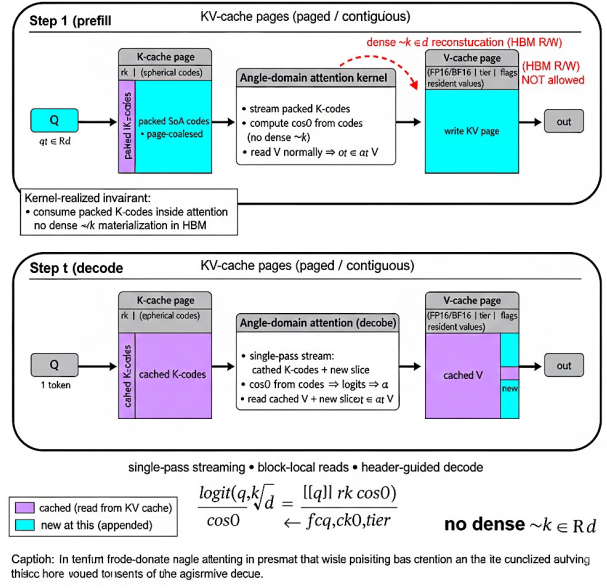}
  \caption{\textbf{Angle-Domain Attention in Practice (Paged KV; kernel-realized, \emph{no reconstruction}).}
  This diagram shows how \textbf{Angle-Domain Attention} fits into a standard paged KV-cache decode loop without paying a hidden densification tax.
  \textbf{Top (prefill):} for each token, we write a \textbf{paged/contiguous K-cache page} that stores \textbf{compact spherical K-codes}—a scalar \(\,r_k\,\) (radius) and \(\,c_k^\theta\) (packed angle codes), plus lightweight \textbf{tier/flags} metadata that makes the layout \textbf{header-guided} and kernel-friendly.
  In parallel, we write a standard \textbf{V-cache page} in FP16/BF16 (resident values), preserving deployable serving assumptions (paged KV, pointer tables, ragged batching).
  \textbf{Bottom (decode):} at each step \(t\), we \textbf{reuse cached pages} (shown as the large cached region) and \textbf{append a thin new slice} for the newly generated token, keeping the write path incremental and page-aligned.
  The \textbf{critical-path difference} is inside the attention kernel: instead of reconstructing a dense \(\tilde{\mathbf{k}}\in\mathbb{R}^d\) (which would reintroduce HBM read/write traffic and extra passes), the kernel \textbf{consumes the packed K-codes directly} and computes the similarity via \(\cos\theta\) \emph{from codes} (hence \( \mathbf{q}\!\cdot\!\mathbf{k}\propto \lVert \mathbf{q}\rVert\, r_k\,\cos\theta\)).
  The resulting logits produce \(\alpha_t=\mathrm{softmax}(\cdot)\), and we then \textbf{read \(V\) normally} to form \(o_t=\alpha_t V\); i.e., \textbf{K is compressed-domain consumed, V remains standard-domain read}.
  \textbf{Kernel-realized invariant (what makes the throughput gain “real”):} the method forbids any dense-key reconstruction stage in the decode hot loop, ensuring a \textbf{single-pass, block-local stream} over packed K-codes and page-coalesced reads, so reductions in KV bytes translate into reduced bandwidth pressure and higher tok/s under paged serving.}
  \label{fig:appndx_angle_domain_attention_practice}
\end{figure}

\paragraph{\textbf{Spherical key coding (what is cached).}}
We represent each key by a \textbf{spherical tuple}
\[
\mathbf{k}\ \longmapsto\ \big(r_k,\ c_k^\theta;\ b,\ \mathrm{flags}\big),
\]
where \(r_k\approx \|\mathbf{k}\|\) is a scalar radius, \(c_k^\theta\) is a \emph{packed angle/direction code} at tier \(b\in\mathcal{B}\), and \(\mathrm{flags}\) include protection/outlier/segment metadata used by the controller and kernel.
The intent is \textbf{not} to store \(\hat{\mathbf{k}}\in\mathbb{R}^d\) explicitly, but to store a compact representation from which the kernel can recover \(\widehat{\cos\theta}\) directly.

\paragraph{\textbf{Query-side coding (computed on the fly).}}
At decode step \(t\), the query is ephemeral and computed for the current token. We form
\[
r_q \doteq \|\mathbf{q}\|,\qquad \hat{\mathbf{q}}=\mathbf{q}/\|\mathbf{q}\|,
\qquad c_q^\theta \leftarrow \mathrm{Encode}(\hat{\mathbf{q}}, b),
\]
so the kernel receives \(r_q\) (or \(\|\mathbf{q}\|\)) and \(c_q^\theta\) for the active tier \(b\). The encoding is chosen to be \textbf{fast}, \textbf{branch-light}, and \textbf{GPU-friendly}.

\paragraph{\textbf{Angle-domain similarity from codes.}}
Define a compressed-domain similarity operator
\[
\widehat{\cos\theta}
\ \doteq\
f\!\left(c_q^\theta,\ c_k^\theta;\ b\right),
\]
where \(f\) is the \emph{angular recurrence} (or code-domain dot-product surrogate) implemented inside the attention kernel. Importantly, \(f\) operates on \textbf{packed code streams} (SoA layout) and is designed to run with \textbf{block-local, coalesced reads} consistent with fused attention kernels \citep{dao2022flashattention,dao2023flashattention2} and paged KV serving \citep{kwon2023efficient}.

\paragraph{\textbf{No-reconstruction invariant (kernel-realized).}}
The central systems constraint is an explicit invariant:
\begin{quote}\vspace{-6pt}
\noindent\textbf{No-reconstruction invariant.} The decode kernel computes logits using \(\widehat{\cos\theta}=f(c_q^\theta,c_k^\theta;b)\) and radius \(r_k\), \emph{without materializing} any dense \(\tilde{\mathbf{k}}\in\mathbb{R}^d\) in HBM.
\vspace{-6pt}\end{quote}
Accordingly, Spherical KV uses the \textbf{angle-domain logit}
\[
\hat{\ell}(\mathbf{q},\mathbf{k})
\ \doteq\
\frac{r_q\,r_k}{\sqrt{d}}\ \widehat{\cos\theta},
\qquad r_q=\|\mathbf{q}\|.
\]
Softmax and value mixing remain standard:
\[
\alpha_i=\frac{\exp(\hat{\ell}_i)}{\sum_j \exp(\hat{\ell}_j)},
\qquad \mathbf{o}=\sum_i \alpha_i \mathbf{v}_i,
\]
where values \(\mathbf{v}_i\) are kept in FP16/BF16 (or another serving-native format) and read normally. Thus, ADA targets \textbf{HBM traffic} specifically in the \(\mathbf{q}^\top\mathbf{k}\) path, where long-context decoding is bandwidth-bound.

\paragraph{\textbf{Logit error propagation (distortion enters as angular error).}}
Let the true cosine be \(\cos\theta\) and the code-derived estimate be \(\widehat{\cos\theta}=\cos\theta+\varepsilon_\theta\).
Assume radii are either stored exactly or with small error \(r_k=\|\mathbf{k}\|+\varepsilon_r\).
Then the logit error is
\[
\Delta \ell
\doteq
\hat{\ell}-\ell
=
\frac{r_q}{\sqrt{d}}
\Big[
(\|\mathbf{k}\|+\varepsilon_r)(\cos\theta+\varepsilon_\theta)
-\|\mathbf{k}\|\cos\theta
\Big],
\]
so
\[
|\Delta \ell|
\le
\frac{r_q}{\sqrt{d}}
\Big(
|\varepsilon_r|\,|\cos\theta|
+\|\mathbf{k}\|\,|\varepsilon_\theta|
+|\varepsilon_r|\,|\varepsilon_\theta|
\Big).
\]
In typical regimes where \(|\varepsilon_r|\,|\varepsilon_\theta|\) is negligible, the dominant term is
\[
|\Delta \ell|
\approx
\frac{r_q\,\|\mathbf{k}\|}{\sqrt{d}}\ |\varepsilon_\theta|.
\]
This clarifies the design target: \textbf{control angular error} \(|\varepsilon_\theta|\) by choosing tier \(b\) and by protecting brittle states (Sec.~\ref{sec:rate_distortion_retention}).

\paragraph{\textbf{From logit drift to attention drift.}}
Softmax is Lipschitz in the logits in standard norms; a simple bound is obtained by noting that for any two logit vectors \(\boldsymbol{\ell},\hat{\boldsymbol{\ell}}\),
\[
\|\mathrm{softmax}(\boldsymbol{\ell})-\mathrm{softmax}(\hat{\boldsymbol{\ell}})\|_1
\le
2\,\|\boldsymbol{\ell}-\hat{\boldsymbol{\ell}}\|_\infty.
\]
Thus, controlling \(\|\Delta \boldsymbol{\ell}\|_\infty\) (via tier selection and protection) directly controls attention-weight drift and downstream output drift.
This is precisely why we treat \textbf{stability} as a first-class constraint rather than an afterthought: ADA supplies an explicit knob (\(b\)) that trades rate for distortion, and RDR chooses that knob under budget.

\paragraph{\textbf{Why ADA is not ``KV quantization as usual''.}}
Many KV compression approaches reduce stored bytes but still \emph{reconstruct dense vectors} (unpack/dequantize) for dot-products, paying a reconstruction tax that reintroduces bandwidth pressure and inhibits kernel fusion. ADA is explicitly designed around the \textbf{kernel consumption path}: \emph{compressed-domain similarity} first, dense values later. This aligns with the philosophy of fused attention kernels \citep{dao2022flashattention,dao2023flashattention2} and serving stacks that emphasize paged locality \citep{kwon2023efficient}.

\subsection{Rate--Distortion Retention}
\label{app:rdr}
\vspace{-2pt}
\paragraph{\textbf{KV control is joint keep/drop + precision.}}
Under a strict memory budget, selecting a subset of tokens is insufficient: one must also choose the \emph{precision tier} of the retained states.
We index cache states by \(i\) (token \(\times\) head; optionally layer), and choose:
\[
z_i\in\{0,1\}\ \ (\text{drop/keep}),\qquad 
b_i\in\mathcal{B}\ \ (\text{tier}).
\]
Let \(\mathrm{cost}(b)\) be bytes-per-state under tier \(b\) (including headers/amortized metadata), and let \(\mathcal{D}_i(b)\) denote the expected distortion incurred by coding state \(i\) at tier \(b\). The controller solves
\[
\min_{\{z_i,b_i\}}
\ \mathbb{E}\big[\mathcal{L}(\mathrm{Decode}_{\text{SphKV}}(\{z_i,b_i\}))\big]
\quad \text{s.t.}\quad 
\sum_i z_i\,\mathrm{cost}(b_i)\le B,
\]
which is a rate--distortion allocation problem in the classical sense \citep{shannon1948}.

\paragraph{\textbf{Lagrangian relaxation and per-state marginal values.}}
Introduce a multiplier \(\lambda\ge 0\) and consider the relaxed objective
\[
\min_{\{z_i,b_i\}}\ 
\mathbb{E}\big[\mathcal{L}(\cdot)\big]
+\lambda\Big(\sum_i z_i\,\mathrm{cost}(b_i)-B\Big).
\]
A useful way to operationalize this is via \emph{marginal improvements}. Define a baseline action (e.g., ``drop'' or ``lowest tier'') and compute, for each state \(i\), the incremental benefit of choosing tier \(b\) instead:
\[
\Delta_i(b)
\ \doteq\
\mathbb{E}\big[\mathcal{L}_i(\text{baseline})\big]
-
\mathbb{E}\big[\mathcal{L}_i(b)\big]
\quad (\text{benefit}).
\]
Then a canonical selection rule is to favor actions with high \emph{benefit per byte}:
\[
\mathrm{vpB}_i(b)
\ \doteq\
\frac{\Delta_i(b)}{\mathrm{cost}(b)-\mathrm{cost}(\text{baseline})}.
\]
In discrete-tier settings, this becomes a multiple-choice knapsack (choose at most one tier for each \(i\)), which is NP-hard in general; however, greedy-by-\(\mathrm{vpB}\) with a tuned \(\lambda\) is a strong practical baseline and yields near-Pareto behavior in serving contexts \citep{martello1990knapsack}.

\paragraph{\textbf{Connecting distortion to ADA's logit error.}}
For Spherical KV, the distortion of a key state \(i\) primarily enters through angular error \(\varepsilon_{\theta,i}(b)\) and (optionally) radius error \(\varepsilon_{r,i}(b)\).
From the ADA derivation,
\[
|\Delta \ell_i(b)|
\lesssim
\frac{r_q\,\|\mathbf{k}_i\|}{\sqrt{d}}\ |\varepsilon_{\theta,i}(b)|
\ +\
\frac{r_q}{\sqrt{d}}|\varepsilon_{r,i}(b)|.
\]
Thus, tiers can be interpreted as choosing a target bound on \(|\varepsilon_{\theta,i}|\), which directly controls logit drift and attention drift. This yields an RD control policy that is \textbf{mechanistically grounded}: it spends bits where logit errors would be amplified (large \(r_q\|\mathbf{k}\|\), high-sensitivity states) and saves bits where the model is robust.

\paragraph{\textbf{Protection constraints (hard constraints, not heuristics).}}
Some states are \emph{brittle} and must be pinned regardless of their immediate \(\mathrm{vpB}\) score:
retrieved evidence spans, safety-critical segments, outliers, or low-margin decisions.
We introduce a protection indicator \(\mathbb{I}_{\text{prot}}(i)\in\{0,1\}\) and enforce
\[
\mathbb{I}_{\text{prot}}(i)=1
\ \Rightarrow\
z_i=1,\quad b_i=b_{\max}.
\]
This is the controller-side mirror of ADA's stability goal: if a state is identified as high-risk, the system does not gamble with compression.

\begin{figure}[H]
  \centering
  \includegraphics[width=\linewidth]{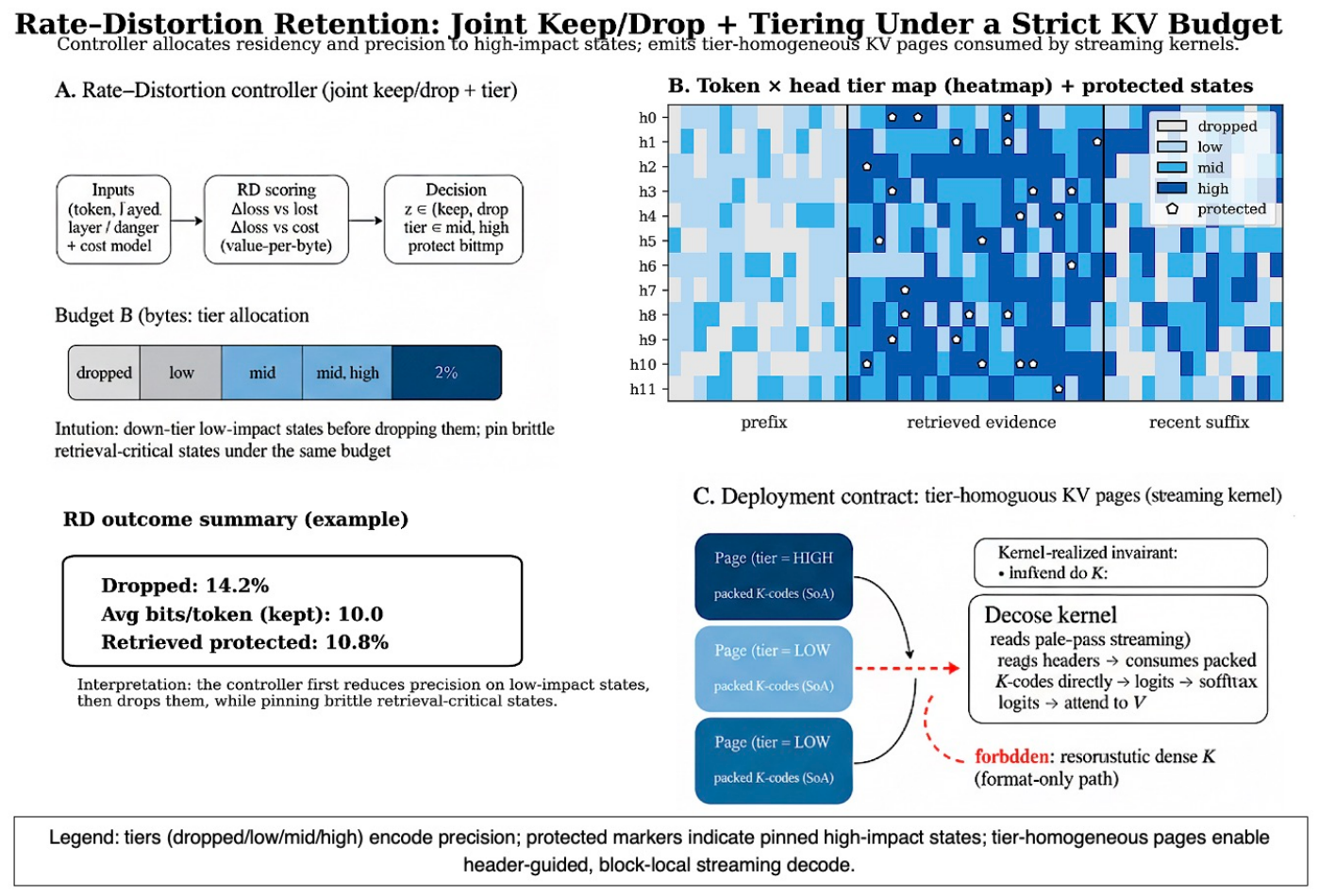}
  \vspace{-4pt}
  \caption{\textbf{Rate--Distortion Retention: joint keep/drop + tiering under a strict KV budget.}
  We treat KV residency as a constrained \emph{rate--distortion} allocation problem: under a fixed byte budget, the controller spends precision on states that are likely to matter later and removes low-utility states first.
  \textbf{(A) RD controller.} Inputs include token/head/layer context plus a cost model; an RD score (e.g., predicted $\Delta$loss vs.\ byte-cost / value-per-byte, augmented with protect/outlier flags) yields a discrete decision: \emph{drop} or \emph{keep} with a \emph{tier} (low/mid/high) and an optional \emph{protect bitmap}.
  The budget bar visualizes how total bytes are partitioned across tiers, emphasizing the policy: \emph{down-tier before dropping} and \emph{pin brittle, retrieval-critical states} when budgets tighten.
  \textbf{(B) Token $\times$ head tier map (mechanism evidence).} Rows are attention heads, columns are tokens; cell color encodes the chosen tier (dropped/low/mid/high), with vertical separators marking prompt segments (prefix / retrieved evidence / recent suffix) and markers denoting \emph{protected} states. The intended signature is segment-aware allocation (retrieved evidence biased toward higher tiers), head heterogeneity (a subset of long-horizon “keeper” heads), and budget-consistent structure (precision reduced on low-impact states before dropping).
  \textbf{(C) Deployment contract: tier-homogeneous KV pages.} The controller emits pages grouped by tier so the decode kernel can do single-pass, header-guided streaming reads: it consumes packed $K$-codes directly (compressed-domain attention) and reads $V$ normally to form outputs.
  The red dashed path highlights the forbidden \emph{format-only} alternative—reconstructing dense $K$—which would reintroduce HBM traffic and erase the kernel-realized benefit.}
  \label{fig:appndx_rd_retention_neurips}
  \vspace{-6pt}
\end{figure}

\paragraph{\textbf{A concrete controller template (deployment-realistic).}}
A minimal controller that is faithful to paged serving proceeds in three stages:
\begin{enumerate}[leftmargin=1.3em,itemsep=2pt,topsep=2pt]
  \item \textbf{Score.} Compute \(\Delta_i(b)\) (or a proxy) for each state and tier, along with \(\mathrm{cost}(b)\). 
        Enforce protection constraints immediately.
  \item \textbf{Allocate tiers before dropping.} Start from a safe tier for all kept states; then down-tier states with lowest \(\mathrm{vpB}\) until the budget is met; only then drop the least valuable states.
  \item \textbf{Write tier-homogeneous pages.} Group states by tier into contiguous pages with fixed headers, enabling specialized kernels and coalesced reads.
\end{enumerate}
This ``down-tier then drop'' structure is important: it avoids brittle discontinuities induced by hard truncation and typically yields smoother quality degradation under tighter budgets.

\subsection{Working Together: One Contract, Two Mechanisms}
\label{app:joint-contract}
\vspace{-2pt}
\paragraph{\textbf{Orthogonality and composition.}}
ADA and RDR act on different axes:
\textbf{ADA} specifies \emph{how} logits are computed cheaply (from codes), while
\textbf{RDR} specifies \emph{which} states exist and \emph{at what tier}.
This separation is what makes the system composable: improvements to ADA kernels do not change the controller objective, and improvements to the controller do not alter the kernel invariant.

\paragraph{\textbf{Serving-native KV layout (paged, tier-homogeneous).}}
We formalize the deployment contract as:
\begin{itemize}[leftmargin=1.2em,itemsep=2pt,topsep=2pt]
  \item \textbf{Paged KV with pointer tables.} KV is stored in pages compatible with ragged batching and reuse across decode steps \citep{kwon2023efficient}.
  \item \textbf{Tier-homogeneous pages.} Each page contains a single tier \(b\), enabling predictable bandwidth and tier-specialized kernels.
  \item \textbf{Header-guided streaming.} A small fixed header stores \((b,\text{len},\text{flags})\), and the kernel streams payloads in one pass.
  \item \textbf{SoA-packed K payload.} Keys are stored as \((r_k, c_k^\theta)\) in SoA form to maximize coalescing; the kernel computes \(\widehat{\cos\theta}\) directly.
  \item \textbf{Standard V payload.} Values remain FP16/BF16 and are read normally to form \(\mathbf{o}=\alpha V\).
\end{itemize}
This contract guarantees that compression translates into real throughput: the hot-path reads are compact and fused (ADA), and the memory budget is enforced by construction (RDR).

\paragraph{\textbf{End-to-end signature (what should improve).}}
In KV-dominated regimes (8K--128K, retrieval-heavy prompts, agent/tool rollouts), the combined mechanism targets a \textbf{frontier shift}:
\emph{lower effective KV bytes/token} (RDR) and \emph{higher tok/s} (ADA) at \emph{bounded iso-quality loss}.
Crucially, the stability story is explicit: ADA provides a distortion knob (tier \(b\)), RDR allocates that knob using value-per-byte plus hard protection, and the serving contract ensures the kernel can realize the promised bandwidth savings \citep{dao2022flashattention,dao2023flashattention2,kwon2023efficient,shannon1948}.

\begin{tcolorbox}[sphalgo={\textbf{Algorithm 1: Spherical KV (End-to-End: Prefill + Decode, Kernel-Realized)}}]
\label{alg:sphkv_end2end_detailed}

\begin{algorithmic}[1]
\Require Model: $L$ layers, $H$ heads, head dim $d_h$, page size $P$ items/page
\Require Prefill tokens $x_{1:T_p}$, decode steps $T_d$, batch size $B$, ragged paging substrate
\Require Tiers $\mathcal{T}=\{t_0,\dots,t_K\}$ with $t_0=\textsf{drop}$ and bitwidths $(b^\theta_t,b^r_t)$
\Require Global budget $\mathcal{B}$ in \emph{bits} (or bytes) for all retained KV at a given layer group
\Require Protect set $\mathcal{P}$ (token/head/layer ids) and tradeoff $\lambda>0$
\Require Quantizers: angle $Q^\theta_t(\cdot)$, radius $Q^r_t(\cdot)$, deterministic decode $D^\theta_t(\cdot)$, $D^r_t(\cdot)$
\Require Angle map $\Phi(\hat{k})\in\mathbb{R}^{d_h-1}$ and \emph{angular-logit routine} $\mathrm{CosFromAngles}(\hat{q},c^\theta,t)$
\Require Meta-bits per item $b^{\mathrm{meta}}_t$ (e.g., flags, offsets); $R(t)=(d_h-1)b^\theta_t+b^r_t+b^{\mathrm{meta}}_t$
\Ensure Tier-homogeneous paged SoA KV store + decode kernel that consumes codes \textbf{without dense reconstruction}

\vspace{4pt}
\State \textbf{Data layout (SoA, tier-homogeneous pages).}
\State For each layer/head $(\ell,h)$, maintain pages $g\in\mathcal{G}_{\ell,h}$ where every page has a single tier $t_g$
\State Page header stores: $(t_g,\ n_g\le P,\ \textsf{scale}_g^r,\ \textsf{scale}_g^\theta,\ \textsf{flags bitmap},\ \textsf{offsets})$
\State Page payload stores SoA streams: $C^\theta_g$ (packed angle codes), $C^r_g$ (packed radius codes), optional $C^v_g$ for values
\State Pointer table $\Pi$ maps $(\ell,h,\textsf{page\_id})\rightarrow$ base addresses of $(C^\theta_g,C^r_g,\textsf{header}_g)$

\vspace{6pt}
\State \textbf{Phase I: Prefill (Write/Allocate).}
\State Run the model on $x_{1:T_p}$ to obtain keys/values $\{k_i^{(\ell,h)},v_i^{(\ell,h)}\}$ for all $i,\ell,h$
\State \textbf{1) Spherical parameterization (per item).}
\For{$\ell\gets1$ to $L$}
  \For{$h\gets1$ to $H$}
    \For{$i\gets1$ to $T_p$}
      \State $r_{i}^{(\ell,h)} \gets \|k_{i}^{(\ell,h)}\|_2$
      \State $\hat{k}_{i}^{(\ell,h)} \gets k_{i}^{(\ell,h)}/(r_{i}^{(\ell,h)}+\epsilon)$
      \State $\phi_{i}^{(\ell,h)} \gets \Phi(\hat{k}_{i}^{(\ell,h)})$ \Comment{$d_h{-}1$ angles}
    \EndFor
  \EndFor
\EndFor

\State \textbf{2) Controller features (replicable scalars).}
\State Define token age $a_i\gets T_p-i$ and segment $\mathrm{seg}(i)\in\{\textsf{prefix},\textsf{retrieved},\textsf{recent}\}$
\State Define segment weights $\omega_{\textsf{prefix}},\omega_{\textsf{retrieved}},\omega_{\textsf{recent}}$ (fixed constants)
\For{$\ell\gets1$ to $L$}
  \For{$h\gets1$ to $H$}
    \State Compute head scalars (examples, fixed recipes):
    \State \hspace{1.0em}$\hat{u}_{\ell,h}\gets$ reuse proxy (e.g., prefill mean attention mass into head $(\ell,h)$)
    \State \hspace{1.0em}$\hat{s}_{\ell,h}\gets$ stability proxy (e.g., margin-danger estimate; see Sec.~\ref{sec:kernel_realized_efficiency})
  \EndFor
\EndFor

\State \textbf{3) Distortion proxy and per-tier score.}
\State Define weights (deterministic):
\State \hspace{1.0em}$w^\theta_{i,\ell,h}\gets \alpha_\theta\,\hat{u}_{\ell,h}\,\omega_{\mathrm{seg}(i)}$, \ \ 
$w^r_{i,\ell,h}\gets \alpha_r\,(1-\hat{s}_{\ell,h})\,\omega_{\mathrm{seg}(i)}$
\For{$\ell\gets1$ to $L$}
  \For{$h\gets1$ to $H$}
    \For{$i\gets1$ to $T_p$}
      \ForAll{$t\in\mathcal{T}$}
        \State $\varepsilon^\theta_{i,\ell,h}(t)\gets$ angle quant error model at tier $t$ (lookup or closed-form)
        \State $\varepsilon^r_{i,\ell,h}(t)\gets$ radius quant error model at tier $t$
        \State $D_{i,\ell,h}(t)\gets w^\theta_{i,\ell,h}\varepsilon^\theta_{i,\ell,h}(t)+w^r_{i,\ell,h}\varepsilon^r_{i,\ell,h}(t)$
        \State $S_{i,\ell,h}(t)\gets -D_{i,\ell,h}(t)-\lambda R(t)$
      \EndFor
      \If{$(i,\ell,h)\in\mathcal{P}$}
        \State Enforce $t\neq t_0$ by restricting $\arg\max$ to $\mathcal{T}\setminus\{t_0\}$
      \EndIf
      \State $t^\star_{i,\ell,h}\gets \arg\max_{t\in\mathcal{T}} S_{i,\ell,h}(t)$
      \State $\nu_{i,\ell,h}\gets \dfrac{D_{i,\ell,h}(t_0)-D_{i,\ell,h}(t^\star_{i,\ell,h})}{R(t^\star_{i,\ell,h})+\epsilon}$ \Comment{value/bit}
    \EndFor
  \EndFor
\EndFor

\State \textbf{4) Budgeted allocation with tier-homogeneous paging.}
\State Initialize $z_{i,\ell,h}\gets 0$, $t_{i,\ell,h}\gets t_0$, remaining budget $\mathcal{B}_{\mathrm{rem}}\gets \mathcal{B}$
\State Sort all items $(i,\ell,h)$ by decreasing $\nu_{i,\ell,h}$ (stable tie-break: lexicographic $(\ell,h,i)$)
\ForAll{items $(i,\ell,h)$ in sorted order}
  \State $t\gets t^\star_{i,\ell,h}$
  \If{$R(t)\le \mathcal{B}_{\mathrm{rem}}$ \textbf{and} feasible to assign into a tier-homogeneous page}
    \State $z_{i,\ell,h}\gets 1$; $t_{i,\ell,h}\gets t$; $\mathcal{B}_{\mathrm{rem}}\gets \mathcal{B}_{\mathrm{rem}}-R(t)$
  \EndIf
\EndFor
\State Pack retained items into pages: group by $(\ell,h,t)$, chunk into size $\le P$ to form pages $g$ with tier $t_g=t$
\State Compute per-page scales (replicable choice):
\State \hspace{1.0em}$\textsf{scale}_g^r\gets \max_{(i,\ell,h)\in g} r_i^{(\ell,h)}$, \ 
$\textsf{scale}_g^\theta\gets 1$ (or per-head calibration constants)

\State \textbf{5) Quantize \& write pages (SoA, pointer table).}
\ForAll{pages $g$}
  \State Write header $(t_g,n_g,\textsf{scale}_g^r,\textsf{flags})$
  \ForAll{$(i,\ell,h)\in g$}
    \State $c^\theta_{i,\ell,h}\gets Q^\theta_{t_g}(\phi_{i}^{(\ell,h)})$ \Comment{pack to $b^\theta_{t_g}$}
    \State $c^r_{i,\ell,h}\gets Q^r_{t_g}\!\Big(\dfrac{r_{i}^{(\ell,h)}}{\textsf{scale}_g^r+\epsilon}\Big)$
    \If{values coded}
      \State $c^v_{i,\ell,h}\gets Q^v_{t_g}(v_i^{(\ell,h)})$ \Comment{or keep dense $v$}
    \EndIf
  \EndFor
  \State Pack $\{c^\theta\}$ and $\{c^r\}$ as SoA bitstreams; update $\Pi$ with base pointers
\EndFor

\vspace{6pt}
\State \textbf{Phase II: Decode (Kernel-realized attention, no reconstruction).}
\State Instrumentation (for replication): place NVTX ranges around \textsf{page\_lookup}, \textsf{kv\_read}, \textsf{angle\_logits}, \textsf{softmax}, \textsf{proj}
\For{$t\gets1$ to $T_d$}
  \For{$\ell\gets1$ to $L$}
    \For{$h\gets1$ to $H$}
      \State \textbf{6) Query.} Compute $q_t^{(\ell,h)}$, $\hat{q}_t^{(\ell,h)}\gets q_t^{(\ell,h)}/(\|q_t^{(\ell,h)}\|+\epsilon)$
      \State \textbf{7) Stream pages.} Iterate pages $g\in\mathcal{G}_{\ell,h}$ via $\Pi$ (paged, ragged batch)
      \ForAll{pages $g\in\mathcal{G}_{\ell,h}$}
        \State Load header $(t_g,n_g,\textsf{scale}_g^r,\textsf{flags})$
        \State Stream SoA code slices for $n_g$ items: $C^\theta_g[1{:}n_g]$, $C^r_g[1{:}n_g]$ (single-pass)
        \For{$j\gets1$ to $n_g$}
          \State $\tilde{r}\gets \textsf{scale}_g^r \cdot D^r_{t_g}(C^r_g[j])$
          \State $c\gets \mathrm{CosFromAngles}(\hat{q}_t^{(\ell,h)}, C^\theta_g[j], t_g)$ \Comment{$c\approx\cos\theta$}
          \State $\ell_{t,(g,j)}^{(\ell,h)} \gets \dfrac{\|q_t^{(\ell,h)}\| \cdot \tilde{r} \cdot c}{\sqrt{d_h}}$
        \EndFor
      \EndFor
      \State \textbf{8) Softmax \& apply.} $\alpha\gets \mathrm{softmax}(\ell)$; accumulate output using corresponding $v$ (dense or coded)
      \State \textbf{Invariant:} Do \emph{not} reconstruct dense $\tilde{k}$ anywhere in this loop.
    \EndFor
  \EndFor

  \State \textbf{9) Append new KV.} For each $(\ell,h)$ encode $k_{T_p+t}^{(\ell,h)}\rightarrow(r,\phi)\rightarrow(c^r,c^\theta)$
  \State \hspace{1.0em}Assign to tier/page using the \emph{same} score recipe (either fixed tier for decode, or gated rule below)

  \State \textbf{10) Optional bounded refresh (if enabled).}
  \If{$t \bmod C = 0$ \textbf{or} a page boundary is crossed}
    \State Recompute controller scalars for \emph{new} items only; do not rewrite old pages (keeps overhead bounded)
    \State Update only future allocations and a small protect bitmap (bounded metadata traffic)
  \EndIf
\EndFor

\vspace{4pt}
\State \textbf{Replication checklist (pin these in the artifact).}
\State (i) $(P,B,T_p,T_d)$, page-table implementation, kernel family, and GPU model;
\State (ii) tier set $\mathcal{T}$ with $(b^\theta_t,b^r_t,b_t^{\mathrm{meta}})$; quantizers $(Q^\theta,Q^r)$ and decoders $(D^\theta,D^r)$;
\State (iii) feature recipes $(\hat{u}_{\ell,h},\hat{s}_{\ell,h},\omega_{\mathrm{seg}},\alpha_\theta,\alpha_r,\lambda)$ and protect set $\mathcal{P}$;
\State (iv) measurement window + NVTX ranges + counters for $b_{\mathrm{HBM}}$ and tok/s (median over trials).
\end{algorithmic}
\end{tcolorbox}

\subsection{Complexity Analysis}
\label{sec:complexity_analysis}

\paragraph{\textbf{Setup and notation.}}
Let $L$ be the number of layers, $H$ the number of attention heads, head dimension $d_h$, prefill length $T_p$, decode length $T_d$, and page capacity $P$ (items per page). Let $N_{\ell,h}$ denote the number of \emph{retained} KV items (after RD keep/drop) for head $(\ell,h)$, with total retained items $N \triangleq \sum_{\ell,h} N_{\ell,h}$. Define the \emph{effective} retained fraction $\rho \triangleq N/(LHT_p)$, and the average retained items per head $\bar{N}\triangleq N/(LH)$. Tiers are $\mathcal{T}=\{t_0,\dots,t_K\}$ with $t_0=\textsf{drop}$ and per-item rate
\[
R(t) \;=\; (d_h-1)\,b^\theta_t \;+\; b^r_t \;+\; b_t^{\mathrm{meta}}\,,
\qquad R(t_0)=0.
\]
We report complexity for (i) \emph{prefill write/allocate} and (ii) \emph{decode read/compute}. Throughout, we emphasize the \textbf{kernel-realized} path: logits are computed from code streams without dense $\tilde{k}$ reconstruction, which is the dominant difference from format-only baselines.

\paragraph{\textbf{Prefill: spherical parameterization and feature extraction.}}
Spherical parameterization computes $r=\|k\|_2$ and $\phi=\Phi(\hat{k})$ per item. This is $\mathcal{O}(LHT_p\cdot d_h)$ arithmetic (norm + normalization + angle map). The controller features are scalar summaries; with a fixed recipe (segment labels, head reuse/stability proxies computed from prefill statistics), feature extraction is
\[
\mathcal{O}(LHT_p)\ \text{(token scalars)} \quad+\quad \mathcal{O}(LH)\ \text{(head scalars)}.
\]
These are one-time costs amortized over $T_d$ decode steps; in long-context settings ($T_d \gg 1$), they are negligible in per-token terms.

\paragraph{\textbf{Prefill allocation: RD tier selection and budgeted packing.}}
For each item $(i,\ell,h)$ we evaluate $K{+}1$ tier scores $S_{i,\ell,h}(t)$ using closed-form distortion proxies. This is
\[
\mathcal{O}(LHT_p\cdot |\mathcal{T}|) \ \text{time}, \qquad \mathcal{O}(LHT_p)\ \text{aux storage (scores or best tier + value/bit)}.
\]
We then sort by value/bit $\nu_{i,\ell,h}$, which costs
\[
\mathcal{O}\big((LHT_p)\log(LHT_p)\big)\,,
\]
followed by a single greedy pass $\mathcal{O}(LHT_p)$ to assign tiers under the global budget. Finally, packing into tier-homogeneous pages is linear in retained items:
\[
\mathcal{O}(N) \ \text{time}, \qquad \mathcal{O}(N/P) \ \text{page headers}.
\]
\emph{Practical note.} The sort is the only superlinear step in prefill; it remains amortized outside the decode hot loop. If desired, it can be replaced by approximate top-$m$ selection (e.g., bucketed $\nu$ or partial selection) without changing the kernel contract.

\paragraph{\textbf{Memory footprint: resident KV bytes/token.}}
Dense KV stores (at minimum) $K$ and $V$ in (say) FP16/BF16, giving per-token bytes roughly
\[
b_{\mathrm{KV}}^{\mathrm{dense}} \;\approx\; 2 \cdot L \cdot H \cdot d_h \cdot 2\ \text{bytes} \;=\; 4LHd_h\ \text{bytes/token}\,,
\]
ignoring metadata and alignment. Spherical KV stores (for keys) radius + angle codes, plus metadata, and optionally codes for values. For keys only, the \emph{effective} per-token resident bytes is
\[
b_{\mathrm{KV}}^{\mathrm{sph}} \;\approx\; \rho \cdot \frac{\mathbb{E}[R(t)]}{8}\ \text{bytes/token}\;+\;\underbrace{\mathcal{O}\!\left(\frac{1}{P}\right)}_{\text{amortized headers}}\!,
\]
where $\mathbb{E}[R(t)]$ is averaged over retained items under the controller’s tier distribution. This makes explicit the two levers:
\emph{retention} ($\rho$) and \emph{rate} (tiered $R(t)$), with page headers amortized by $P$.

\begin{table}[ht!]
\centering
\footnotesize
\setlength{\tabcolsep}{4.2pt}
\renewcommand{\arraystretch}{1.03}
\vspace{-2pt}
\caption{\textbf{Comparison against methods used in our plots (with asymptotics).}
Let $T$ be (retained) context length at decode, $H$ heads, $d_h$ head dim, $b$ effective bits/value (method-dependent), and $k$ the kept set size after eviction/windowing ($k\le T$). We report dominant per-token \emph{decode-step} complexity and memory-traffic scaling (coarse but decision-relevant).}
\label{tab:compare_plot_methods_3col_compact_bigO}
\begin{tabular}{p{0.20\linewidth} p{0.40\linewidth} p{0.34\linewidth}}
\toprule
\textbf{Method} & \textbf{Stored KV \& decode behavior} & \textbf{Dominant per-step scaling (Big-O)} \\
\midrule
\textbf{Dense KV} &
Dense $K,V$ for all past tokens; standard attention computes $QK^\top$, then $\alpha V$. &
\textbf{Compute:} $O(H\,T\,d_h)$.
\textbf{HBM reads:} $O\!\left(H\,T\,(d_h\,b_K + d_v\,b_V)\right)$ bytes $\approx O(H\,T\,d_h)$ (bandwidth-bound). \\
\textbf{StreamingLLM} &
Dense KV but truncate/keep via streaming rule (recency+sinks); dense attention over kept set. &
\textbf{Compute:} $O(H\,k\,d_h)$.
\textbf{HBM reads:} $O\!\left(H\,k\,(d_h\,b_K + d_v\,b_V)\right)$.
(Control: $O(1)$ or $O(k)$ depending on rule.) \\
\textbf{H2O} &
Dense KV with importance-based eviction; dense attention over retained subset. &
\textbf{Compute:} $O(H\,k\,d_h)$.
\textbf{HBM reads:} $O\!\left(H\,k\,(d_h\,b_K + d_v\,b_V)\right)$.
(Control scoring: typically $O(H\,T)$--$O(H\,T\,d_h)$ depending on proxy.) \\
\textbf{KVQuant} &
Quantized $K,V$ (INT8/INT4) with quantized/fused decode \emph{if kernel-native}; else dequantization. &
\textbf{Compute:} $O(H\,T\,d_h)$.
\textbf{HBM reads:} $O\!\left(H\,T\,(d_h\,b_K + d_v\,b_V)\right)$ with smaller $b$.
If dequant/staging: extra $O(H\,T\,d_h)$ movement/ops (hidden tax). \\
\textbf{PolarQuant} &
Polar/spherical quantized format unless paired with code-domain decode kernels. &
If logits require dense reconstruction: \textbf{HBM} includes \emph{reconstructed} dense $K$ so effective reads revert toward $O(H\,T\,d_h)$ plus decode overhead; otherwise similar to KVQuant. \\
\midrule
\textbf{SphKV (Angle-only)} &
Stores spherical K-codes (radius + angle codes); \emph{code-domain} logits; values standard; no RD policy. &
\textbf{Compute:} $O(H\,T\,d_h)$ (still must score $T$ keys),
but \textbf{HBM reads for $K$:} $O\!\left(H\,T\,b_{\text{code}}\right)$ with $b_{\text{code}}\ll d_h b_K$ (no dense-$K$ materialization).
\textbf{HBM reads for $V$:} $O(H\,T\,d_v\,b_V)$. \\
\textbf{SphKV (RD-only)} &
RD keep/drop+tiering under budget, but attention uses dense logits (needs dense $K$). &
\textbf{Compute:} $O(H\,k\,d_h)$.
\textbf{HBM reads:} dense $K,V$ over kept set: $O\!\left(H\,k\,(d_h\,b_K + d_v\,b_V)\right)$.
\textbf{Controller:} typically $O(H\,T\,|\mathcal{T}|)$ scoring + sorting $O(H\,T\log(HT))$ (amortized at prefill / refresh). \\
\textbf{Spherical KV (full)} &
\textbf{Angle-Domain Attention + RD Retention}: tier-homogeneous paged K-codes; kernel consumes codes directly; controller allocates keep/drop+tier. &
\textbf{Compute:} $O(H\,k\,d_h)$ logits over retained $k$ (often $k<T$).
\textbf{HBM reads:}
$K$ codes $O(H\,k\,b_{\text{code}})$ (no dense-$K$),
$V$ reads $O(H\,k\,d_v\,b_V)$.
\textbf{Controller:} $O(H\,T\,|\mathcal{T}| + HT\log(HT))$ at allocation time (prefill / periodic refresh), \emph{not} per decode step. \\
\bottomrule
\end{tabular}
\vspace{-6pt}
\end{table}

\paragraph{\textbf{Decode: paging overhead and streamed HBM traffic.}}
At decode time, each head $(\ell,h)$ streams its retained items once in a block-local pattern through the page table. The number of pages touched is $G_{\ell,h}\approx \lceil N_{\ell,h}/P\rceil$, so pointer-table indirection is
\[
\mathcal{O}\!\left(\sum_{\ell,h} G_{\ell,h}\right) \;=\; \mathcal{O}\!\left(\frac{N}{P}\right),
\]
which is dominated by streaming the payloads. The \textbf{primary systems witness} is HBM bytes/token:
\[
b_{\mathrm{HBM}}^{\mathrm{sph}} \;\approx\; \sum_{\ell,h}\Bigg[\underbrace{G_{\ell,h}\cdot b_{\mathrm{hdr}}}_{\text{headers}} \;+\; \underbrace{N_{\ell,h}\cdot \frac{\mathbb{E}[R(t)]}{8}}_{\text{code payloads}} \Bigg] \;+\; b_{\mathrm{softmax/proj}},
\]
with $b_{\mathrm{softmax/proj}}$ common across methods. Since $b_{\mathrm{hdr}}$ amortizes as $P$ grows, the dominant term is the code payload, which is precisely what the angle-domain kernel consumes without densification.

\paragraph{\textbf{Decode compute: angle-domain logits vs reconstruct-then-dot.}}
Dense attention performs a dot-product $\langle q,k\rangle$ costing $\mathcal{O}(d_h)$ flops per retained item, i.e.,
\[
\mathcal{O}\!\left(\sum_{\ell,h} N_{\ell,h}\cdot d_h\right)\;=\;\mathcal{O}(N d_h)\ \text{per decode token}.
\]
Spherical KV replaces dense dot-products with \emph{angular-logit computation} from compact codes. With a fixed implementation, $\mathrm{CosFromAngles}(\hat{q},c^\theta,t)$ costs $\mathcal{O}(d_h)$ arithmetic in the simplest form (e.g., a recurrence over $d_h{-}1$ angles), but critically \emph{does not require reading $d_h$-dimensional dense keys from HBM}. Thus:
\[
\text{Compute: }\mathcal{O}(N d_h)\ \text{(same order)} \qquad\text{but}\qquad
\text{HBM read: }\mathcal{O}\!\left(N\cdot \mathbb{E}[R(t)]\right)\ \text{bits},
\]
which is the intended regime shift in memory-bound decoding. In contrast, any reconstruct-then-dot baseline incurs additional traffic and staging:
\[
b_{\mathrm{HBM}}^{\mathrm{recon}} \;\approx\; b_{\mathrm{HBM}}^{\mathrm{sph}} \;+\; \underbrace{\Theta(N d_h)}_{\text{dense materialization read/write}},
\]
and typically extra kernel launches, which is why it is labeled \emph{format-only} for throughput claims.

\paragraph{\textbf{Controller overhead during decode.}}
If the controller is \emph{prefill-only} (recommended for a clean systems story), decode overhead is $\mathcal{O}(1)$ per token beyond fixed metadata checks. If a bounded refresh cadence $C$ is enabled, and only the newest $\Delta N$ items are scored/packed every $C$ steps, the amortized overhead is
\[
\mathcal{O}\!\left(\frac{\Delta N\cdot |\mathcal{T}|}{C}\right)\ \text{per token} \quad\text{plus}\quad
\mathcal{O}\!\left(\frac{\Delta N}{C}\log \Delta N\right)\ \text{if sorting is used},
\]
and can be kept strictly below the decode critical path by design (control-plane execution and page-boundary alignment).

\paragraph{\textbf{Takeaway (what matters asymptotically and operationally).}}
Spherical KV changes the long-context bottleneck by (i) reducing \emph{resident} KV via $\rho$ and tiered $R(t)$, and (ii) ensuring the decode kernel consumes compact codes directly, making the dominant term in $b_{\mathrm{HBM}}$ proportional to $\sum_{\ell,h} N_{\ell,h}\,\mathbb{E}[R(t)]$ rather than $\Theta(N d_h)$ dense-key streaming. Prefill-time allocation is superlinear only due to a one-time sort, while decode-time overhead remains linear in retained items with header costs amortized by $P$.

\section{Experimental Setup, Metrics, and Evaluation Protocol}
\label{app:exp-protocol}

\paragraph{\textbf{Goal and philosophy.}}
We evaluate \textbf{\emph{Spherical KV}} precisely at the operating points where inference is \textbf{\emph{memory- and bandwidth-bound}}: long contexts, ragged retrieval prompts, and long-horizon decode rollouts. The central object throughout is a \textbf{\emph{memory--quality--throughput frontier}}: at fixed operating budgets, a method should simultaneously \textbf{(i)} reduce \textbf{\emph{KV-resident footprint}}, \textbf{(ii)} reduce \textbf{\emph{HBM (High Bandwidth Memory) traffic}}, and \textbf{(iii)} preserve \textbf{\emph{task quality and behavioral stability}}. Our protocol is intentionally designed to prevent ``paper speedups'' that disappear under realistic serving assumptions, including \textbf{\emph{paged/ragged KV layouts}}, \textbf{\emph{fused decode kernels}}, and \textbf{\emph{full end-to-end cost accounting}} \citep{kwon2023efficient}. The section is organized around six reviewer-facing questions: \emph{(1) what runtime substrate is assumed, (2) what models and workloads are covered, (3) what quantities are measured, (4) what makes the comparison fair, (5) what baselines and ablations are included, and (6) how are the reported frontier numbers constructed?}

\subsection{Serving Substrate, Kernel Realism, and Runtime Contract}
\label{sec:systems_substrate}

\paragraph{\textbf{Paged/ragged KV substrate (deployment-faithful by construction).}}
All experiments run on a \textbf{\emph{paged, ragged KV cache}} in the style of \textbf{\emph{PagedAttention}} (as popularized in vLLM), where each request’s KV states are stored as a list of fixed-size \textbf{\emph{token blocks}} (pages) allocated from a global pool and indexed through a \textbf{\emph{pointer table}} \citep{kwon2023efficient,kwon2023pagedattention}. This matches real serving more faithfully than a contiguous-cache abstraction: prompts are variable-length, batches are dynamic, preemption is common, and memory reuse is mediated through block-local allocation rather than dense tensor continuity. Concretely, with block size \(P\) tokens and effective request length \(T\), each layer/head KV is stored in
\[
N_{\mathrm{blk}}(T;P)\;=\;\left\lceil \frac{T}{P}\right\rceil
\]
blocks. Token \(i\) maps to block
\[
\mathrm{blk}(i)\;=\;\left\lfloor \frac{i}{P}\right\rfloor,
\qquad
\mathrm{off}(i)\;=\;i \bmod P,
\]
so both \textbf{\emph{block locality}} and \textbf{\emph{ragged addressing}} become first-class systems constraints. Any method that claims decode-time speedups must therefore survive pointer-table resolution, page headers, allocator fragmentation, and non-contiguous addressability in the hot loop.

\paragraph{\textbf{HBM realism and why we measure bytes/token.}}
In long-context decoding, the dominant cost is typically not FLOPs but repeated reads and writes of historical KV states from \textbf{\emph{HBM}}. Each generated token forces attention over a long prefix, making memory traffic the effective systems currency in the long-context regime \citep{dao2022flashattention,dao2023flashattention2}. We therefore report not only throughput but also an explicit traffic metric:
\[
b_{\mathrm{HBM}}
\;\triangleq\;
\frac{\text{device DRAM bytes read + written during decode}}
{\#\text{ generated tokens}}.
\]
We obtain this via profiler counters (e.g., Nsight Compute DRAM read/write metrics) on a fixed decode window, and we always report the exact counter names and tool versions. This quantity is the closest \textbf{\emph{systems-grounded}} surrogate for the KV bottleneck under a paged substrate, because it directly measures whether a representation change reduces the dominant bandwidth term instead of merely moving work elsewhere.

\paragraph{\textbf{Kernel realism rule: no hidden densification in the critical path.}}
We only claim throughput gains if the \textbf{\emph{decode critical path}} consumes the stored representation \textbf{\emph{directly}}, i.e., remains \textbf{\emph{streaming + fused}}:
\begin{itemize}[leftmargin=1.25em,itemsep=2pt,topsep=2pt]
  \item \textbf{Streaming (one-pass reads).} Each decode step should read KV \emph{once} from HBM in a block-local pattern (paged blocks + pointer lookups), without auxiliary staging passes over the same KV.
  \item \textbf{Fused (no dense materialization).} Similarity/logit computation must operate on the stored format itself. Methods that require \textbf{dense reconstruction} (e.g., dequantize \(\rightarrow\) unpack \(\rightarrow\) rebuild dense \(K/V\) vectors) are classified as \textbf{\emph{format-only}} and reported separately, because their additional passes often erase the apparent benefit in memory-bound regimes.
\end{itemize}
Our baseline kernel stack is reported explicitly: FlashAttention-family kernels for dense attention \citep{dao2022flashattention,dao2023flashattention2}, and FlashInfer-style kernels for paged/sparse attention where applicable \citep{ye2025flashinfer}. If a baseline uses a different kernel family, we treat it as a \textbf{\emph{different operating point}} rather than silently merging it into a single comparison bucket.

\paragraph{\textbf{What ``fair'' means at the systems layer.}}
To isolate algorithmic improvements from serving artifacts, we enforce three system-level fairness rules:
\begin{itemize}[leftmargin=1.25em,itemsep=2pt,topsep=2pt]
  \item \textbf{Same substrate.} All methods run on the same paged/ragged KV implementation: same block size \(P\), same allocator, same preemption behavior, and same pointer-table semantics \citep{kwon2023efficient}.
  \item \textbf{Matched KV budgets.} Each method is evaluated at the same \textbf{\emph{effective KV bytes/token}} budget,
  \[
  \mathrm{KVBytes/token}
  \;\triangleq\;
  \frac{\text{total KV bytes resident for the request}}{T},
  \]
  where total bytes include any metadata required for decode (headers, pointer tables, bitwidth/tier tags, protect bitmaps, outlier maps, and fragmentation overhead).
  \item \textbf{Kernel parity where possible.} When two methods store comparable formats (e.g., both low-bit KV), we run them through the same decode kernel family. When that is impossible, we report kernel differences explicitly (compile flags, tensor-core usage, fusion points, and staging behavior).
\end{itemize}

\paragraph{\textbf{Control-plane bound: the controller cannot become the bottleneck.}}
The \textbf{\emph{Rate--Distortion Retention}} controller is restricted to \textbf{\emph{prefill-time}} and must emit a \textbf{\emph{write contract}} that keeps decode-time work constant:
\begin{itemize}[leftmargin=1.25em,itemsep=2pt,topsep=2pt]
  \item \textbf{Prefill-only compute.} The controller runs once per prompt (or once per segment in batched prefill) and emits \textbf{(i)} a retain mask \(z_{i,\ell,h}\), \textbf{(ii)} per-token/head tier assignments, and \textbf{(iii)} a tier-homogeneous page map.
  \item \textbf{Constant-time decode lookup.} During decoding, the runtime performs only pointer-table resolution, page-header reads (tier metadata, offsets), and a single streaming pass over the packed payload. No per-step recomputation of retention or utility scores is allowed.
  \item \textbf{Explicit accounting.} Controller time is included as \textbf{\emph{prefill overhead}} and reported in absolute ms as well as percentage of end-to-end latency.
\end{itemize}
This matters because a budget-aware controller is only useful if it does not merely relocate the bottleneck from HBM traffic to control overhead.

\paragraph{\textbf{Timing protocol (reproducible tok/s and latency).}}
We separate \textbf{\emph{warm-up}} (kernel compilation/JIT, allocator warm-up, cache warming) from \textbf{\emph{measurement}}. Throughput is measured over a fixed decode window of \(N_{\mathrm{gen}}\) generated tokens:
\[
\mathrm{tok/s}
\;\triangleq\;
\frac{N_{\mathrm{gen}}}{t_{\mathrm{decode}}},
\qquad
t_{\mathrm{decode}}
\;=\;
t_{\mathrm{attn}}+t_{\mathrm{mlp}}+t_{\mathrm{sampling}}+t_{\mathrm{runtime}}.
\]
We report: \textbf{(i)} the exact \(N_{\mathrm{gen}}\), \textbf{(ii)} whether prefill is excluded or included, \textbf{(iii)} p50/p95 latency under dynamic batching, and \textbf{(iv)} the random seeds and batching policy used for repeatability. Warm-up and measurement windows are held fixed across methods. This avoids conflating model differences with compilation and runtime transients.

\paragraph{\textbf{Hardware and runtime (audit-ready table).}}
We report all knobs required to reproduce memory and throughput numbers, including the \textbf{\emph{exact}} GPU SKU and software stack, and we pin runtime commits so ``paged-KV behavior'' is not a moving target \citep{kwon2023efficient}.

\begin{table}[t]
\centering
\scriptsize
\setlength{\tabcolsep}{5.2pt}
\renewcommand{\arraystretch}{1.12}
\caption{\textbf{Hardware + runtime environment.} Every reported throughput/memory number is tied to an explicit stack configuration (GPU, kernels, runtime commits, and timing protocol).}
\label{tab:hw_runtime}
\begin{tabular}{l l}
\hline
\textbf{Component} & \textbf{Setting (reported in all runs)} \\
\hline
GPU(s) & Exact SKU (e.g., A100-80GB SXM / H100-80GB SXM), \#GPUs, clocks/power mode \\
HBM & Capacity + vendor peak BW; measured DRAM counters used for \(b_{\mathrm{HBM}}\) \\
Host & CPU model, RAM, PCIe/NVLink topology, OS + kernel version \\
Software & Driver, CUDA, cuBLAS/cuDNN, NCCL (if multi-GPU), compiler versions \\
Runtime & vLLM commit; PagedAttention block size \(P\); allocator/preemption settings \citep{kwon2023efficient} \\
Attention kernels & FlashAttention-family version/flags \citep{dao2022flashattention}; FlashInfer version if used \citep{ye2025flashinfer} \\
Batching & Static vs dynamic; max batch; admission control; preemption on/off \\
Timing & Warm-up steps; \(N_{\mathrm{gen}}\) window; CUDA-event vs wall-clock; p50/p95 protocol \\
\hline
\end{tabular}
\end{table}

\subsection{Models and Workload Regimes}
\label{sec:models_chosen}

\paragraph{\textbf{Models evaluated.}}
We choose three instruct-tuned LLMs spanning \textbf{\emph{scale}} and \textbf{\emph{ecosystem diversity}}, while keeping the study implementable under a kernel-realistic evaluation:
\begin{itemize}[leftmargin=1.5em,itemsep=2pt,topsep=2pt]
  \item \textbf{\emph{Llama-3.1-8B-Instruct}} (8B-class): representative of widely deployed single-GPU serving.
  \item \textbf{\emph{Qwen2.5-14B-Instruct}} (14B-class): higher bandwidth pressure; more heads/layers stress KV traffic.
  \item \textbf{\emph{gpt-oss-20b}} (open-weight reasoning-oriented model): emphasizes strong reasoning + long context, serving as an open-weight reference point in the 20B range.
\end{itemize}

\begin{table}[t]
\centering
\scriptsize
\setlength{\tabcolsep}{4.8pt}
\renewcommand{\arraystretch}{1.12}
\caption{\textbf{Model summary.} We report architectural attributes relevant to KV footprint and decode kernels. For models with grouped-query attention, we report query-head and KV-head counts separately.}
\label{tab:model_summary}
\begin{tabular}{l r r c l r}
\hline
\textbf{Model} & \textbf{Params} & \textbf{Layers} & \textbf{Heads (Q/KV)} & \textbf{Positional} & \textbf{Max ctx} \\
\hline
\href{https://huggingface.co/meta-llama/Llama-3.1-8B-Instruct}{Llama-3.1-8B-Instruct}
& 8B
& 32
& 32 / 8
& RoPE
& 128K \\
\href{https://huggingface.co/Qwen/Qwen2.5-14B-Instruct}{Qwen2.5-14B-Instruct}
& 14.7B
& 48
& 40 / 8
& RoPE
& 131{,}072 \\
\href{https://openai.com/index/introducing-gpt-oss/}{gpt-oss-20b}
& 21B
& 24
& 64 / 8
& RoPE
& 128K \\
\hline
\end{tabular}
\end{table}

\noindent
For each model we additionally report head dimension, grouped/multi-query attention settings, KV layout (per head / grouped), and any RoPE scaling, because KV bandwidth is sensitive to the exact attention geometry and cache layout.

\paragraph{\textbf{Workload design principle.}}
We evaluate \textbf{\emph{three workload regimes}} chosen to satisfy two conditions simultaneously: \textbf{(i)} they expose the KV bottleneck directly (peak KV footprint + HBM traffic), and \textbf{(ii)} they stress distinct failure modes (forgetting, distractor sensitivity, and error amplification in long rollouts). Across all workloads we fix \textbf{(a)} identical tokenization and prompt templates across methods, \textbf{(b)} identical max context \(T\in\{8\text{K},32\text{K},128\text{K}\}\) when supported, and \textbf{(c)} the same paged/ragged substrate so improvements reflect \textbf{\emph{kernel-realized decode behavior}} rather than contiguous-KV artifacts \citep{kwon2023pagedattention}.

\paragraph{\textbf{(W1) Long-context language modeling (pure KV pressure).}}
\begin{itemize}[leftmargin=2em,itemsep=2pt,topsep=2pt]
  \item \textbf{Dataset.} \textbf{\emph{PG-19}} (book-length documents), chosen to measure long-range LM under sustained KV growth \citep{rae2020compressive}.
  \item \textbf{Metric.} Token-level NLL and perplexity with a \textbf{\emph{strided/sliding-window}} causal protocol that preserves causality while reaching long effective contexts \citep{dai2019transformerxl,rae2020compressive}.
  \item \textbf{Why this workload.} PG-19 explicitly stresses long-context dependency, making it a clean probe of how \textbf{\emph{KV footprint}} and \textbf{\emph{HBM bytes/token}} translate into throughput.
  \item \textbf{Systems logs.} decode tok/s, \(\,b_{\mathrm{HBM}}\), and effective \(b_{\mathrm{KV}}\) after retention+tiering.
\end{itemize}

\paragraph{\textbf{(W2) Retrieval-heavy QA with ragged prompts (paging + distractors).}}
\begin{itemize}[leftmargin=2em,itemsep=2pt,topsep=2pt]
  \item \textbf{Benchmark bucket.} \textbf{\emph{LongBench}} as the umbrella long-context benchmark \citep{bai2024longbench}.
  \item \textbf{Concrete datasets.} \textbf{\emph{HotpotQA}} and \textbf{\emph{2WikiMultiHopQA}} as multi-doc / multi-hop QA tasks that naturally induce ragged retrieval blocks \citep{bai2024longbench,ho2020constructing,yang2018hotpotqa}.
  \item \textbf{Prompt serialization.} \textbf{prefix instructions} \(+\) \textbf{retrieved block} (multi-document with separators + lightweight metadata) \(+\) \textbf{recent turns}, yielding \textbf{\emph{ragged paging}} under paged-KV layouts.
  \item \textbf{Distractor protocol.} We append \(K_d\) retrieval distractors (topic-adjacent, answer-irrelevant) and sweep both \textbf{(i)} distractor count and \textbf{(ii)} answer position (early / middle / late) within the retrieved block.
  \item \textbf{Metrics and mechanism evidence.} We report EM/F1 together with tok/s, \(b_{\mathrm{HBM}}\), and \(b_{\mathrm{KV}}\); additionally, we report \textbf{\emph{segment-wise retention+tiering}} (prefix vs.\ retrieved vs.\ recent) as direct controller evidence.
\end{itemize}

\paragraph{\textbf{(W3) Agentic/tool rollouts (long decode + error amplification).}}
\begin{itemize}[leftmargin=2em,itemsep=2pt,topsep=2pt]
  \item \textbf{Datasets.} \textbf{\emph{AgentBench}} and \textbf{\emph{ToolBench}} \citep{liu2023agentbench,qin2023toolbench}.
  \item \textbf{Why this workload.} Long rollouts amplify small logit perturbations into divergent action sequences; compression must therefore demonstrate \textbf{\emph{trajectory stability}}, not only average-case quality.
  \item \textbf{Protocol.} Bounded-horizon episodes (max steps \(S_{\max}\)), fixed tool schema, deterministic tool responses when possible; evaluate \textbf{(a)} success/task completion and \textbf{(b)} seed stability (agreement of tool-call sequences / final answers under identical prompts).
  \item \textbf{Stability metrics.} \textbf{\emph{length drift}} \(\Delta\) tokens versus dense KV, \textbf{\emph{trajectory sensitivity}} across \(N\) seeds, plus tok/s and \(b_{\mathrm{HBM}}\).
\end{itemize}

\paragraph{\textbf{Fairness and serving realism across workloads.}}
All workloads run on the same \textbf{\emph{paged/ragged KV substrate}} \citep{kwon2023pagedattention}. Any baseline that requires \textbf{dense reconstruction} in the decode critical path is evaluated end-to-end with that overhead included, preserving the paper’s notion of a \textbf{\emph{fair systems comparison}}: improvements must survive realistic serving.

\subsection{Metrics and Deployment-True Measurement}
\label{sec:metrics}

\paragraph{\textbf{Principle: evaluation must match the actual bottleneck.}}
We evaluate methods along \textbf{\emph{four coupled axes}}: \textbf{\emph{memory}}, \textbf{\emph{throughput}}, \textbf{\emph{quality}}, and \textbf{\emph{stability}}.
A method counts as an improvement only if it moves the deployment-relevant frontier in the desired direction: \textbf{lower resident KV cost}, \textbf{lower memory traffic}, and/or \textbf{higher decode throughput}, while preserving \textbf{task quality} and avoiding \textbf{behavioral instability}. This is necessary because long-context KV methods can otherwise appear favorable under one axis while silently failing on another; e.g., a method can reduce resident bytes yet increase HBM traffic through reconstruction, or improve average accuracy while destabilizing long-horizon rollouts. Our metric design therefore treats the evaluation problem itself as \textbf{\emph{multi-objective}} rather than reducing it to a single headline number.

\paragraph{\textbf{Memory metrics: what remains resident, and what that really costs.}}
We report memory using both \textbf{\emph{absolute footprint}} and \textbf{\emph{normalized budget}} views.

\noindent\textbf{(i) Peak KV footprint.}
For a request of effective context length \(T\), the total resident KV memory is denoted
\[
M_{\mathrm{KV}}^{\mathrm{resident}}(T).
\]
This quantity includes the full decode-resident state: key/value payloads, compressed codes, per-page headers, pointer tables, protect bitmaps, tier tags, and fragmentation overhead induced by paged allocation. We report the \textbf{\emph{peak}} value across decoding at fixed context lengths \(T\in\{8\mathrm{K},32\mathrm{K},128\mathrm{K}\}\), since peak residency is what constrains batching and maximum supported context under a fixed GPU memory budget.

\noindent\textbf{(ii) Effective resident KV budget (bytes/token).}
Because absolute memory alone is not comparable across requests of different lengths, we normalize residency by the number of active/addressable tokens:
\[
b_{\mathrm{KV}}
\;\triangleq\;
\frac{M_{\mathrm{KV}}^{\mathrm{resident}}}{T_{\mathrm{active}}}.
\]
This is our primary memory budget variable. It should be read as the \textbf{\emph{effective KV cost per active token}} under the actual serving layout. Importantly, \(b_{\mathrm{KV}}\) is \textbf{not} just payload size divided by token count. In a paged substrate, the true resident cost can be decomposed as
\[
M_{\mathrm{KV}}^{\mathrm{resident}}
\;=\;
M_{\mathrm{payload}}
\;+\;
M_{\mathrm{headers}}
\;+\;
M_{\mathrm{ptr}}
\;+\;
M_{\mathrm{tags}}
\;+\;
M_{\mathrm{prot}}
\;+\;
M_{\mathrm{frag}},
\]
where the terms denote payload bytes, page headers, pointer tables, tier/bitwidth tags, protection metadata, and allocator fragmentation, respectively. Therefore,
\[
b_{\mathrm{KV}}
\;=\;
\frac{
M_{\mathrm{payload}}
+
M_{\mathrm{headers}}
+
M_{\mathrm{ptr}}
+
M_{\mathrm{tags}}
+
M_{\mathrm{prot}}
+
M_{\mathrm{frag}}
}{
T_{\mathrm{active}}
}.
\]
This definition is intentionally strict: metadata and paging effects are part of the real cost and cannot be omitted without overstating efficiency \citep{kwon2023efficient}.

\noindent\textbf{(iii) Page-overhead breakdown.}
Whenever useful, we separately report the fraction of resident memory attributable to non-payload terms:
\[
\eta_{\mathrm{meta}}
\;\triangleq\;
\frac{
M_{\mathrm{headers}}
+
M_{\mathrm{ptr}}
+
M_{\mathrm{tags}}
+
M_{\mathrm{prot}}
+
M_{\mathrm{frag}}
}{
M_{\mathrm{KV}}^{\mathrm{resident}}
}.
\]
This is helpful because two methods may achieve the same nominal bitwidth but differ materially in deployment efficiency due to page-local metadata or fragmentation behavior.

\paragraph{\textbf{Throughput metrics: what the runtime actually experiences.}}
We report throughput in a way that distinguishes \textbf{\emph{steady-state decode speed}} from \textbf{\emph{end-to-end latency}}.

\noindent\textbf{(i) Decode throughput (tok/s).}
For a decode window containing \(N_{\mathrm{gen}}\) generated tokens, we define
\[
s
\;\triangleq\;
\frac{N_{\mathrm{gen}}}{t_{\mathrm{decode}}}
\qquad \text{(tokens/s)}.
\]
This is the primary throughput metric in the long-context regime because it isolates the repeated decode loop where KV traffic dominates. We also decompose
\[
t_{\mathrm{decode}}
\;=\;
t_{\mathrm{attn}}
\;+\;
t_{\mathrm{mlp}}
\;+\;
t_{\mathrm{sampling}}
\;+\;
t_{\mathrm{runtime}},
\]
where \(t_{\mathrm{attn}}\) is attention-kernel time, \(t_{\mathrm{mlp}}\) is feed-forward compute, \(t_{\mathrm{sampling}}\) includes logits-to-token sampling overhead, and \(t_{\mathrm{runtime}}\) captures runtime/engine overhead. This decomposition matters because the claimed benefit of Spherical KV should primarily manifest as a reduction in the attention-side memory term, not as an unrelated shift elsewhere.

\noindent\textbf{(ii) End-to-end latency.}
We additionally report full latency measures, including:
- \textbf{TTFT} (time to first token),
- per-token latency after decode warm-up,
- and p50/p95 latency under batching.
Where relevant, the complete decomposition is
\[
t_{\mathrm{e2e}}
\;=\;
t_{\mathrm{prefill}}
\;+\;
t_{\mathrm{allocate/write}}
\;+\;
t_{\mathrm{decode}}.
\]
This ensures that any controller cost, page construction overhead, or reconstruction/offload cost is included in the real runtime story rather than hidden outside the decode loop.

\noindent\textbf{(iii) HBM bytes/token (primary systems KPI).}
The decisive systems metric is
\[
b_{\mathrm{HBM}}
\;\triangleq\;
\frac{\texttt{HBM\_read}+\texttt{HBM\_write}}
     {\#\text{ decode tokens}},
\]
where \(\texttt{HBM\_read}\) and \(\texttt{HBM\_write}\) are measured from GPU profiling counters over the decode window. This quantity is critical because long-context decoding is typically \textbf{\emph{memory-bandwidth bound}} rather than compute-bound \citep{dao2022flashattention,dao2023flashattention2}. A method that reduces stored bytes but still requires extra decode-time staging, dequantization, or dense reconstruction may lower \(b_{\mathrm{KV}}\) while leaving \(b_{\mathrm{HBM}}\) nearly unchanged, or even increasing it. Thus, in this paper, \(b_{\mathrm{HBM}}\) is the primary test of whether a method reduces the \textbf{\emph{true runtime bottleneck}}.

\noindent\textbf{(iv) Throughput-per-budget views.}
For selected summaries, we also examine normalized efficiency quantities such as
\[
\frac{s}{b_{\mathrm{KV}}}
\qquad\text{and}\qquad
\frac{s}{b_{\mathrm{HBM}}},
\]
which provide intuition about whether a method is translating resident-memory savings or traffic savings into realized decode throughput. These are secondary summaries rather than primary metrics.

\paragraph{\textbf{Quality metrics: preserving utility under compression.}}
Quality is workload-dependent, but the logic is unified: a systems gain only matters if the resulting model remains useful.

\noindent\textbf{(i) Task metrics.}
We report:
\begin{itemize}[leftmargin=1.5em,itemsep=2pt,topsep=2pt]
  \item \textbf{\emph{Long-context LM}}: token-level NLL and perplexity.
  \item \textbf{\emph{Retrieval-heavy QA}}: EM/F1 or task-appropriate exactness metrics.
  \item \textbf{\emph{Agentic/tool rollouts}}: success / completion rate and trajectory-level correctness.
\end{itemize}

\noindent\textbf{(ii) Long-context degradation curves.}
For each method and workload, we report quality as a function of context length and budget:
\[
Q = Q(T,\mathcal{B}),
\]
where \(T\) is context length and \(\mathcal{B}\) denotes the effective budget setting. This makes degradation with increasing KV pressure explicit, rather than reporting only one operating point.

\noindent\textbf{(iii) Budget-normalized quality.}
Because different methods operate at different effective budgets, we also compare quality at matched \(b_{\mathrm{KV}}\). Formally, if \(Q_m(\mathcal{B})\) is the quality of method \(m\) at budget \(\mathcal{B}\), then budget-normalized comparisons evaluate \(Q_m(\mathcal{B})\) across methods under shared \(\mathcal{B}\). This prevents a method from claiming higher accuracy simply because it was allocated a larger effective memory budget.

\noindent\textbf{(iv) Depth-conditioned quality.}
For long-context workloads, we additionally report quality conditioned on answer or evidence depth, e.g.,
\[
Q_{\mathrm{depth}}(d),
\]
where \(d\) indexes relative position (early / middle / late). This is especially important because long-context failures often concentrate in the middle or late parts of the prompt, even when average quality appears stable \citep{liu2023lost}.

\paragraph{\textbf{Stability metrics: beyond average accuracy.}}
A distinctive concern for KV approximation is that it can perturb the generation trajectory even when average task metrics change only slightly. We therefore treat \textbf{\emph{stability}} as a separate measured axis rather than an informal remark.

\noindent\textbf{(i) Seed sensitivity.}
For each operating point, we run \(N\) seeds and compute the variance of quality and throughput:
\[
\mathrm{Var}_{\mathrm{seed}}[Q],
\qquad
\mathrm{Var}_{\mathrm{seed}}[s].
\]
A method that appears strong only under one seed but unstable across seeds is not considered robust.

\noindent\textbf{(ii) Length drift.}
For long-horizon generation, we track deviation in output length relative to dense KV:
\[
\Delta_{\mathrm{tok}}
\;\triangleq\;
\#\text{tokens}_{\mathrm{method}}
-
\#\text{tokens}_{\mathrm{dense}}.
\]
This detects premature EOS, runaway continuation, or altered stopping behavior.

\noindent\textbf{(iii) Trajectory disagreement.}
For tool-augmented or agentic workloads, we report the fraction of prompts whose action/tool-call trajectory differs across seeds or relative to dense KV:
\[
\mathrm{Disagree}
\;\triangleq\;
\frac{1}{N_{\mathrm{ep}}}
\sum_{e=1}^{N_{\mathrm{ep}}}
\mathbb{1}\!\left[\tau_e^{(m)} \neq \tau_e^{(\mathrm{dense})}\right].
\]
Here \(\tau_e\) denotes the discrete trajectory for episode \(e\). This quantity is more sensitive than final success alone because small internal perturbations can cascade into very different action sequences.

\noindent\textbf{(iv) Failure-mode accounting.}
We explicitly categorize failures such as:
- looping,
- derailment,
- early refusal,
- malformed tool calls,
- and structured-output corruption.
These are qualitative categories, but they are attached to concrete exemplars and counts rather than left anecdotal.

\noindent\textbf{(v) Logit-drift diagnostics.}
To connect stability back to mechanism, we measure the distribution of attention-logit perturbations induced by approximation:
\[
\Delta \ell_{t,i}
\;\triangleq\;
\ell_{t,i}^{\mathrm{approx}}
-
\ell_{t,i}^{\mathrm{dense}}.
\]
We summarize this distribution using statistics such as mean absolute drift, quantiles, and tail probability
\[
\Pr\!\left(|\Delta \ell_{t,i}| > \epsilon\right),
\]
over held-out prompts. This is useful because many practical failures are caused not by average drift, but by rare large perturbations on fragile positions.

\paragraph{\textbf{How the four axes interact.}}
These metrics are designed to be interpreted jointly. In particular:
\begin{itemize}[leftmargin=1.5em,itemsep=2pt,topsep=2pt]
  \item \textbf{Memory} tells us what remains resident.
  \item \textbf{HBM traffic} tells us what still moves through the real bottleneck.
  \item \textbf{Throughput} tells us whether the traffic reduction becomes a real runtime gain.
  \item \textbf{Quality and stability} tell us whether that gain is usable.
\end{itemize}
A method is therefore strong only if it moves the frontier in the right direction under all four axes, rather than optimizing one while silently degrading another.

\paragraph{\textbf{What this subsection contributes to the evaluation logic.}}
This metric suite makes the experimental claims \textbf{\emph{deployment-true}}. The paper does not ask only whether a representation is smaller, but whether it is:
\textbf{(i)} cheaper to keep resident,
\textbf{(ii)} cheaper to stream through HBM,
\textbf{(iii)} faster in the actual decode loop,
and \textbf{(iv)} behaviorally safe enough to use.
That is the correct standard for long-context KV methods operating under realistic serving constraints.

\subsection{Fairness Rules and Operating-Point Definitions}
\label{sec:protocol}

\paragraph{\textbf{Principle: fixed budgets, matched operating points.}}
For every \((\text{model},\text{workload})\), we compare methods only at \textbf{\emph{matched budget}} \(b_{\mathrm{KV}}\) and under the shared substrate. Because different methods may yield different quality at the same budget, we report two fairness-preserving projections \citep{lin2024kvquant,zhang2023h2o}.

\paragraph{\textbf{Iso-quality throughput gain.}}
Let \(q^\star\) be a target quality level. Define the best throughput achievable by method \(m\) while meeting \(q^\star\):
\[
s_m^{\mathrm{isoQ}}(q^\star)
\;\triangleq\;
\max_{\mathcal{B}}
\Big\{
s_m(\mathcal{B}) : q_m(\mathcal{B}) \ge q^\star
\Big\}.
\]
We then report the relative gain against dense-KV serving:
\[
G_m^{\mathrm{isoQ}}(q^\star)
\;\triangleq\;
\frac{s_m^{\mathrm{isoQ}}(q^\star)}
{s_{\text{dense}}^{\mathrm{isoQ}}(q^\star)}.
\]
This prevents a method from claiming speed by silently degrading quality.

\paragraph{\textbf{Iso-throughput memory reduction.}}
Let \(s^\star\) be a target decode throughput. Define the minimum KV budget required by method \(m\) to achieve \(s^\star\) while preserving quality above threshold \(q_{\min}\):
\[
\mathcal{B}_m^{\mathrm{isoS}}(s^\star)
\;\triangleq\;
\min_{\mathcal{B}}
\Big\{
\mathcal{B}: s_m(\mathcal{B}) \ge s^\star \ \wedge\ q_m(\mathcal{B}) \ge q_{\min}
\Big\}.
\]
We report memory reduction versus dense KV:
\[
R_m^{\mathrm{isoS}}(s^\star)
\;\triangleq\;
\frac{\mathcal{B}_{\text{dense}}^{\mathrm{isoS}}(s^\star)}
{\mathcal{B}_m^{\mathrm{isoS}}(s^\star)}.
\]
This prevents a method from claiming a ``memory win'' only at an unfairly slower operating point.

\paragraph{\textbf{Warm-up and measurement windows.}}
We separate warm-up and measurement to avoid compilation and cache-cold artifacts \citep{kwon2023efficient}. For each run:
\begin{itemize}[leftmargin=1.1em,itemsep=2pt,topsep=2pt]
  \item \textbf{Warm-up:} generate \(N_{\mathrm{warm}}\) tokens (excluded from metrics).
  \item \textbf{Measurement:} generate \(N_{\mathrm{meas}}\) tokens and compute tok/s, \(b_{\mathrm{HBM}}\), and quality metrics on the same window.
\end{itemize}
We report \((N_{\mathrm{warm}},N_{\mathrm{meas}})\) and use identical values across methods.

\paragraph{\textbf{Identical decoding settings.}}
We fix decoding hyperparameters (temperature, top-\(p\), max new tokens), prompt templates, and tokenization across methods. Any deviation is treated as a fairness violation and is explicitly reported. For agentic tasks, we fix the tool schema and tool responses whenever possible to isolate KV effects \citep{liu2023agentbench,qin2023toolbench}.

\paragraph{\textbf{Full cost accounting (no hidden reconstruction).}}
We report the complete end-to-end decomposition:
\[
t_{\mathrm{e2e}}
\;=\;
t_{\mathrm{prefill}}
\;+\;
t_{\mathrm{allocate/write}}
\;+\;
t_{\mathrm{decode}}
\quad
(\text{and } b_{\mathrm{HBM}} \text{ over the same decode window}).
\]
A baseline that requires dense reconstruction (dequantize/unpack/materialize) must pay that time in \(t_{\mathrm{decode}}\), and any offload/transfer must be included in both time and traffic reporting \citep{lin2024kvquant,kwon2023efficient}. This is essential because many KV methods shift the bottleneck rather than removing it.

\paragraph{\textbf{Seed control and stability reporting.}}
For each operating point we run \(N\) random seeds and report mean \(\pm\) std for quality and tok/s. For long-horizon rollouts we additionally report a \textbf{\emph{trajectory disagreement rate}}: the fraction of episodes whose tool-call sequences differ across seeds, because compounding error is the defining failure mode in agentic settings \citep{liu2023lost,liu2023agentbench}.

\subsection{Baselines and Ablation Design}
\label{subsec:baselines}

\paragraph{\textbf{Baseline design principle.}}
We compare against representative method families and enforce the same \textbf{\emph{kernel realism rule}} used for our method: throughput claims must be measured end-to-end on a paged cache and include all critical-path costs \citep{kwon2023efficient,dao2022flashattention}.

\paragraph{\textbf{Retention / eviction baselines.}}
\begin{itemize}[leftmargin=1.1em,itemsep=2pt,topsep=2pt]
  \item \textbf{\emph{Sliding window + attention sinks.}} A strong streaming baseline that preserves a small set of sink tokens while dropping older context \citep{xiao2023streamingllm}.
  \item \textbf{\emph{Heavy-hitter eviction (H\(_2\)O).}} Retains tokens predicted to dominate future attention, providing a competitive learned/heuristic retention policy \citep{zhang2023h2o}.
\end{itemize}

\paragraph{\textbf{KV quantization baselines.}}
\begin{itemize}[leftmargin=1.1em,itemsep=2pt,topsep=2pt]
  \item \textbf{\emph{Uniform low-bit KV (2/4-bit).}} A clean baseline implemented on the same paged substrate and reported in bytes/token and HBM bytes/token.
  \item \textbf{\emph{KVQuant.}} A state-of-the-art KV-cache quantization baseline targeting extreme contexts; we include its end-to-end packing/unpacking, staging, and kernel constraints explicitly \citep{lin2024kvquant}.
\end{itemize}

\paragraph{\textbf{Closest conceptual neighbor (direction/magnitude coding).}}
We include a \textbf{\emph{polar/angle-encoding baseline}} that parameterizes direction and magnitude but requires partial/full reconstruction before dot-products, specifically to isolate the value of \textbf{\emph{angle-domain compute without reconstruction}} (A1) \citep{lin2024kvquant,dao2022flashattention}.

\paragraph{\textbf{Ablation families.}}
We include ablations that correspond directly to the paper’s two ``hero'' ideas:
\begin{itemize}[leftmargin=1.1em,itemsep=2pt,topsep=2pt]
  \item \textbf{Angle-domain compute vs.\ reconstruction.} Same spherical storage, but \textbf{(a)} fused angle-domain logits versus \textbf{(b)} reconstruct-then-dot.
  \item \textbf{Retention-only vs.\ quant-only vs.\ joint RD policy.} \textbf{(a)} keep/drop only, \textbf{(b)} fixed low-bit for all kept tokens, \textbf{(c)} joint keep/drop + per-head (or per-token) rate tiers under the same budget.
  \item \textbf{Per-head heterogeneity.} Shared tier globally versus head-wise tiers.
  \item \textbf{Budget sweeps.} Multiple KV budgets to produce a \textbf{\emph{frontier curve}}, not a single operating point.
  \item \textbf{Segment sensitivity and stability gating.} Prefix/retrieved/recent sensitivity and gate-vs.-nogate analyses to expose mechanism evidence and failure predictability.
\end{itemize}

\subsection{Frontier Construction and Reporting Protocol}
\label{subsec:frontier}

\paragraph{\textbf{Why the frontier is the correct comparison object.}}
In long-context inference, a single-point comparison is often misleading. A method can appear faster only because it is evaluated at a looser quality target, or appear smaller only because metadata, paging overhead, or decode-time reconstruction costs are excluded. For this reason, we treat the experimental object as a \textbf{\emph{frontier}} rather than a single operating point. The frontier jointly characterizes the trade among \textbf{\emph{resident KV memory}}, \textbf{\emph{HBM traffic}}, \textbf{\emph{decode throughput}}, and \textbf{\emph{quality/stability}} under a fixed serving substrate. This is the natural comparison object for long-context KV methods because the relevant question is not ``which method is best at one arbitrary knob setting,'' but rather \textbf{\emph{which method dominates the feasible operating region under shared deployment assumptions}} \citep{kwon2023efficient,lin2024kvquant,zhang2023h2o,xiao2023streamingllm}. 

\paragraph{\textbf{Budget sweeps: what is varied for each method.}}
For each method \(m\), model \(\mathcal{M}\), workload \(\mathcal{W}\), and context regime \(T\), we evaluate a finite set of operating budgets
\[
\mathcal{B}\in\{b_1,\dots,b_K\},
\]
where \(\mathcal{B}\) denotes the method’s \textbf{\emph{natural budget-control knob}}. The exact meaning of \(\mathcal{B}\) depends on the method family:
\begin{itemize}[leftmargin=1.25em,itemsep=2pt,topsep=2pt]
  \item for \textbf{\emph{Spherical KV}}, \(\mathcal{B}\) spans both a \textbf{retention level} and a \textbf{tier/rate allocation schedule}, since the method jointly controls \emph{which} states are retained and \emph{at what precision};
  \item for \textbf{\emph{retention/eviction baselines}}, \(\mathcal{B}\) spans quantities such as window size, retain ratio, or eviction aggressiveness \citep{xiao2023streamingllm,zhang2023h2o};
  \item for \textbf{\emph{quantization baselines}}, \(\mathcal{B}\) spans bitwidth, outlier handling, or mixed-precision settings \citep{lin2024kvquant};
  \item for \textbf{\emph{hybrid or format-only baselines}}, \(\mathcal{B}\) may also include reconstruction/staging modes, which are charged explicitly if they affect the decode path.
\end{itemize}
Each evaluated budget setting yields a measured tuple
\[
x_m(\mathcal{B})
\;\triangleq\;
\Big(
b_{\mathrm{KV},m}(\mathcal{B}),
\ b_{\mathrm{HBM},m}(\mathcal{B}),
\ s_m(\mathcal{B}),
\ q_m(\mathcal{B}),
\ \sigma_m(\mathcal{B})
\Big),
\]
where:
\begin{itemize}[leftmargin=1.25em,itemsep=2pt,topsep=2pt]
  \item \(b_{\mathrm{KV},m}\) is the \textbf{effective resident KV budget},
  \item \(b_{\mathrm{HBM},m}\) is the \textbf{HBM bytes per generated token},
  \item \(s_m\) is \textbf{steady-state decode throughput},
  \item \(q_m\) is the \textbf{task-quality score},
  \item \(\sigma_m\) denotes \textbf{stability statistics} (e.g., seed variance, length drift, trajectory disagreement), when relevant.
\end{itemize}
Thus, each point on a plotted frontier is the projection of a higher-dimensional measured operating point, not merely a pair of numbers.

\paragraph{\textbf{Deployment-true operating budget.}}
The primary memory budget variable is the \textbf{\emph{effective resident KV cost per active token}}
\[
b_{\mathrm{KV}}
\;\triangleq\;
\frac{M_{\mathrm{KV}}^{\mathrm{resident}}}{T_{\mathrm{active}}},
\]
where \(T_{\mathrm{active}}\) is the number of addressable tokens in the current request and
\[
M_{\mathrm{KV}}^{\mathrm{resident}}
=
M_{\mathrm{payload}}
+
M_{\mathrm{headers}}
+
M_{\mathrm{ptr}}
+
M_{\mathrm{tags}}
+
M_{\mathrm{prot}}
+
M_{\mathrm{frag}}.
\]
Here \(M_{\mathrm{payload}}\) denotes KV payload bytes, \(M_{\mathrm{headers}}\) page headers, \(M_{\mathrm{ptr}}\) pointer tables, \(M_{\mathrm{tags}}\) tier/bitwidth tags, \(M_{\mathrm{prot}}\) protection metadata, and \(M_{\mathrm{frag}}\) allocator fragmentation induced by paged/ragged serving. This definition is intentionally strict: it matches the true resident footprint seen by the runtime and avoids the common mistake of comparing compressed payloads while ignoring decode-required metadata \citep{kwon2023efficient}. In other words, \textbf{\emph{paging and metadata are part of the algorithmic cost model}}.

\paragraph{\textbf{HBM traffic as the systems-side companion metric.}}
Because long-context decoding is governed by repeated KV streaming from HBM, we measure the decode-side traffic
\[
b_{\mathrm{HBM}}
\;\triangleq\;
\frac{\mathrm{HBM}_{\mathrm{read}}+\mathrm{HBM}_{\mathrm{write}}}
{\#\text{ decode tokens}},
\]
over the same decode window used for throughput measurement. This metric is essential because reducing \(b_{\mathrm{KV}}\) does not guarantee reducing \(b_{\mathrm{HBM}}\): a method may compress KV at rest yet still incur a decode-time \textbf{densification tax} through dequantization, unpacking, or dense reconstruction. Therefore, in this paper, a memory-side gain is considered \textbf{\emph{kernel-realized}} only if it is accompanied by a corresponding reduction in \(b_{\mathrm{HBM}}\) under the shared substrate \citep{dao2022flashattention,dao2023flashattention2,kwon2023efficient}.

\paragraph{\textbf{Feasible sets under quality and fairness constraints.}}
Not every measured operating point is eligible for headline comparison. We first define a \textbf{\emph{quality-feasible set}}. Let \(q^\star\) denote the target quality threshold used for iso-quality comparison, or equivalently let \(Q^{\star}_{\mathrm{dense}}\) denote the dense-KV reference score under the same decode settings. Then the feasible set for method \(m\) is
\[
\mathcal{F}_m(q^\star)
\;\triangleq\;
\Big\{
\mathcal{B}\in\{b_1,\dots,b_K\}
\;:\;
q_m(\mathcal{B}) \ge q^\star
\Big\}.
\]
In the dense-relative form used in the main paper, one may write
\[
\mathcal{F}_m(\Delta)
\;\triangleq\;
\Big\{
\mathcal{B}
\;:\;
q_m(\mathcal{B}) \ge Q^{\star}_{\mathrm{dense}}-\Delta
\Big\},
\]
where \(\Delta\) is the allowed quality tolerance. Only operating points in \(\mathcal{F}_m\) are allowed to contribute to the \textbf{iso-quality frontier}. This prevents a method from claiming speed by silently sacrificing task utility.

In addition, all points are required to satisfy the \textbf{\emph{fairness contract}} of the appendix: shared paged substrate, shared decode settings, explicit end-to-end cost accounting, and no hidden exclusion of reconstruction/offload cost. Thus the true feasible set is the intersection of \textbf{quality-feasible} and \textbf{fairness-valid} points.

\paragraph{\textbf{Frontier extraction as a Pareto envelope.}}
Given the feasible set, the frontier is defined as the set of \textbf{\emph{non-dominated}} operating points under the chosen projection. For the main memory--throughput plots, the relevant projection is \((b_{\mathrm{KV}}, s)\), with lower \(b_{\mathrm{KV}}\) and higher \(s\) preferred. Formally, for method \(m\), define the feasible projected set
\[
\Pi_m
\;\triangleq\;
\Big\{
\big(b_{\mathrm{KV},m}(\mathcal{B}),\, s_m(\mathcal{B})\big)
\;:\;
\mathcal{B}\in\mathcal{F}_m
\Big\}.
\]
A point \(p=(b,s)\in\Pi_m\) is \textbf{dominated} if there exists another point \(p'=(b',s')\in\Pi_m\) such that
\[
b' \le b,\qquad s' \ge s,
\]
with at least one inequality strict. The \textbf{Pareto envelope} (or upper envelope) is then
\[
\mathrm{Env}_m
\;\triangleq\;
\Big\{
p\in\Pi_m:\ p\ \text{is not dominated in}\ \Pi_m
\Big\}.
\]
This envelope is the mathematically correct object reported in the frontier plots: it discards inferior operating points and retains only those that define the achievable trade-off surface under shared conditions.

\paragraph{\textbf{Iso-quality and iso-throughput summaries extracted from the frontier.}}
From the frontier we derive two reviewer-auditable summary statistics.

\noindent\textbf{(i) Iso-quality throughput gain.}
For a target quality \(q^\star\), define
\[
s_m^{\mathrm{isoQ}}(q^\star)
\;\triangleq\;
\max_{\mathcal{B}\in \mathcal{F}_m(q^\star)}
s_m(\mathcal{B}).
\]
The relative gain against dense KV is
\[
G_m^{\mathrm{isoQ}}(q^\star)
\;\triangleq\;
\frac{s_m^{\mathrm{isoQ}}(q^\star)}
{s_{\mathrm{dense}}^{\mathrm{isoQ}}(q^\star)}.
\]
This quantity answers: \emph{at fixed quality, how much decode speed can the method deliver?}

\noindent\textbf{(ii) Iso-throughput memory reduction.}
For a target throughput \(s^\star\), define the minimum feasible budget
\[
\mathcal{B}_m^{\mathrm{isoS}}(s^\star)
\;\triangleq\;
\min_{\mathcal{B}}
\Big\{
b_{\mathrm{KV},m}(\mathcal{B})
\;:\;
s_m(\mathcal{B}) \ge s^\star
\;\wedge\;
q_m(\mathcal{B}) \ge q_{\min}
\Big\}.
\]
The corresponding reduction relative to dense KV is
\[
R_m^{\mathrm{isoS}}(s^\star)
\;\triangleq\;
\frac{
\mathcal{B}_{\mathrm{dense}}^{\mathrm{isoS}}(s^\star)
}{
\mathcal{B}_{m}^{\mathrm{isoS}}(s^\star)
}.
\]
This quantity answers: \emph{at fixed throughput, how much resident KV memory can the method save without violating a quality floor?}

Together, \(G_m^{\mathrm{isoQ}}\) and \(R_m^{\mathrm{isoS}}\) summarize the two deployment-relevant ways a method can improve the frontier: \textbf{\emph{more speed at fixed quality}} or \textbf{\emph{less memory at fixed speed}}.

\paragraph{\textbf{Why ablations belong on the same frontier.}}
We include ablation variants such as \textbf{Angle-only}, \textbf{RD-only}, \textbf{retention-only}, \textbf{quant-only}, and other mechanism-isolating variants on the same frontier plots because the frontier itself is the correct object for mechanism attribution. A component is meaningful only if its addition expands the feasible non-dominated region; a knob that changes some internal quantity without moving the envelope does not contribute to the deployment-relevant claim. Thus, the frontier plays a dual role: \textbf{\emph{comparison object}} across methods and \textbf{\emph{mechanism audit}} within the method family \citep{kwon2023efficient,lin2024kvquant,zhang2023h2o,xiao2023streamingllm}. 

\paragraph{\textbf{Reporting rule and interpretation.}}
A method is considered superior only when it improves the feasible frontier \emph{under the shared substrate, shared accounting rules, and shared quality constraints}. In particular, we do \textbf{\emph{not}} allow a method to claim wins by:
\begin{itemize}[leftmargin=1.2em,itemsep=2pt,topsep=2pt]
  \item changing the kernel family without disclosure,
  \item silently degrading quality to gain speed,
  \item excluding metadata from the reported memory budget,
  \item excluding controller/reconstruction/offload overhead from end-to-end cost,
  \item or comparing one method at a favorable but behaviorally unstable operating point against another at a stricter operating point.
\end{itemize}
This subsection therefore serves both as \textbf{\emph{experimental documentation}} and as a \textbf{\emph{fairness contract}}. That dual role is essential for long-context systems papers, because the meaning of a reported speedup depends not only on the algorithm but also on the runtime assumptions, budget definitions, and quality filters under which the result is produced.

\paragraph{\textbf{Practical reading guide for the main-paper figures.}}
The frontier plots in the main paper should therefore be read as follows: each point corresponds to a fully measured operating point \(\,x_m(\mathcal{B})\), filtered through quality and fairness constraints; the drawn curve is the Pareto envelope of the surviving points in the projected comparison plane; and the reported headline gains are summary functionals extracted from that envelope. In this sense, the frontier is not a visualization convenience but the actual empirical statement of the paper.

\section{Extended Results, Ablations, and Kernel-Realized Evidence}
\label{app:extended_results}

This appendix section provides the \textbf{\emph{full empirical support}} for the paper’s central claim: \textbf{\emph{Spherical KV}} improves the \textbf{memory--quality--throughput frontier} of long-context decoding under \textbf{\emph{deployment-realistic}} serving assumptions. The emphasis here is deliberately stronger than in the main paper. Rather than reporting only headline gains, we make explicit \textbf{(i)} the retained \textbf{\emph{iso-quality frontier}} across models and context regimes, \textbf{(ii)} the \textbf{\emph{kernel-realized}} evidence showing that throughput gains are accompanied by reductions in \textbf{HBM bytes/token} rather than hidden reconstruction overheads, \textbf{(iii)} the \textbf{\emph{causal ablations}} that separate compute-path effects from retention/rate-allocation effects, \textbf{(iv)} the \textbf{\emph{mechanism evidence and stability diagnostics}} showing how the controller allocates memory and prevents brittle long-horizon failures, and \textbf{(v)} the \textbf{\emph{representative operating points}} that anchor the frontier with concrete, reproducible read-offs. Throughout this section, all results are interpreted under the same \textbf{\emph{paged/ragged KV substrate}}, the same \textbf{\emph{end-to-end cost accounting}} rules, and the same \textbf{\emph{dense-relative quality tolerance}} used in the main paper. Accordingly, the appendix should be read not as a collection of auxiliary plots, but as the paper’s \textbf{\emph{audit trail}}: it establishes that the reported gains are \textbf{\emph{not}} artifacts of contiguous-cache assumptions, omitted metadata, relaxed quality targets, or format-only compression, but arise from a genuine shift in the feasible operating region of long-context serving.

\subsection{Main Result: Iso-Quality Frontier Shift}
\label{sec:appendix_main_frontier}

\paragraph{\textbf{Scope of this subsection.}}
The central empirical claim of \textbf{\emph{Spherical KV}} is a \textbf{\emph{frontier shift}} in the long-context, memory-bounded regime: under a shared \textbf{\emph{paged/ragged}} serving substrate and at \textbf{\emph{matched task quality}}, Spherical KV achieves \textbf{\emph{lower resident KV bytes/token}}, \textbf{\emph{lower HBM bytes/generated-token}}, and \textbf{\emph{higher steady-state decode throughput}} simultaneously. This subsection formalizes the frontier object used throughout the appendix, explains how it is instantiated in the existing \(3\times 3\) frontier figure, and defines the envelope-derived summary quantities reported in the paper. The emphasis is intentionally \textbf{\emph{deployment-true}}: all retained points obey the same substrate assumptions, the same quality tolerance, and the same end-to-end accounting rules \citep{kwon2023efficient,dao2022flashattention,dao2023flashattention2}.

\subsubsection{Operating points, budget sweeps, and matched-quality feasible sets}
\label{sec:appendix_frontier_operating_points}

\paragraph{\textbf{Operating points as measured deployment states.}}
For a fixed method \(m\), model \(\mathcal{M}\), workload \(\mathcal{W}\), and context regime \(L\), we define an \textbf{\emph{operating point}} \(p\) to be a concrete deployment configuration together with its measured outcomes. Operationally, an operating point is induced by selecting a method-specific budget/control setting \(\mathcal{B}\) and evaluating the resulting decode behavior under the shared substrate. Thus, for each method we sweep a discrete set of budget settings
\[
\mathcal{B}\in\{b_1,\dots,b_K\},
\]
where the semantic meaning of \(\mathcal{B}\) depends on the method family:
\begin{itemize}[leftmargin=1.3em,itemsep=2pt,topsep=2pt]
    \item for \textbf{\emph{Spherical KV}}, \(\mathcal{B}\) spans a joint setting of \textbf{retention level} and \textbf{tier/rate schedule}, since the method controls both \emph{which} states remain resident and \emph{at what precision};
    \item for \textbf{\emph{retention/eviction baselines}}, \(\mathcal{B}\) spans window size, eviction aggressiveness, or retain ratio \citep{xiao2023streamingllm,zhang2023h2o};
    \item for \textbf{\emph{KV quantization baselines}}, \(\mathcal{B}\) spans bitwidth, outlier handling, or mixed-precision settings \citep{lin2024kvquant};
    \item for any \textbf{\emph{format-only}} baseline, \(\mathcal{B}\) may additionally encode whether decode-time reconstruction or staging is required, in which case that cost is explicitly charged in the runtime and traffic metrics.
\end{itemize}

\paragraph{\textbf{Measured tuple associated with each budget setting.}}
Every budget setting \(\mathcal{B}\) yields a fully measured tuple
\[
x_m(\mathcal{B})
\;=\;
\Big(
b_{\mathrm{KV},m}(\mathcal{B}),
\;
b_{\mathrm{HBM},m}(\mathcal{B}),
\;
s_m(\mathcal{B}),
\;
q_m(\mathcal{B})
\Big),
\]
where:
\begin{itemize}[leftmargin=1.3em,itemsep=2pt,topsep=2pt]
    \item \(b_{\mathrm{KV},m}(\mathcal{B})\) is the \textbf{\emph{effective resident KV budget}} (bytes per active/addressable token),
    \item \(b_{\mathrm{HBM},m}(\mathcal{B})\) is the \textbf{\emph{HBM bytes per generated token}},
    \item \(s_m(\mathcal{B})\) is the \textbf{\emph{steady-state decode throughput}} (tok/s),
    \item \(q_m(\mathcal{B})\) is the \textbf{\emph{task-quality score}} for the workload under study.
\end{itemize}
When stability is relevant, the measured record is augmented with witness statistics such as seed variance, length drift, or trajectory disagreement; however, the frontier figure itself is a projection of the four primary quantities above.

\paragraph{\textbf{Deployment-true resident budget.}}
The primary x-axis quantity is the \textbf{\emph{effective resident KV budget}}
\[
b_{\mathrm{KV}}
\;\triangleq\;
\frac{M_{\mathrm{KV}}^{\mathrm{resident}}}{T_{\mathrm{active}}},
\]
where \(T_{\mathrm{active}}\) is the number of addressable tokens and
\[
M_{\mathrm{KV}}^{\mathrm{resident}}
=
M_{\mathrm{payload}}
+
M_{\mathrm{headers}}
+
M_{\mathrm{ptr}}
+
M_{\mathrm{tags}}
+
M_{\mathrm{prot}}
+
M_{\mathrm{frag}}.
\]
Here \(M_{\mathrm{payload}}\) denotes the actual KV/code payload, while \(M_{\mathrm{headers}}, M_{\mathrm{ptr}}, M_{\mathrm{tags}}, M_{\mathrm{prot}}, M_{\mathrm{frag}}\) account for page headers, pointer tables, tier/bitwidth tags, protection metadata, and fragmentation under paged/ragged allocation. This definition is intentionally strict: the frontier is meant to reflect the \textbf{\emph{actual cost model seen by the runtime}} rather than an idealized payload-only compression ratio \citep{kwon2023efficient}.

\paragraph{\textbf{HBM traffic as the systems-side companion variable.}}
The primary systems-side witness is
\[
b_{\mathrm{HBM}}
\;\triangleq\;
\frac{\mathrm{HBM}_{\mathrm{read}}+\mathrm{HBM}_{\mathrm{write}}}
{\#\text{ decode tokens}},
\]
measured over the same decode window used for throughput. This quantity is not plotted directly on the x-axis of the frontier figure, but it is recorded for every operating point and used to certify whether a nominal memory-side gain is \textbf{\emph{kernel-realized}}. A method may reduce \(b_{\mathrm{KV}}\) while failing to reduce \(b_{\mathrm{HBM}}\) if it pays a decode-time densification tax; this is precisely why both quantities are tracked \citep{dao2022flashattention,dao2023flashattention2}.

\paragraph{\textbf{Dense-relative quality filtering.}}
The frontier is not constructed from all swept points, but only from those that remain sufficiently close in quality to the dense-KV reference. Let
\[
Q^{\star}_{\mathrm{dense}}
\]
denote the best dense-KV quality achieved under the same model, workload, context regime, decode settings, and serving substrate. Fix a single tolerance \(\Delta\), shared across all methods in the comparison. Then the \textbf{\emph{matched-quality feasible set}} for method \(m\) is
\[
\mathcal{P}_m^{\Delta}
=
\Big\{
\mathcal{B}
:\;
q_m(\mathcal{B}) \ge Q^{\star}_{\mathrm{dense}}-\Delta
\Big\}.
\]
Equivalently, if \(p\) denotes the operating point induced by budget \(\mathcal{B}\), we may write
\[
p\in\mathcal{P}_m^{\Delta}
\quad\Longleftrightarrow\quad
q(p)\ge Q^{\star}_{\mathrm{dense}}-\Delta.
\]
Only points in \(\mathcal{P}_m^{\Delta}\) are retained for the iso-quality frontier. This filtering step ensures that the frontier remains a \textbf{\emph{quality-conditioned systems object}} rather than collapsing into a generic speed--memory plot \citep{liu2023lost,lin2024kvquant}.

\paragraph{\textbf{Interpretation of retained points.}}
Every retained point therefore has three simultaneous properties:
\begin{enumerate}[leftmargin=1.35em,itemsep=2pt,topsep=2pt]
    \item it is measured under the \textbf{\emph{same paged/ragged substrate}} and the same decode settings as the reference,
    \item it satisfies the \textbf{\emph{same dense-relative quality tolerance}} \(\Delta\),
    \item it carries a complete measured tuple \(\big(b_{\mathrm{KV}}, b_{\mathrm{HBM}}, s, q\big)\), with end-to-end accounting.
\end{enumerate}
This is why the frontier points are not merely visualization artifacts; they are the actual \textbf{\emph{feasible operating states}} of the deployment system under an explicit fairness contract.

\subsubsection{Main frontier result across models and context lengths}
\label{sec:appendix_frontier_grid}

\paragraph{\textbf{Structure of the existing \(3\times 3\) frontier figure.}}
Figure~\ref{fig:iso_quality_frontier} is the appendix’s primary result artifact. It visualizes the retained iso-quality frontier for three \textbf{\emph{models}} and three \textbf{\emph{context regimes}} under a fixed paged/ragged serving substrate. The figure is organized as follows:
\begin{itemize}[leftmargin=1.3em,itemsep=2pt,topsep=2pt]
    \item \textbf{rows} correspond to context regimes \(L\in\{8\mathrm{K},32\mathrm{K},128\mathrm{K}\}\),
    \item \textbf{columns} correspond to models (\textbf{Llama-3.1-8B}, \textbf{Qwen2.5-14B}, \textbf{gpt-oss}),
    \item the x-axis is \(b_{\mathrm{KV}}\) (lower is better),
    \item the y-axis is \(s\) (higher is better).
\end{itemize}
Thus, in every panel, the preferred direction is \textbf{\emph{up-left}}: lower effective resident KV at higher throughput.

\paragraph{\textbf{What each panel actually contains.}}
Each point in a panel corresponds to a fully measured operating point \(x_m(\mathcal{B})\), filtered through the matched-quality set \(\mathcal{P}_m^\Delta\). Gray or excluded points correspond to settings whose quality gap exceeds the tolerance \(\Delta\), and therefore are not eligible to support the main claim. For every method, the black polyline traces the \textbf{\emph{Pareto envelope}} over the retained points in the \((b_{\mathrm{KV}}, s)\) plane. Formally, if
\[
\Pi_m^\Delta
\;\triangleq\;
\Big\{
\big(b_{\mathrm{KV},m}(\mathcal{B}),\, s_m(\mathcal{B})\big)
:\;
\mathcal{B}\in\mathcal{P}_m^\Delta
\Big\},
\]
then the envelope consists of the non-dominated elements of \(\Pi_m^\Delta\), i.e., the points for which no other retained point is simultaneously cheaper in \(b_{\mathrm{KV}}\) and faster in \(s\).

\paragraph{\textbf{What the frontier figure establishes.}}
The figure supports one central statement:
\begin{quote}
\textbf{Under matched quality, Spherical KV shifts the feasible memory--throughput operating region up-left relative to Dense KV and strong baselines.}
\end{quote}
This statement is stronger than a single operating-point comparison because it does not merely say that one particular configuration is favorable; it says that the \textbf{\emph{retained trade-off surface itself}} improves. In other words, the method does not simply choose a different point on the same curve; it changes the curve.

\paragraph{\textbf{Why the frontier view is necessary.}}
A raw speedup number is ambiguous unless it is tied to a quality level, a budget definition, and a declared serving substrate. Likewise, a raw memory reduction says little if decode-time reconstruction erases the runtime benefit. The frontier representation resolves these ambiguities by:
\begin{enumerate}[leftmargin=1.35em,itemsep=2pt,topsep=2pt]
    \item fixing the \textbf{\emph{quality regime}} via \(\mathcal{P}_m^\Delta\),
    \item fixing the \textbf{\emph{deployment substrate}} via shared paged/ragged serving,
    \item displaying the \textbf{\emph{trade-off surface}} rather than isolated cherry-picked points.
\end{enumerate}
This is why Fig.~\ref{fig:iso_quality_frontier} is the correct top-level result figure for the appendix.

\begin{figure}[H]
  \centering
  \includegraphics[width=\textwidth]{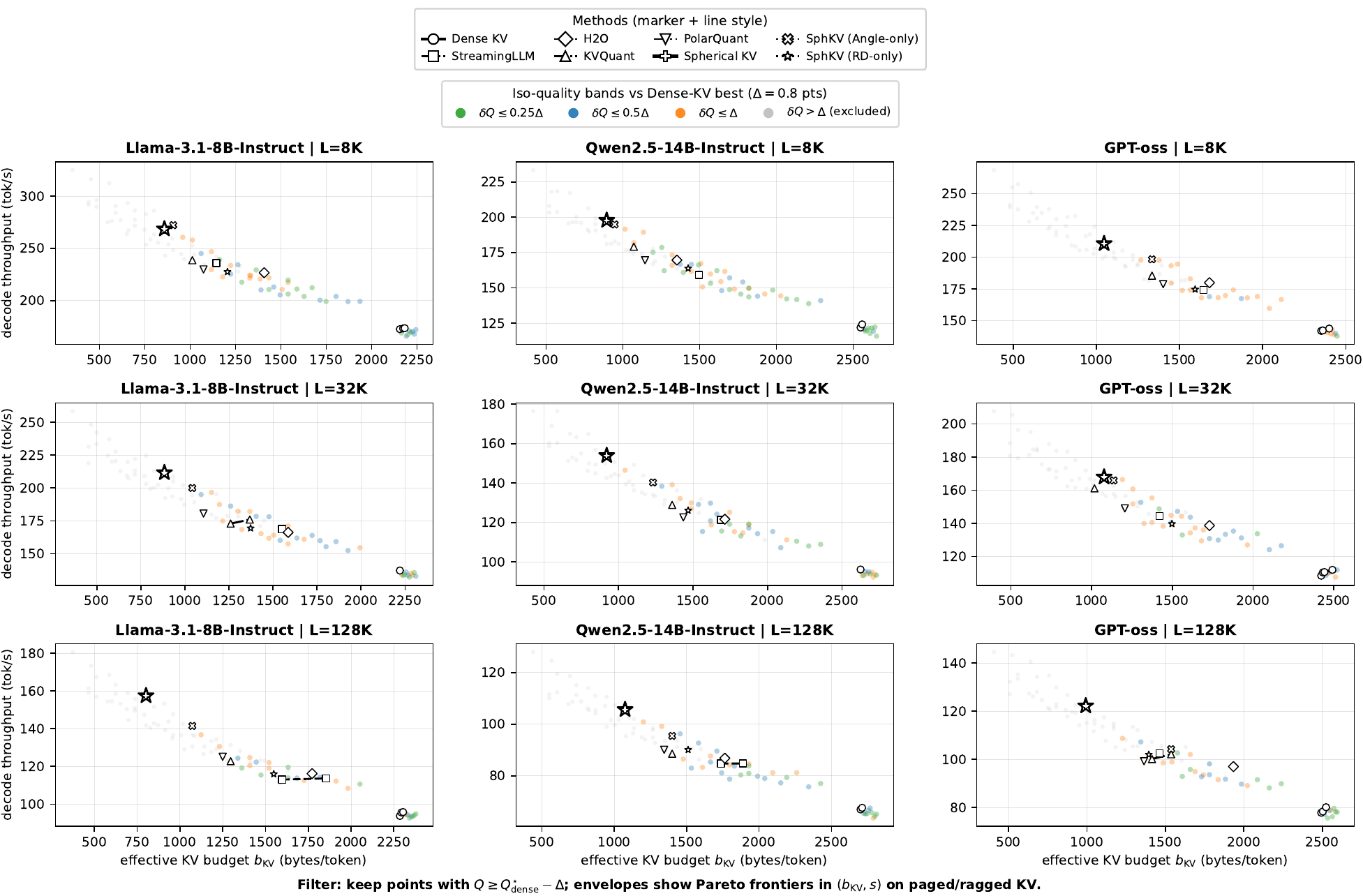}
  \caption{\textbf{Iso-quality Pareto frontiers for \emph{memory-bounded} decoding.}
Each panel plots \textbf{decode throughput} (tok/s; \textbf{higher is better}) versus \textbf{effective KV budget} $b_{\mathrm{KV}}$ (bytes/token; \textbf{lower is better}) under \textbf{paged/ragged serving}, across three models and context lengths $L\in\{8\mathrm{K},32\mathrm{K},128\mathrm{K}\}$.
Let $Q$ be the \textbf{quality score} (\textbf{higher is better}; defined in \S\,[X]) and let $Q^{\star}_{\mathrm{dense}}$ denote the \textbf{best Dense-KV quality} in the panel.
We enforce an \textbf{iso-quality constraint} by retaining only configurations with \textbf{$Q \geq Q^{\star}_{\mathrm{dense}}-\Delta$} (\textbf{$\Delta=0.8$} points).
Background dots are colored by the \textbf{quality gap} $\delta Q = Q^{\star}_{\mathrm{dense}}-Q$ (\textbf{gray}: excluded, $\delta Q>\Delta$).
For each method, the black polyline traces the \textbf{Pareto envelope} over retained points in $(b_{\mathrm{KV}},\text{tok/s})$.
The \textbf{$\star$} marks the \textbf{best throughput-per-byte} operating point for \textbf{Spherical KV} among retained configurations, highlighting an \textbf{up-left} shift of the \textbf{iso-quality frontier}.}
  \label{fig:appndx_iso_quality_frontier}
\end{figure}

\begin{table*}[ht!]
\centering
\scriptsize
\setlength{\tabcolsep}{4.2pt}
\renewcommand{\arraystretch}{1.13}
\caption{\textbf{Panel-by-panel matched-quality read-off for the iso-quality frontier.}
For each \((\text{model},L)\) panel, we compare the \(\text{Dense-KV}\) reference against the \(\star\)-marked representative \(\text{Spherical-KV}\) operating point, defined as the retained Spherical-KV configuration maximizing throughput-per-byte \(s/b_{\mathrm{KV}}\) under the iso-quality constraint \(Q \ge Q^{\star}_{\mathrm{dense}}-\Delta\) with \(\Delta=0.8\). Values are computed directly from the frontier CSV used to generate the figure.}
\label{tab:full_frontier_readoff}
\resizebox{\textwidth}{!}{%
\begin{tabular}{l c
                r r r
                r r r r
                r r}
\toprule
\multirow{2}{*}{\textbf{Model}} &
\multirow{2}{*}{\textbf{L}} &
\multicolumn{3}{c}{\textbf{Dense KV reference}} &
\multicolumn{4}{c}{\textbf{Representative Spherical KV (\(\star\))}} &
\multicolumn{2}{c}{\textbf{Derived gains}} \\
\cmidrule(lr){3-5} \cmidrule(lr){6-9} \cmidrule(lr){10-11}
&
&
\textbf{\(Q^{\star}_{\mathrm{dense}}\)} &
\textbf{\(b_{\mathrm{KV}}\)} &
\textbf{\(s\)} &
\textbf{\(Q\)} &
\textbf{\(\delta Q\)} &
\textbf{\(b_{\mathrm{KV}}\)} &
\textbf{\(s\)} &
\textbf{\(s/s_{\mathrm{dense}}\)} &
\textbf{\(b_{\mathrm{KV}}/b_{\mathrm{KV,dense}}\)} \\
\midrule

Llama-3.1-8B-Instruct & 8K
& 74.203 & 2138.400 & 173.392
& 73.951 & 0.252 & 844.105 & 268.519
& 1.549 & 0.395 \\

Llama-3.1-8B-Instruct & 32K
& 73.110 & 2156.000 & 134.138
& 72.980 & 0.130 & 847.623 & 211.604
& 1.578 & 0.393 \\

Llama-3.1-8B-Instruct & 128K
& 71.467 & 2251.600 & 95.816
& 71.152 & 0.314 & 784.316 & 157.472
& 1.644 & 0.348 \\

\midrule

Qwen2.5-14B-Instruct & 8K
& 78.627 & 2602.000 & 122.132
& 78.198 & 0.428 & 881.068 & 197.800
& 1.619 & 0.339 \\

Qwen2.5-14B-Instruct & 32K
& 77.563 & 2620.000 & 93.304
& 77.173 & 0.390 & 883.609 & 153.949
& 1.650 & 0.337 \\

Qwen2.5-14B-Instruct & 128K
& 75.889 & 2692.000 & 65.306
& 75.153 & 0.736 & 1054.298 & 105.639
& 1.618 & 0.392 \\

\midrule

GPT-oss & 8K
& 80.506 & 2510.000 & 137.749
& 79.775 & 0.731 & 1072.902 & 210.616
& 1.529 & 0.427 \\

GPT-oss & 32K
& 79.073 & 2528.000 & 111.464
& 78.776 & 0.297 & 1089.939 & 167.982
& 1.507 & 0.431 \\

GPT-oss & 128K
& 77.452 & 2600.000 & 75.640
& 77.252 & 0.200 & 1018.071 & 122.250
& 1.616 & 0.392 \\

\bottomrule
\end{tabular}%
}
\end{table*}

\subsubsection{Cross-model and cross-context invariance}
\label{sec:appendix_frontier_invariance}

\paragraph{\textbf{Why the grid is arranged by rows and columns.}}
The \(3\times 3\) structure is designed to test two orthogonal forms of invariance.

\noindent\textbf{Vertical scan (down a column).}
Holding the model fixed and moving from \(8\mathrm{K}\rightarrow 32\mathrm{K}\rightarrow 128\mathrm{K}\) increases the \textbf{\emph{KV-pressure regime}}. Resident footprint rises, paging and fragmentation effects become more consequential, and the decode path becomes increasingly memory-bound. A method whose advantage disappears under this downward scan is not robust in the true long-context regime.

\noindent\textbf{Horizontal scan (across a row).}
Holding the context regime fixed and moving across the three models tests \textbf{\emph{architectural generalization}}. If the gain only appears on one model family, it may reflect an accidental compatibility with one tokenizer, head geometry, or kernel layout. A method whose frontier shift remains visible across the row is more likely to reflect a general systems property rather than a model-specific artifact.

\paragraph{\textbf{Persistence across model families.}}
The first claim extracted from the grid is \textbf{\emph{cross-model persistence}}: the retained frontier shift remains visible for all three model families. This matters because the models differ in scale, head configuration, and practical decode characteristics, yet the same pattern persists: \textbf{\emph{at matched quality, Spherical KV retains more favorable memory--throughput operating points than Dense KV and the strongest baselines.}} This is the relevant form of generalization for a systems paper, because the claim is about deployment behavior rather than benchmark-specific fine-tuning.

\paragraph{\textbf{Strengthening with context length.}}
The second claim extracted from the grid is \textbf{\emph{context-dependent strengthening}}: the relative advantage becomes more informative, and typically larger, as the context regime becomes more difficult. The \(128\mathrm{K}\) row is the decisive stress test. At this scale,
\begin{itemize}[leftmargin=1.3em,itemsep=2pt,topsep=2pt]
    \item KV residency dominates the memory budget,
    \item paged allocation overhead becomes materially visible,
    \item decode-time memory traffic dominates arithmetic throughput,
    \item and naïve compression or eviction methods are more likely to fall outside the retained iso-quality band.
\end{itemize}
Thus, if the frontier shift remains visible in the bottom row, it is evidence of a \textbf{\emph{real operating-curve change}} rather than a short-context tuning artifact.

\paragraph{\textbf{Why \(128\mathrm{K}\) is the decisive regime.}}
The appendix should explicitly treat \(128\mathrm{K}\) as the hardest regime, not merely the largest one. At lower lengths, several methods may look superficially competitive because the decode loop is not yet fully dominated by KV traffic. By contrast, at \(128\mathrm{K}\), the system is decisively in the memory-bounded regime: the interaction of resident bytes, HBM traffic, block-local serving, and kernel fusion determines performance. Therefore, the persistence of the matched-quality up-left shift in the bottom row is the most direct evidence that Spherical KV changes the \textbf{\emph{true long-context deployment frontier}}.

\paragraph{\textbf{Global interpretation of the grid.}}
The correct appendix-level reading of the figure is therefore:
\begin{quote}
\textbf{The matched-quality frontier shift is visible across all rows and columns, and it becomes most informative in the \(128\mathrm{K}\) stress regime.}
\end{quote}
This is the right scientific summary: the effect is not tied to one model or one context, but to the consistent shape of the retained frontier under increasingly difficult operating regimes.

\subsubsection{Envelope-derived summary deltas}
\label{sec:appendix_frontier_deltas}

\paragraph{\textbf{Why scalar summaries are still needed.}}
The frontier curve is the correct comparison object, but readers also need compact, reproducible summary quantities that can be cited in text and compared across settings. These summaries should be derived from the retained envelope rather than from arbitrary single points, and they should respect the same quality and fairness constraints as the plotted frontier. For this reason, we report two envelope-derived quantities: \textbf{\emph{iso-quality speedup}} and \textbf{\emph{iso-throughput memory ratio}}.

\paragraph{\textbf{Iso-quality speedup.}}
Let \(s_{\mathrm{dense}}\) denote the dense-KV decode throughput under the same model, workload, context, and decode settings. Over the retained set \(\mathcal{P}_m^\Delta\), define
\[
\Gamma_s
\;=\;
\max_{p\in \mathcal{P}_m^\Delta}
\frac{s(p)}{s_{\mathrm{dense}}}.
\]
Equivalently, in budget notation,
\[
\Gamma_s
\;=\;
\max_{\mathcal{B}\in\mathcal{P}_m^\Delta}
\frac{s_m(\mathcal{B})}{s_{\mathrm{dense}}}.
\]
This quantity answers the systems question:
\begin{quote}
\emph{Among all quality-matched operating points, what is the largest throughput advantage the method can realize over Dense KV?}
\end{quote}
Because the maximization is restricted to the retained set, \(\Gamma_s\) cannot be inflated by sacrificing quality beyond the allowed tolerance.

\paragraph{\textbf{Iso-throughput memory ratio.}}
The complementary summary asks the inverse question: at a dense-like throughput target, how much effective resident memory is required? Let \(b_{\mathrm{KV,dense}}\) denote the dense-KV effective resident budget and define
\[
\Gamma_m
\;=\;
\min_{p:\, s(p)\ge s_{\mathrm{dense}}}
\frac{b_{\mathrm{KV}}(p)}{b_{\mathrm{KV,dense}}}.
\]
Equivalently,
\[
\Gamma_m
\;=\;
\min_{\mathcal{B}:\, s_m(\mathcal{B})\ge s_{\mathrm{dense}}}
\frac{b_{\mathrm{KV},m}(\mathcal{B})}{b_{\mathrm{KV,dense}}}.
\]
This quantity is smaller-is-better. It answers:
\begin{quote}
\emph{Among all operating points that are at least as fast as Dense KV, what is the smallest resident KV budget the method can achieve?}
\end{quote}
Thus \(\Gamma_m\) converts the frontier into a precise statement about \textbf{\emph{memory efficiency at fixed speed}}.

\paragraph{\textbf{Why these summaries are principled.}}
These two envelope-derived quantities are preferable to raw single-point speedups for four reasons:
\begin{enumerate}[leftmargin=1.35em,itemsep=2pt,topsep=2pt]
    \item \textbf{Quality conditioning.} \(\Gamma_s\) is computed only over \(\mathcal{P}_m^\Delta\), so it does not reward hidden quality degradation.
    \item \textbf{Speed conditioning.} \(\Gamma_m\) is computed only over points that meet or exceed dense throughput, so it does not reward trivially slow but memory-thrifty settings.
    \item \textbf{Shared accounting.} Both quantities use the same deployment-true definition of \(b_{\mathrm{KV}}\), including paging and metadata overhead.
    \item \textbf{Shared substrate.} Both are measured under the same paged/ragged serving substrate and the same end-to-end runtime contract.
\end{enumerate}
For a systems paper, these properties are essential; without them, a reported gain can simply reflect a weaker operating condition rather than a genuinely better method.

\paragraph{\textbf{Why they matter more than raw single-point speedups.}}
A raw single-point speedup reports one ratio at one operating condition, but it does not specify whether the point lies on the best retained part of the curve, whether its quality is comparable to the reference, whether its memory cost is realistic under paged serving, or whether another nearby point dominates it. By contrast, \(\Gamma_s\) and \(\Gamma_m\) summarize the best retained operating points \textbf{\emph{after}} the frontier has been constructed. They therefore inherit the logic of the envelope itself rather than bypassing it. In this sense, they are not ad hoc statistics; they are \textbf{\emph{functionals of the feasible retained frontier}}.

\paragraph{\textbf{Connection to the main-paper claims.}}
These summary deltas are the quantitative bridge between the appendix and the main paper. The main paper claims that Spherical KV:
\begin{itemize}[leftmargin=1.3em,itemsep=2pt,topsep=2pt]
    \item improves decode throughput at matched quality,
    \item reduces effective resident KV cost,
    \item and does so most clearly in the long-context, memory-bounded regime.
\end{itemize}
The appendix frontier and the two envelope-derived quantities formalize exactly those claims. Specifically:
\begin{itemize}[leftmargin=1.3em,itemsep=2pt,topsep=2pt]
    \item \(\Gamma_s>1\) certifies a matched-quality throughput gain,
    \item \(\Gamma_m<1\) certifies a resident-memory reduction at dense-like speed,
    \item and the row-wise behavior of these quantities across \(L\in\{8\mathrm{K},32\mathrm{K},128\mathrm{K}\}\) shows whether the gain strengthens in the true long-context regime.
\end{itemize}
Thus, the appendix formalism is not separate from the paper’s headline claim; it is the rigorous statement underlying it.

\paragraph{\textbf{Companion traffic annotation.}}
Although not part of the two scalar definitions above, we additionally record the corresponding
\[
b_{\mathrm{HBM}}(p)
\]
at the operating points achieving \(\Gamma_s\) and \(\Gamma_m\). This matters because, in a memory-bounded decode regime, any meaningful throughput gain should be explainable by a reduction in HBM traffic. Therefore, the appendix should interpret \(\Gamma_s\) and \(\Gamma_m\) together with their associated \(b_{\mathrm{HBM}}\) values as the complete deployment-relevant summary of the retained frontier.

\subsection{Kernel-Realized Efficiency: Why the Speedup Is Real}
\label{sec:kernel_realized_efficiency_appendix}

\paragraph{\textbf{Purpose of this section.}}
The central systems claim of \textbf{\emph{Spherical KV}} is stronger than ``the stored representation is smaller.'' The claim is that the reduction in resident KV state is \textbf{\emph{realized by the decode kernel}} as a reduction in \textbf{\emph{HBM traffic}} and therefore as an increase in \textbf{\emph{steady-state decode throughput}} under a \textbf{\emph{paged/ragged}} serving substrate. This distinction matters because many KV methods improve \emph{storage format} while leaving the actual decode bottleneck largely unchanged: they reduce bytes-at-rest but still pay a decode-time \textbf{\emph{densification tax}} through unpacking, dequantization, staging, or dense reconstruction before dot-products. In a memory-bound regime, such methods can look impressive in payload statistics while failing to move the actual serving frontier. The role of this subsection is therefore to formalize what it means for a gain to be \textbf{\emph{kernel-realized}}, define the primary systems witness used to certify that claim, and specify the exact runtime protocol under which the claim is measured \citep{dao2022flashattention,dao2023flashattention2,kwon2023efficient,ye2025flashinfer}. 

\subsubsection{Critical-path accounting and the no-reconstruction rule}
\label{sec:critical_path_accounting}

\paragraph{\textbf{Decode-time cost must be stated at the critical-path level.}}
Long-context inference is dominated by the repeated decode loop rather than by one-time prefill. For that reason, any claim about runtime improvement must be expressed in terms of the \textbf{\emph{critical path}} of one decode step. We write the total decode-time cost as
\[
T_{\mathrm{decode}}
\;=\;
T_{\mathrm{page\ lookup}}
\;+\;
T_{\mathrm{KV\ read}}
\;+\;
T_{\mathrm{similarity}}
\;+\;
T_{\mathrm{softmax/attn}}
\;+\;
T_{\mathrm{proj}}
\;+\;
T_{\mathrm{misc}}.
\]
Each term has a precise systems meaning:
\begin{itemize}[leftmargin=1.35em,itemsep=2pt,topsep=2pt]
    \item \(T_{\mathrm{page\ lookup}}\): indirection through pointer tables and block headers under paged/ragged KV serving;
    \item \(T_{\mathrm{KV\ read}}\): global-memory movement required to fetch keys/values or their compressed surrogates from HBM;
    \item \(T_{\mathrm{similarity}}\): computation of query--key similarity or its compressed-domain analogue;
    \item \(T_{\mathrm{softmax/attn}}\): normalization, weighting, and attention accumulation;
    \item \(T_{\mathrm{proj}}\): value aggregation and output projection;
    \item \(T_{\mathrm{misc}}\): residual decode-time overhead, including launch overheads, runtime bookkeeping, and non-attention residual costs.
\end{itemize}
This decomposition is not decorative. It makes explicit \textbf{\emph{where}} a KV method can legitimately claim gains, and it prevents compression claims that merely move work across hidden stages.

\paragraph{\textbf{Kernel-realized versus format-only methods.}}
We distinguish two classes of methods at the systems level.

\noindent\textbf{Kernel-realized methods.}
A method is \textbf{\emph{kernel-realized}} if the representation stored in the KV cache is \textbf{\emph{consumed directly in the decode kernel}}. In that case, the reduction in resident bytes can propagate into reductions in \(T_{\mathrm{KV\ read}}\) and possibly \(T_{\mathrm{similarity}}\), because the hot loop does not first reconstruct a dense surrogate of the original KV state.

\noindent\textbf{Format-only methods.}
A method is \textbf{\emph{format-only}} if it stores a compact representation but requires decode-time \textbf{unpacking}, \textbf{dequantization}, or \textbf{dense materialization} before the similarity kernel can run. Such methods may improve \(M_{\mathrm{KV}}^{\mathrm{resident}}\), but they often reintroduce global-memory movement and additional launches in the decode path, thereby erasing much of the apparent systems gain.

\paragraph{\textbf{The no-reconstruction rule.}}
This appendix adopts a strict \textbf{\emph{no-reconstruction rule}}:

\begin{quote}
\textbf{A compressed representation supports a throughput claim only if the decode kernel consumes that representation directly, without reconstructing dense keys in the critical path.}
\end{quote}

Formally, let \(r_i\) and \(\phi_i\) denote the compact radial and angular codes for key state \(i\), and let \(q_t\) denote the query at decode step \(t\). In a kernel-realized implementation, similarity is computed in compressed domain:
\[
\tilde{\ell}_{t,i}
\;=\;
f_{\mathrm{comp}}\!\big(q_t,\tilde{r}_i,\tilde{\phi}_i\big),
\]
where \(f_{\mathrm{comp}}\) is a fused compressed-domain kernel. By contrast, a format-only implementation first computes
\[
\tilde{k}_i
\leftarrow
\mathrm{decode}(\tilde{r}_i,\tilde{\phi}_i),
\qquad
\tilde{\ell}_{t,i}
\leftarrow
\langle q_t,\tilde{k}_i\rangle.
\]
The latter path inserts an additional decode-stage transformation into the critical path and therefore changes not only the arithmetic but the memory-traffic profile. This distinction is the heart of the paper’s systems thesis.

\paragraph{\textbf{A traffic-level view of reconstruction cost.}}
At the level of memory traffic, the reconstruct-then-dot path adds an explicit densification term. If \(b_{\mathrm{read}}(\mathrm{codes})\) denotes reading compressed codes, \(b_{\mathrm{write}}(\mathrm{dense}\ K)\) the staging of dense keys, and \(b_{\mathrm{read}}(\mathrm{dense}\ K)\) the subsequent read of those dense keys by the similarity kernel, then a reconstruction-based method incurs approximately
\[
b_{\mathrm{HBM}}^{\mathrm{Recon}}
\;\approx\;
b_{\mathrm{HBM}}^{\mathrm{NoRecon}}
\;+\;
b_{\mathrm{write}}(\mathrm{dense}\ K)
\;+\;
b_{\mathrm{read}}(\mathrm{dense}\ K)
\;+\;
b_{\mathrm{aux}},
\]
where \(b_{\mathrm{aux}}\) absorbs auxiliary traffic such as staging buffers and launch-side bookkeeping. In long-context decode, where the system is already near the bandwidth roofline, this densification term is often large enough to nullify much of the apparent compression gain \citep{dao2022flashattention,lin2024kvquant}.

\paragraph{\textbf{Implication for empirical claims.}}
Accordingly, this appendix treats resident-memory reduction alone as insufficient evidence. A method may be interesting as a \textbf{\emph{representation}} even if it is not kernel-realized, but it supports a \textbf{\emph{throughput}} claim only if the reduction in stored bytes is visible in the critical-path terms \(T_{\mathrm{KV\ read}}\) and \(T_{\mathrm{similarity}}\) under the shared substrate. This is the formal standard used throughout the rest of the results section.

\subsubsection{HBM bytes/token as the primary systems witness}
\label{sec:hbm_primary_witness}

\paragraph{\textbf{Why HBM traffic is the decisive systems quantity.}}
In long-context decode, the dominant runtime bottleneck is usually not arithmetic throughput but \textbf{\emph{global-memory movement}}. Each newly generated token must attend over a large historical prefix, forcing repeated reads of historical KV pages. In this regime, the most direct witness of an improvement is not merely resident footprint, but the number of bytes that the decode step must stream through HBM. For this reason, the appendix uses
\[
b_{\mathrm{HBM}}
\;\triangleq\;
\frac{\mathrm{HBM}_{\mathrm{read}}+\mathrm{HBM}_{\mathrm{write}}}
{\#\text{ decode tokens}}
\]
as the primary systems witness. Here \(\mathrm{HBM}_{\mathrm{read}}\) and \(\mathrm{HBM}_{\mathrm{write}}\) are measured from GPU profiling counters over the decode measurement window.

\paragraph{\textbf{Lower resident memory does not imply lower HBM traffic.}}
A common failure mode in KV-compression evaluation is to assume that lower resident bytes automatically translate into lower runtime traffic. This is false in general. The resident-memory budget
\[
b_{\mathrm{KV}}
\;=\;
\frac{M_{\mathrm{KV}}^{\mathrm{resident}}}{T_{\mathrm{active}}}
\]
measures the size of the cache state that remains addressable during decode, but it says nothing about how many times that state is read, transformed, or rewritten during the critical path. A method can therefore reduce \(b_{\mathrm{KV}}\) and still fail to improve runtime if decode-time reconstruction, staging, or auxiliary passes inflate
\[
b_{\mathrm{HBM}}
\;=\;
\frac{\mathrm{HBM}_{\mathrm{read}}+\mathrm{HBM}_{\mathrm{write}}}{\#\text{ decode tokens}}.
\]
This is exactly why both quantities must be tracked: \(b_{\mathrm{KV}}\) measures the \textbf{\emph{resident-state burden}}, while \(b_{\mathrm{HBM}}\) measures the \textbf{\emph{runtime bottleneck}}.

\paragraph{\textbf{Why reconstruction can erase apparent resident-memory gains.}}
Suppose two methods achieve the same resident-memory budget \(b_{\mathrm{KV}}\), but one uses compressed-domain similarity while the other reconstructs dense keys before attention. Then, even though the resident cache appears equally compact, the second method may incur a substantially larger decode traffic:
\[
b_{\mathrm{HBM}}^{\mathrm{recon}}
\;=\;
b_{\mathrm{HBM}}^{\mathrm{stream}}
\;+\;
b_{\mathrm{densify}}
\;+\;
b_{\mathrm{stage}},
\]
where \(b_{\mathrm{densify}}\) captures reconstruction traffic and \(b_{\mathrm{stage}}\) captures auxiliary buffer movement. In the memory-bound regime, throughput is governed far more directly by \(b_{\mathrm{HBM}}\) than by \(b_{\mathrm{KV}}\). Therefore, resident-memory reduction without traffic reduction is not sufficient evidence of a real systems win.

\paragraph{\textbf{Why IO reduction is the real proof in long-context decode.}}
The motivation follows directly from IO-aware attention analyses \citep{dao2022flashattention,dao2023flashattention2}. Once the decode step is bandwidth-limited, the most meaningful way to certify a systems improvement is to show that the number of bytes streamed through HBM per generated token decreases. If tok/s improves while \(b_{\mathrm{HBM}}\) does not change materially, then one of three explanations is more plausible than a genuine KV improvement:
\begin{enumerate}[leftmargin=1.35em,itemsep=2pt,topsep=2pt]
    \item transient runtime noise or cache luck,
    \item a shifted cost hidden in another stage,
    \item or an unfair comparison in kernel/runtime configuration.
\end{enumerate}
By contrast, when lower \(b_{\mathrm{KV}}\), lower \(b_{\mathrm{HBM}}\), and higher \(s\) co-occur under fixed decode settings and a shared substrate, the systems argument becomes much harder to challenge.

\paragraph{\textbf{Measured witness tuple.}}
For every representative operating point and every frontier read-off, the appendix therefore treats the systems witness tuple
\[
\big(
b_{\mathrm{KV}},
\;
b_{\mathrm{HBM}},
\;
s
\big)
\]
as inseparable. A method supports the paper’s central systems claim only when these three quantities move in the intended direction under matched quality:
\begin{itemize}[leftmargin=1.35em,itemsep=2pt,topsep=2pt]
    \item \(b_{\mathrm{KV}}\downarrow\) : less resident cache state,
    \item \(b_{\mathrm{HBM}}\downarrow\) : less traffic through the actual bottleneck,
    \item \(s\uparrow\) : higher realized decode throughput.
\end{itemize}
This is the paper’s operational definition of a \textbf{\emph{kernel-realized speedup}}.

\subsubsection{Streaming and fused consumption under paged KV}
\label{sec:streaming_fused_paged}

\paragraph{\textbf{What ``streaming + fused'' means in this paper.}}
The phrase \textbf{\emph{streaming + fused}} is used in a specific systems sense. Under paged/ragged KV serving, a decode kernel is said to be streaming and fused if it satisfies the following conditions:
\begin{enumerate}[leftmargin=1.35em,itemsep=2pt,topsep=2pt]
    \item \textbf{One-pass block-local reads.} KV blocks are read once from HBM in a page-local pattern, with pointer-table resolution but without repeated revisits to the same block for staging or reconstruction.
    \item \textbf{No auxiliary staging buffers.} The decode path does not write intermediate dense keys or auxiliary reconstructed representations to HBM.
    \item \textbf{No dense key materialization.} Similarity is computed directly from the stored representation, not from a reconstructed dense surrogate.
    \item \textbf{Shared paged substrate.} The same paged/ragged KV substrate is used for all compared methods, so differences in traffic arise from representation consumption rather than from different serving backends.
\end{enumerate}

\paragraph{\textbf{Formalizing one-pass block-local reads.}}
Let the KV cache be partitioned into pages of size \(P\) tokens, indexed by a pointer table. For decode step \(t\), let \(\mathcal{G}_t\) denote the set of accessed pages. In an ideal streaming implementation, each page \(g\in\mathcal{G}_t\) is fetched once and consumed directly, so the read-side traffic is approximately
\[
\mathrm{HBM}_{\mathrm{read},t}
\;\approx\;
\sum_{g\in\mathcal{G}_t}
\mathrm{bytes}(g)
\;+\;
\mathrm{bytes}(\mathrm{ptr/header}),
\]
up to bounded runtime overhead. By contrast, a non-streaming implementation may revisit or restage the same page multiple times, yielding
\[
\mathrm{HBM}_{\mathrm{read},t}^{\mathrm{nonstream}}
\;\approx\;
\sum_{g\in\mathcal{G}_t}
\kappa_g\,\mathrm{bytes}(g)
\;+\;
\mathrm{bytes}(\mathrm{ptr/header})
\;+\;
\mathrm{bytes}(\mathrm{stage}),
\]
with \(\kappa_g>1\) for some pages. This multiplicative revisit factor is exactly what streaming/fused kernels are designed to avoid.

\paragraph{\textbf{Why no dense key materialization matters.}}
Dense key materialization is especially harmful in the long-context regime because it introduces both writes and rereads to HBM. If the compressed state for key \(i\) is \((\tilde{r}_i,\tilde{\phi}_i)\), then the reconstruct-then-dot path typically performs
\[
(\tilde{r}_i,\tilde{\phi}_i)
\;\rightarrow\;
\tilde{k}_i
\;\rightarrow\;
\langle q_t,\tilde{k}_i\rangle,
\]
where the middle stage \(\tilde{k}_i\) may be materialized as a dense vector. In a fused compressed-domain path, the similarity is computed instead as
\[
\tilde{\ell}_{t,i}
=
f_{\mathrm{comp}}(q_t,\tilde{r}_i,\tilde{\phi}_i),
\]
so the dense intermediate never exists in HBM. This distinction is what allows a reduction in \(b_{\mathrm{KV}}\) to survive into a reduction in \(b_{\mathrm{HBM}}\).

\paragraph{\textbf{Shared paged substrate as a fairness requirement.}}
The same paged/ragged KV substrate is used for all methods. This requirement matters because otherwise one method could appear better simply by benefiting from a more favorable backend. In the appendix, ``streaming + fused'' is therefore not an implementation flourish; it is a declared comparison contract. It means that the compared methods must live under the same page size, pointer-table structure, allocator behavior, and batching regime, so any observed reduction in traffic is attributable to the method’s representation and kernel consumption path, not to a hidden substrate change \citep{kwon2023efficient,ye2025flashinfer}.


\begin{figure*}[t]
  \centering
  \includegraphics[width=\textwidth]{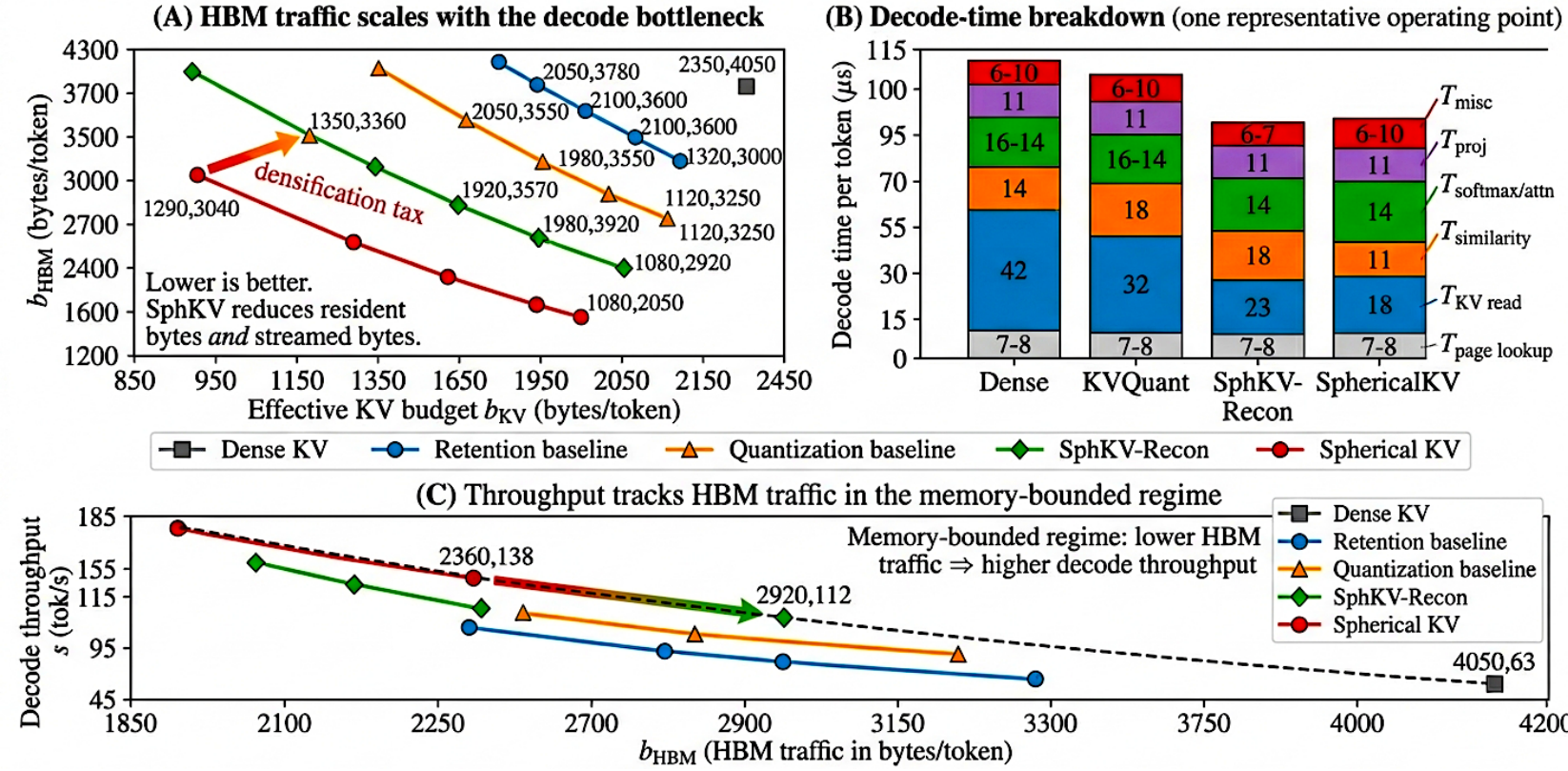}
  \caption{\textbf{Kernel-realized efficiency witness under paged KV serving.}
  \textbf{(A)} \textbf{HBM bytes/token} versus \textbf{effective resident KV budget} \(b_{\mathrm{KV}}\). A representation-level improvement is only deployment-meaningful if lower resident bytes are accompanied by lower HBM traffic in the decode path. \textbf{Spherical KV} moves down-left relative to \textbf{Dense KV} and reconstruction-heavy baselines, showing that its memory savings survive the kernel path. \textbf{(B)} \textbf{Decode-time breakdown} at one representative operating point, decomposed as
  \(T_{\mathrm{decode}}
  =
  T_{\mathrm{page\ lookup}}
  +
  T_{\mathrm{KV\ read}}
  +
  T_{\mathrm{similarity}}
  +
  T_{\mathrm{softmax/attn}}
  +
  T_{\mathrm{proj}}
  +
  T_{\mathrm{misc}}\).
  The dominant reduction for \textbf{Spherical KV} appears in \(T_{\mathrm{KV\ read}}\), without the compensating reconstruction cost paid by format-only baselines. \textbf{(C)} \textbf{Decode throughput} \(s\) versus \textbf{HBM traffic} \(b_{\mathrm{HBM}}\), showing that in the memory-bounded long-context regime, lower HBM traffic tracks higher throughput. Together, these panels certify that the observed speedup is \textbf{kernel-realized}, not merely \textbf{format-only}.}
  \label{fig:kernel_realized_proof}
\end{figure*}

\paragraph{\textbf{Interpretive summary.}}
This section supplies the appendix-level proof of the paper’s systems claim. The argument is not that Spherical KV merely stores fewer bytes. The argument is that, under a shared paged/ragged substrate, its stored representation is \textbf{\emph{consumed directly in the hot loop}}, thereby lowering \(b_{\mathrm{HBM}}\) and increasing \(s\) in the actual decode regime that matters. That is the sense in which the speedup is \textbf{\emph{real}}.

\subsubsection{Decode-window protocol and kernel-parity controls}
\label{sec:decode_window_kernel_parity}

\paragraph{\textbf{Why the systems proof must be protocolized.}}
A kernel-realized claim is only convincing if the measurement protocol itself is auditable. Apparent speedups can otherwise arise from warm-up artifacts, first-step paging effects, batch-shape changes, or kernel retuning unrelated to the representation. This subsection therefore states the exact decode-window and parity protocol used to turn the systems argument into a reproducible measurement procedure.

\paragraph{\textbf{Fixed warm-up and fixed measurement window.}}
All throughput and counter measurements are computed on a fixed decode window after warm-up. Let \(N_{\mathrm{warm}}\) denote the number of warm-up tokens and \(N_{\mathrm{meas}}\) the number of measured decode tokens. Then the reported throughput is
\[
s
=
\frac{N_{\mathrm{meas}}}{t_{\mathrm{decode}}^{\mathrm{meas}}},
\]
where \(t_{\mathrm{decode}}^{\mathrm{meas}}\) excludes the warm-up tokens. The same \((N_{\mathrm{warm}}, N_{\mathrm{meas}})\) pair is used across methods within a comparison set. This prevents one method from benefiting from a favorable measurement window or a lower warm-up burden.

\paragraph{\textbf{Repeated trials and summary statistics.}}
To reduce sensitivity to transient system noise, every reported systems number is measured over repeated trials. If \(s^{(r)}\) denotes the throughput in trial \(r\), then the appendix reports the median or mean across trials, together with a variability statistic when appropriate:
\[
\bar{s}
=
\frac{1}{R}\sum_{r=1}^{R}s^{(r)},
\qquad
\mathrm{med}(s^{(1)},\dots,s^{(R)}).
\]
The same repeated-trial policy is applied to \(b_{\mathrm{HBM}}\) and any other profiler-derived statistics. The point is not merely to smooth noise, but to ensure that a claimed gain is stable under repeated execution.

\paragraph{\textbf{Same batching regime.}}
All compared methods are evaluated under the same batching regime: same admission control, same dynamic/static batching policy, same preemption mode, and same batch-size constraints. This matters because batching affects page reuse, allocator pressure, and memory-traffic patterns. A method must not be allowed to appear faster simply because it was run under a more favorable batch geometry.

\paragraph{\textbf{Kernel-parity requirements.}}
Kernel parity is enforced as strictly as the method family allows. In particular:
\begin{itemize}[leftmargin=1.35em,itemsep=2pt,topsep=2pt]
    \item methods with comparable storage/consumption patterns are run under the same kernel family where possible;
    \item page size, head dimension handling, and launch-shape constraints are held fixed unless a method fundamentally requires otherwise;
    \item any divergence in kernel family or launch behavior is explicitly disclosed and treated as part of the method’s operating condition.
\end{itemize}
This requirement is essential because otherwise a claimed representation improvement could be confounded with a kernel retuning advantage unrelated to the algorithm.

\paragraph{\textbf{Why this protocol is enough to support the systems claim.}}
Taken together, fixed warm-up, fixed measurement window, repeated trials, common batching, and kernel parity turn the systems proof into an auditable protocol rather than an implementation anecdote. They ensure that when the appendix reports lower \(b_{\mathrm{HBM}}\) and higher \(s\), the result is not an artifact of measurement asymmetry.

\subsubsection{Kernel-realism checklist}
\label{sec:kernel_realism_checklist}

\paragraph{\textbf{Purpose of the checklist.}}
The checklist below is not a stylistic convenience; it is the appendix’s compact certification block for throughput claims. Its role is to make explicit what must be true before a compression result can be interpreted as a real serving improvement. In other words, it tells the reader what the appendix has verified before calling a gain \textbf{\emph{kernel-realized}}.

\vspace{4pt}
\begin{center}
\fbox{%
\parbox{0.97\linewidth}{%
\textbf{Kernel-realism checklist (enforced for any tok/s claim).}
\begin{itemize}[leftmargin=1.2em,itemsep=2pt,topsep=2pt]
  \item \textbf{P1: End-to-end decode.} Report tok/s from the full decode loop under paged/ragged serving \citep{kwon2023efficient}, not a microkernel in isolation.
  \item \textbf{P2: No hidden reconstruction.} The similarity/attention kernel consumes the stored representation directly; any densification/reconstruction path is labeled \emph{format-only} \citep{dao2022flashattention,lin2024kvquant}.
  \item \textbf{P3: Single-pass streaming.} KV is read once in a block-local pattern; no extra staging reads/writes beyond the kernel-realized contract.
  \item \textbf{P4: HBM bytes as primary witness.} Report DRAM/HBM read+write bytes per generated token; log L2/sector bytes as secondary diagnostics.
  \item \textbf{P5: Fixed decode window.} Same warm-up policy and same measured token window across methods; report median over repeated trials.
  \item \textbf{P6: Kernel config parity.} Same kernel family and launch constraints (batching, page size, head dim, occupancy constraints); differences must arise from \textbf{representation consumption}, not kernel retuning.
\end{itemize}
}%
}
\end{center}

\paragraph{\textbf{What the checklist certifies.}}
If all six items are satisfied, then a throughput improvement can be interpreted as a \textbf{\emph{representation-induced systems gain}} rather than as a side effect of hidden reconstruction, unmatched runtime settings, or unfair kernel engineering. Put differently, the checklist certifies that the observed improvement is attributable to the combination of \textbf{\emph{smaller resident state}} and \textbf{\emph{direct compressed-domain consumption}} under the same serving substrate.

\paragraph{\textbf{Optional appendix figure (if one more systems-proof panel is desired).}}
If additional visual evidence is desired, the following compact figure can be added without changing the main paper. It is intended as a proof-style appendix figure rather than a new headline result.

\subsection{Ablations: What Actually Drives the Frontier Shift}
\label{sec:appendix_ablations}

\paragraph{\textbf{Purpose of this section.}}
The frontier shift established in Sec.~\ref{sec:appendix_main_frontier} is an \textbf{\emph{outcome}}; this subsection asks for its \textbf{\emph{cause}}. In particular, we separate three candidate explanations: \textbf{(i)} a \textbf{\emph{compute-path}} effect, namely that compressed-domain attention removes decode-time densification and therefore lowers HBM traffic; \textbf{(ii)} a \textbf{\emph{control-path}} effect, namely that a better keep/drop + rate policy lowers resident KV cost without unacceptable quality loss; and \textbf{(iii)} a genuinely \textbf{\emph{joint}} effect, in which compressed-domain compute and rate--distortion allocation reinforce one another under paged/ragged serving. The aim of the ablation suite is therefore not merely to show that individual components matter, but to determine whether the observed frontier shift is \textbf{\emph{additive}} or \textbf{\emph{non-additive}}. That distinction is central to the paper’s claim: if the gain were fully explained by one lever alone, then the method would be reducible to a simpler alternative; if not, then the co-design itself is the contribution. 

\subsubsection{Ablation protocol and evaluation setting}
\label{sec:appendix_ablation_protocol}

\paragraph{\textbf{Evaluation setting.}}
All ablations are run under the same \textbf{\emph{paged/ragged KV substrate}}, the same decode engine, the same batching regime, and the same kernel family wherever kernel parity is possible. The goal is to isolate the effect of the ablated mechanism rather than confound it with runtime differences. Concretely, the ablation suite is evaluated at fixed long-context regimes
\[
L\in\{32\mathrm{K},\,128\mathrm{K}\},
\]
with the \(128\mathrm{K}\) regime serving as the primary stress-test slice shown in the existing A0--A5 figure. Within each \((\text{model},L)\) setting, we sweep a common set of effective KV budgets
\[
\mathcal{B}\in\{\mathcal{B}_1,\dots,\mathcal{B}_K\},
\]
where \(\mathcal{B}\) is always interpreted through the deployment-true resident-memory metric
\[
b_{\mathrm{KV}}
=
\frac{M_{\mathrm{KV}}^{\mathrm{resident}}}{T_{\mathrm{active}}}.
\]

\paragraph{\textbf{Shared quality filtering and decode settings.}}
All ablation variants are subjected to the same dense-relative quality filter used for the main frontier:
\[
\mathcal{P}_m^\Delta
=
\Big\{
\mathcal{B}
:\;
q_m(\mathcal{B}) \ge Q_{\mathrm{dense}}^\star-\Delta
\Big\},
\]
with the same fixed tolerance \(\Delta\) across variants. Decode settings are also held fixed: same sampling configuration, same prompt serialization, same maximum new tokens, same warm-up, and same measured decode window. Accordingly, any observed change in the retained frontier should be interpreted as a consequence of the ablated mechanism rather than a change in operating conditions. This is crucial because the purpose of D.3 is explicitly \textbf{\emph{causal}}: to separate \textbf{\emph{compute-path effects}} from \textbf{\emph{control-path effects}} under a shared serving contract. 
\paragraph{\textbf{Observed tuple per ablation point.}}
For each ablation variant \(v\), budget \(\mathcal{B}\), model \(\mathcal{M}\), and regime \(L\), we record
\[
x_v(\mathcal{B})
=
\Big(
b_{\mathrm{KV},v}(\mathcal{B}),
\;
b_{\mathrm{HBM},v}(\mathcal{B}),
\;
s_v(\mathcal{B}),
\;
q_v(\mathcal{B})
\Big),
\]
and, when relevant, associated stability witnesses. The ablation frontier is therefore evaluated in the same measurement space as the main frontier; only the internal mechanism differs.

\begin{figure}[t]
  \centering
  \includegraphics[width=\textwidth]{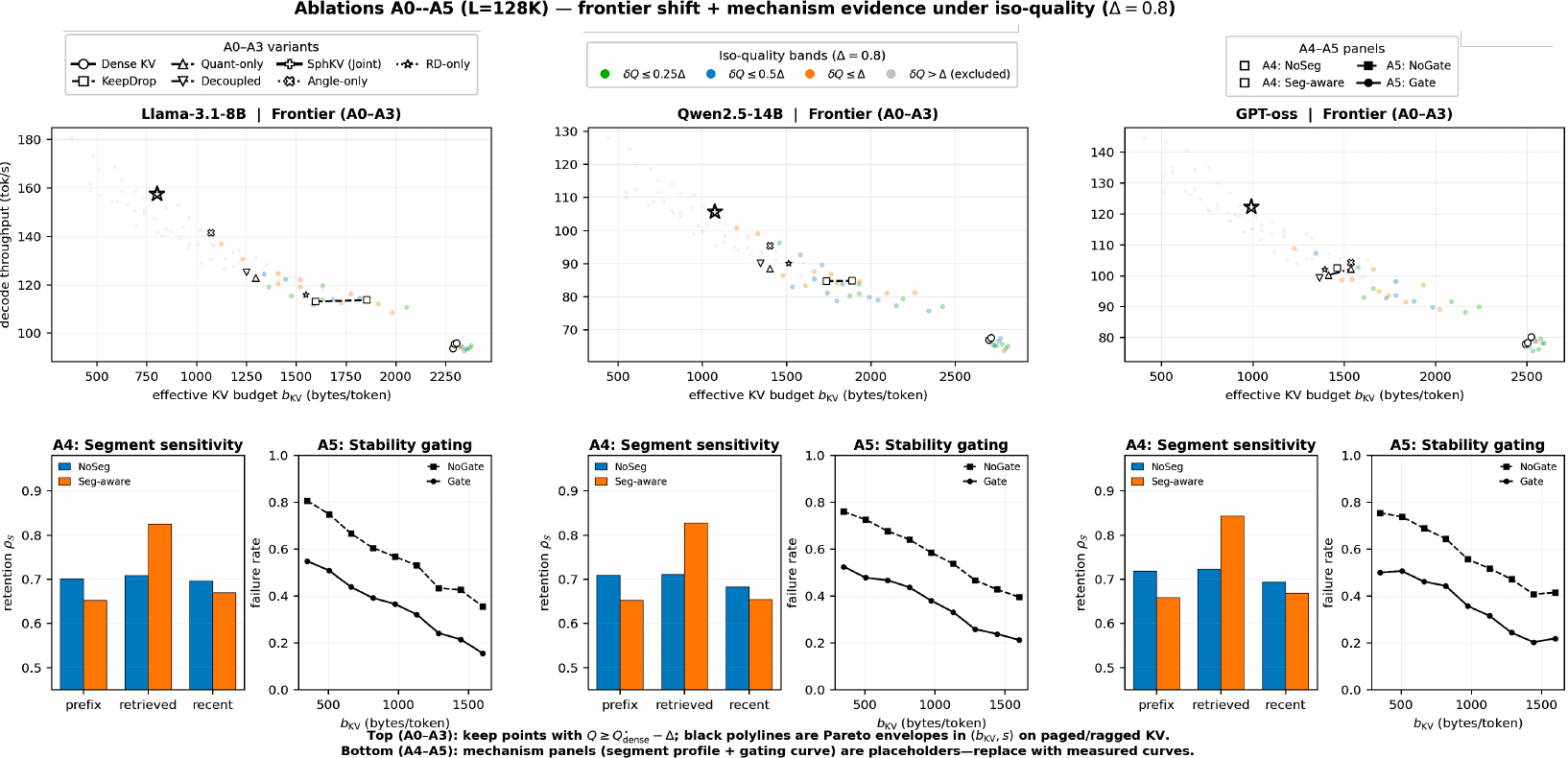}
  \caption{\textbf{Ablations A0--A5 at extreme context (L=128K): frontier shift + mechanism evidence under iso-quality ($\Delta=0.8$).}
  \textbf{Top row (A0--A3, three LLMs):} For each model (columns), we plot decode throughput $s$ (tok/s; \textbf{higher is better}) versus effective resident KV budget $b_{\mathrm{KV}}$ (bytes/token; \textbf{lower is better}) measured on a \textbf{paged/ragged} KV substrate.
  Let $Q^{\star}_{\mathrm{dense}}$ be the best \textbf{Dense KV} quality for that model; we \textbf{retain} operating points satisfying $Q \ge Q^{\star}_{\mathrm{dense}}-\Delta$ and color points by the quality gap $\delta Q \!=\! Q^{\star}_{\mathrm{dense}}-Q$ (green/blue/orange are within-band; gray violates the band and is excluded).
  For each ablation variant, the \textbf{black polyline} traces the \textbf{Pareto envelope} of retained points in $(b_{\mathrm{KV}}, s)$, exposing whether a change \emph{actually shifts} the \textbf{iso-quality frontier}.
  The variants instantiate the A0--A3 story: \textbf{KeepDrop} (retention-only), \textbf{Quant-only}, \textbf{Decoupled} (two-stage), and \textbf{SphKV (Joint)} plus \textbf{Angle-only} and \textbf{RD-only} contributions; the $\star$ marks the best throughput-per-byte operating point for \textbf{SphKV (Joint)} among retained configurations.
  \textbf{Bottom row (A4--A5, mechanism panels):} Segment profiles (A4; \textsf{prefix}/\textsf{retrieved}/\textsf{recent}) and stability gating (A5; \textsc{NoGate} vs.\ \textsc{Gate}) illustrate the intended \textbf{mechanism evidence}; in the final paper, these panels are replaced with the \textbf{measured} segment allocations and failure-vs-budget curves from the same paged deployment.}
  \label{fig:ablation_a0a5_l128k}
\end{figure}

\subsubsection{A0: Full Spherical KV reference}
\label{sec:appendix_a0_reference}

\paragraph{\textbf{Reference object.}}
We denote the full method by \(\textsc{SphKV}\), and use it as the causal reference against which all partial variants are compared. The full method combines four properties simultaneously:
\begin{enumerate}[leftmargin=1.35em,itemsep=2pt,topsep=2pt]
    \item \textbf{\emph{Angle-domain compute.}} Similarity is computed directly from stored radial/angular codes rather than from reconstructed dense keys.
    \item \textbf{\emph{Joint keep/drop + tiering.}} The controller jointly chooses retention and precision under a hard resident-memory budget.
    \item \textbf{\emph{Paged write contract.}} The emitted cache layout respects tier-homogeneous pages and block-local decode consumption.
    \item \textbf{\emph{No reconstruction.}} Dense key materialization is forbidden in the decode hot loop.
\end{enumerate}

\paragraph{\textbf{Reference score path.}}
At decode step \(t\), for query \(q_t\) and retained key state \(i\), the full method computes logits in compressed domain:
\[
\tilde{\ell}_{t,i}
=
f_{\mathrm{ang}}\!\big(q_t,\tilde{r}_i,\tilde{\phi}_i\big),
\]
where \(f_{\mathrm{ang}}\) is the angle-domain similarity kernel. The output of the controller is a joint assignment
\[
(z_{i,\ell,h},\, t_{i,\ell,h}),
\]
where \(z_{i,\ell,h}\in\{0,1\}\) determines whether the state is retained and \(t_{i,\ell,h}\) determines its tier (equivalently, code precision/rate). The full method is therefore the only variant in the ablation suite that simultaneously changes the \textbf{\emph{compute path}} and the \textbf{\emph{memory/control path}}.

\paragraph{\textbf{Why A0 is needed.}}
A0 is not simply a baseline row. It defines the target mechanism whose gains must be decomposed by A1--A3 and whose segment/stability behavior is later analyzed by A4--A5. Without A0, the remaining variants would only provide local comparisons rather than a coherent causal decomposition.

\begin{table}[ht!]
\centering
\caption{\textbf{Ablation summary: what each variant isolates and what witness it is expected to move.}
All variants are evaluated under the same \textbf{paged/ragged KV substrate}, the same decode settings, and the same dense-relative quality filter. The point of the table is causal: a variant matters only if it moves the retained frontier or its associated systems/stability witness in the direction predicted by the mechanism.}
\label{tab:ablation_summary}
\resizebox{\textwidth}{!}{%
\begin{tabular}{l l l l c l}
\toprule
\textbf{Variant} & \textbf{Compute path} & \textbf{Retention policy} & \textbf{Rate / tier policy} & \textbf{Kernel-realized?} & \textbf{Primary witness} \\
\midrule
\textbf{A0: Full Spherical KV} 
& Angle-domain, direct from codes 
& Joint keep/drop 
& Joint adaptive tiering 
& Yes 
& Up-left frontier shift at matched quality \\
\textbf{A1: SphKV-Recon} 
& Reconstruct-then-dot 
& Same as A0 
& Same as A0 
& No 
& Higher \(b_{\mathrm{HBM}}\), weaker tok/s at matched \(b_{\mathrm{KV}}\) \\
\textbf{A2a: KeepDrop} 
& Same as dense / baseline path 
& Adaptive keep/drop only 
& Fixed tier \(t^\star\) 
& Depends on kernel 
& Lower \(b_{\mathrm{KV}}\) with limited tok/s gain \\
\textbf{A2b: Quant-only} 
& Same as dense / baseline path 
& Keep all (\(z_i\equiv 1\)) 
& Uniform or quasi-uniform low-bit tiering 
& Depends on kernel 
& Lower resident bytes, weaker quality/traffic trade \\
\textbf{A2c: Decoupled} 
& Same as baseline path 
& Keep/drop first 
& Quantize retained set afterwards 
& Depends on kernel 
& Inferior to joint policy if coupling matters \\
\textbf{A2d: Joint} 
& Same as A0 
& Joint keep/drop 
& Joint adaptive tiering 
& Yes 
& Positive synergy gap \(\Delta_{\mathrm{joint}}(\mathcal{B})\) \\
\textbf{A3: Uniform-head} 
& Same as A0 
& Same as A0 
& Shared tier across heads 
& Yes 
& Flatter \(a_h\) allocation, weaker frontier under tight budgets \\
\textbf{A3: Adaptive-head} 
& Same as A0 
& Same as A0 
& Head-adaptive tiering 
& Yes 
& Concentrated \(a_h\), evidence of memory-keeper heads \\
\textbf{A4: NoSeg} 
& Same as A0 
& Segment-agnostic 
& Segment-agnostic 
& Yes 
& Weaker retrieved-token allocation; worse depth-conditioned QA \\
\textbf{A5: NoGate} 
& Same as A0 
& Same as A0, but no drift gate 
& Same as A0, but no drift gate 
& Yes 
& Higher failure rate, larger \(S_{\mathrm{traj}}\) / \(\Delta T\) \\
\bottomrule
\end{tabular}%
}
\end{table}

\subsubsection{A1: Angle-domain compute versus reconstruction}
\label{sec:appendix_a1_compute_vs_recon}

\paragraph{\textbf{What A1 isolates.}}
A1 isolates the \textbf{\emph{compute lever}}: does the gain come from a better consumption path through the decode kernel, or is the representation merely a compact storage format whose benefits disappear once decode begins? The comparison is between two methods that share the same compressed representation and, as far as possible, the same resident-memory profile, but differ in how similarity is computed.

\paragraph{\textbf{Two decode paths.}}
Let \((\tilde{r}_i,\tilde{\phi}_i)\) denote the stored radial and angular codes for state \(i\), and let \(q_t\) be the query at decode step \(t\).

\noindent\textbf{A1a: Direct angle-domain similarity (kernel-realized).}
In the kernel-realized path,
\[
\tilde{\ell}_{t,i}^{\mathrm{NoRecon}}
=
f_{\mathrm{ang}}(q_t,\tilde{r}_i,\tilde{\phi}_i),
\]
where \(f_{\mathrm{ang}}\) is evaluated directly inside the attention kernel. No dense key \(\tilde{k}_i\) is materialized.

\noindent\textbf{A1b: Reconstruct-then-dot (negative control).}
In the reconstruction path,
\[
\tilde{k}_i
\leftarrow
\mathrm{decode}(\tilde{r}_i,\tilde{\phi}_i),
\qquad
\tilde{\ell}_{t,i}^{\mathrm{Recon}}
=
\langle q_t,\tilde{k}_i\rangle.
\]
This variant is the negative control: it uses the same stored representation but forces decode-time densification before similarity.

\paragraph{\textbf{Traffic-level densification tax.}}
The distinction between the two paths is best understood at the level of HBM traffic. Let
\begin{itemize}[leftmargin=1.35em,itemsep=2pt,topsep=2pt]
    \item \(b_{\mathrm{read}}(\mathrm{codes})\) denote reading compressed codes from HBM,
    \item \(b_{\mathrm{write}}(\mathrm{dense}\ K/V)\) denote writing dense reconstructed states into staging buffers,
    \item \(b_{\mathrm{read}}(\mathrm{dense}\ K/V)\) denote rereading those dense states for the dot-product kernel.
\end{itemize}
Then the reconstruction path incurs the approximate traffic
\[
b_{\mathrm{HBM}}^{\mathrm{Recon}}
\;\approx\;
b_{\mathrm{HBM}}^{\mathrm{NoRecon}}
\;+\;
b_{\mathrm{read}}(\mathrm{codes})
\;+\;
b_{\mathrm{write}}(\mathrm{dense}\ K/V)
\;+\;
b_{\mathrm{read}}(\mathrm{dense}\ K/V).
\]
This expression is intentionally stronger than a generic ``extra overhead'' statement: it makes clear that the reconstruction path pays both \textbf{\emph{decode}} traffic and \textbf{\emph{re-read}} traffic for dense intermediates. In a long-context memory-bounded regime, these terms are large enough to materially affect throughput even when the resident-memory budget is similar.

\paragraph{\textbf{Interpretation of A1.}}
A1 therefore answers a very precise question:
\begin{quote}
\emph{Holding the storage representation fixed, does direct compressed-domain consumption produce a measurable systems gain over reconstruct-then-dot attention?}
\end{quote}
If the answer is yes, then the gain is not merely a representation artifact. It shows that the \textbf{\emph{compute path itself}} matters, and that the no-reconstruction rule is operationally consequential rather than stylistic.

\subsubsection{A2: Retention-only, quant-only, and joint rate--distortion control}
\label{sec:appendix_a2_joint_control}

\paragraph{\textbf{What A2 isolates.}}
A2 isolates the \textbf{\emph{control lever}}: does the gain come from selecting fewer states, from representing retained states more cheaply, or from solving these two decisions jointly under a hard budget? This is the ablation that most directly tests whether the controller is a true \textbf{\emph{rate--distortion allocator}} or merely a combination of familiar knobs.

\paragraph{\textbf{Three partial control variants and one joint variant.}}
At fixed effective budget \(\mathcal{B}\), we evaluate the following variants.

\noindent\textbf{Retention-only (\(\textsc{KeepDrop}\)).}
The controller chooses retain/drop variables \(z_i\in\{0,1\}\), but all retained states are kept at a fixed tier:
\[
t_i \equiv t^\star.
\]
Thus, the method controls \emph{which} states survive but not \emph{how precisely} they are represented.

\noindent\textbf{Quant-only (\(\textsc{Quant}\)).}
All states are retained,
\[
z_i \equiv 1,
\]
but precision is reduced uniformly or quasi-uniformly to satisfy the target budget. Thus, the method controls \emph{rate} but not \emph{residency}.

\noindent\textbf{Decoupled (\(\textsc{Decoupled}\)).}
A keep/drop rule is applied first, followed by a separate quantization step over the retained set. This is a two-stage pipeline in which the two decisions are not solved jointly.

\noindent\textbf{Joint (\(\textsc{Joint}\) / full Spherical KV).}
The controller solves for \((z_i,t_i)\) together under the same hard budget, so retention and precision are co-optimized rather than composed sequentially.

\paragraph{\textbf{A scalar functional for matched-budget comparison.}}
To compare these variants at the same effective budget \(\mathcal{B}\), we use the scalar functional
\[
\Psi(\mathcal{B})
=
q(\mathcal{B})
+
\beta \log s(\mathcal{B}),
\]
where \(q(\mathcal{B})\) is task quality, \(s(\mathcal{B})\) is throughput, and \(\beta\) is a reporting coefficient controlling the relative emphasis placed on throughput. The appendix should make clear that \(\Psi\) is a \textbf{\emph{summary functional}} rather than the optimization target of the controller itself; it is used only to compare ablation outcomes at fixed budget.

\paragraph{\textbf{Synergy gap.}}
We define the \textbf{\emph{synergy gap}} at budget \(\mathcal{B}\) as
\[
\Delta_{\mathrm{joint}}(\mathcal{B})
=
\Psi_{\textsc{Joint}}(\mathcal{B})
-
\max\Big\{
\Psi_{\textsc{KeepDrop}}(\mathcal{B}),
\Psi_{\textsc{Quant}}(\mathcal{B}),
\Psi_{\textsc{Decoupled}}(\mathcal{B})
\Big\}.
\]
This quantity is the causal heart of A2. A positive
\[
\Delta_{\mathrm{joint}}(\mathcal{B})>0
\]
means that the joint controller reaches an operating point that is not matched by any single-lever or sequential composition at the same budget.

\paragraph{\textbf{Interpretation of a positive synergy gap.}}
A positive synergy gap means more than ``the joint variant is best.'' It means that \textbf{\emph{retention and rate allocation interact non-trivially}}. In other words, deciding who to keep changes which precision decisions become optimal, and deciding precision changes which retention decisions are worthwhile. This is especially important under paged serving because metadata overhead, fragmentation, and tier-homogeneous page construction make the cost of a state depend not only on its own bits but also on how it coexists with neighboring states. Sequential composition is therefore insufficient in general: it solves the wrong optimization problem.

\paragraph{\textbf{Why this matters specifically under paged serving.}}
Under contiguous or idealized storage, one might imagine that keep/drop and bit allocation are approximately separable. Under paged/ragged serving, however, that separability breaks down. The realized cost of a state includes page headers, local tier packing, and page occupancy effects, so the true budget coupling is not token-wise independent. This is why A2 matters specifically in this paper: a positive \(\Delta_{\mathrm{joint}}\) is evidence that the observed gain is not merely a generic compression effect, but a \textbf{\emph{co-design gain under the actual serving substrate}}.

\begin{figure}[ht!]
  \centering
  \includegraphics[width=\linewidth]{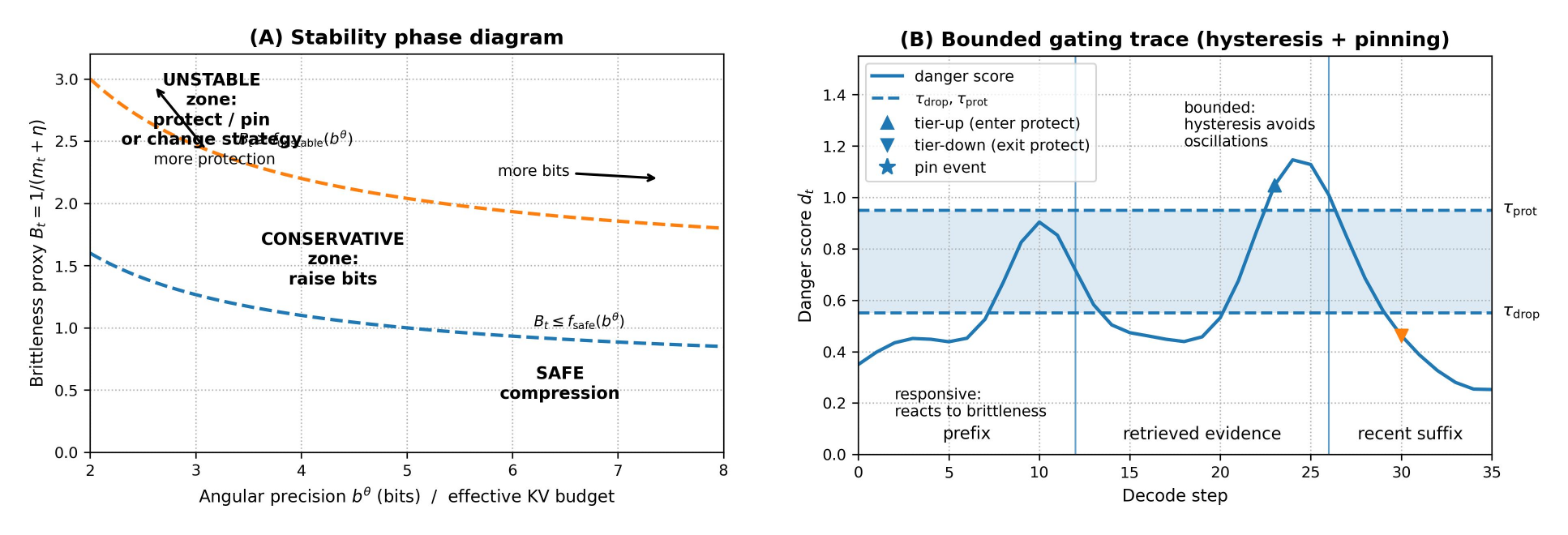}
  \caption{\textbf{Stability phase diagram and bounded gating behavior.}
  \textbf{(A) Phase diagram (policy map).} The x-axis is \textbf{angular precision} $b^\theta$ (equivalently, the effective KV budget allocated to angular codes), and the y-axis is a \textbf{brittleness proxy} $B_t$ that increases when attention becomes sensitive to small logit perturbations (e.g., inverse-margin or a normalized query-norm proxy. The plane separates three regimes: \textbf{safe compression} (low $B_t$, where down-tiering preserves trajectories), a \textbf{conservative zone} (moderate $B_t$, where the controller raises bits before dropping), and an \textbf{unstable zone} (high $B_t$, where the correct action is \textbf{protect/pin} or change strategy rather than compress).
  \textbf{(B) Decode-time gate (bounded + hysteretic).} A representative trace of the danger score $d_t$ over decode steps. Two thresholds define a hysteresis band, $\tau_{\mathrm{drop}}<\tau_{\mathrm{prot}}$: crossing the upper threshold triggers protection/tier-up, while crossing the lower threshold permits tier-down/drop. Markers indicate switch events, showing that the policy is \textbf{responsive} (reacts when brittleness spikes) yet \textbf{bounded} (avoids oscillations).
  \textbf{Takeaway.} The same stability contract that drives the controller is what we audit: when budgets tighten, the controller reallocates bits to brittle states (and pins when necessary), reducing $(S_{\mathrm{traj}},\Delta T)$ without violating the kernel-realized ``no reconstruction'' constraint.}
  \label{fig:stability_phase_gating}
\end{figure}

\subsubsection{A3: Per-head heterogeneity and adaptive tiering}
\label{sec:appendix_a3_head_heterogeneity}

\paragraph{\textbf{What A3 isolates.}}
A3 asks whether the rate--distortion controller benefits from adapting to head-level heterogeneity. Intuitively, not all heads contribute equally to long-horizon memory. Some behave as \textbf{\emph{memory-keeper}} heads whose information persists and remains useful over long spans, while others are predominantly short-range or local-processing heads. If this intuition is correct, then a uniform tiering policy should be suboptimal.

\paragraph{\textbf{Head-wise allocation.}}
For head \(h\), define the realized average bytes/token allocation
\[
a_h
=
\frac{1}{T}
\sum_{i=1}^{T}
\mathrm{bytes}\!\big(z_{i,\ell,h},\,t_{i,\ell,h}\big),
\qquad
\bar{a}
=
\frac{1}{H}
\sum_{h=1}^{H} a_h.
\]
Here \(a_h\) measures how much resident budget head \(h\) receives on average, including the realized effect of retention and tier assignment.

\paragraph{\textbf{Concentration statistics.}}
To characterize how unevenly budget is distributed across heads, we report concentration summaries over the normalized allocation vector
\[
\pi_h
=
\frac{a_h}{\sum_{j=1}^{H} a_j}.
\]
Useful summaries include:
\[
H_{\mathrm{alloc}}
=
-
\sum_{h=1}^{H}\pi_h\log \pi_h
\qquad\text{and}\qquad
G_{\mathrm{alloc}}
=
\frac{\sum_{h=1}^{H}\sum_{j=1}^{H}|a_h-a_j|}
{2H\sum_{h=1}^{H}a_h},
\]
i.e., the entropy and Gini concentration of the head-wise allocation distribution. Low entropy / high Gini indicates concentrated spending on a subset of heads.

\paragraph{\textbf{Uniform versus adaptive tiering.}}
We compare:
\begin{itemize}[leftmargin=1.35em,itemsep=2pt,topsep=2pt]
    \item \textbf{Uniform-head tiering:} \(t_{i,\ell,h}\equiv t^\star\), so all heads share the same effective rate policy;
    \item \textbf{Adaptive-head tiering:} \(t_{i,\ell,h}\) may vary with \(h\) and \(i\), so the controller can preferentially spend memory on heads that support long-horizon utility.
\end{itemize}

\paragraph{\textbf{Interpretation of heterogeneity.}}
If adaptive tiering improves the retained frontier and the allocation concentrates on a subset of heads, then the natural interpretation is that the controller has discovered \textbf{\emph{head-level functional heterogeneity}}. Heads with systematically larger \(a_h\) under tight budgets can be interpreted as \textbf{\emph{memory-keeper heads}}, whereas heads with consistently smaller \(a_h\) behave more like short-range or disposable processing channels. A3 therefore connects the systems result back to a weak mechanistic claim: the memory budget is not spread uniformly because the model itself is not uniform in how it uses context.

\subsubsection{Non-additivity and the synergy gap}
\label{sec:appendix_nonadditivity}

\paragraph{\textbf{Synthesizing A1--A3.}}
A1, A2, and A3 isolate three distinct mechanisms:
\begin{itemize}[leftmargin=1.35em,itemsep=2pt,topsep=2pt]
    \item \textbf{A1: compute lever only} — compressed-domain consumption versus reconstruction;
    \item \textbf{A2: control lever only} — retention/rate variants and their joint optimization;
    \item \textbf{A3: adaptive allocation structure} — whether head-wise heterogeneity is exploited.
\end{itemize}
Taken together, they allow the appendix to distinguish among three hypotheses:
\begin{enumerate}[leftmargin=1.35em,itemsep=2pt,topsep=2pt]
    \item the gain is explained entirely by a better \textbf{compute path};
    \item the gain is explained entirely by a better \textbf{memory-control policy};
    \item the gain is \textbf{non-additive}, i.e., it arises because the compute path and the control policy reinforce one another under paged serving.
\end{enumerate}

\paragraph{\textbf{What non-additivity means here.}}
In this paper, \textbf{\emph{non-additivity}} means that the best operating points reached by the joint system cannot be recovered by simply combining the best single-lever variants independently. Formally, a positive
\[
\Delta_{\mathrm{joint}}(\mathcal{B})>0
\]
across a meaningful budget range indicates that the joint variant lies outside the performance envelope of the partial alternatives at matched budget. At the frontier level, the same idea appears geometrically: if the full method’s retained Pareto envelope lies strictly up-left of both the compute-only and control-only envelopes, then the gain is not reducible to either lever in isolation.

\paragraph{\textbf{Strongest causal statement of D.3.}}
The strongest causal statement supported by this section is therefore:
\begin{quote}
\textbf{The frontier shift is non-additive: compressed-domain compute and joint rate--distortion control each help, but the observed long-context gain arises from their co-design under paged/ragged serving.}
\end{quote}
This is the appendix-level causal conclusion of the ablation suite.

\subsubsection{Extreme-context ablations at 128K}
\label{sec:appendix_128k_ablation_figure}

\paragraph{\textbf{Why 128K is the right stress test.}}
The existing A0--A5 figure at \(128\mathrm{K}\) is the appendix’s most demanding ablation slice. At this length, KV residency is large enough that paging, fragmentation, metadata overhead, and HBM streaming all become first-order effects. Variants that appear competitive at shorter contexts often fail here either by falling outside the iso-quality set or by losing their throughput advantage once the memory system becomes dominant. For that reason, the \(128\mathrm{K}\) figure is the right place to evaluate whether a claimed ablation effect survives the true long-context regime rather than only a mild version of it.

\vspace{2pt}
\begin{table}[ht!]
\centering
\scriptsize
\setlength{\tabcolsep}{4.4pt}
\renewcommand{\arraystretch}{1.16}
\caption{\textbf{Ablation matrix and stability witnesses under long-context decode.}
All ablations are evaluated at \(L\in\{32\mathrm{K},128\mathrm{K}\}\) and at three effective KV budgets under the same \textbf{paged/ragged serving substrate}, identical decode settings, and the same dense-relative quality tolerance. For each row, we specify the mechanism being isolated, the primary \textbf{systems-side witness}, the \textbf{quality constraint} under which the comparison is valid, and the \textbf{stability signal} used to detect whether the ablation preserves usable long-horizon behavior.}
\label{tab:ablation_deltas}
\resizebox{\linewidth}{!}{%
\begin{tabular}{l p{3.8cm} p{3.8cm} p{3.4cm} p{3.8cm}}
\hline
\textbf{Ablation} & \textbf{Mechanism isolated} & \textbf{Primary systems witness} & \textbf{Quality constraint} & \textbf{Stability witness} \\
\hline
\textbf{A1: Angle-domain vs.\ reconstruction}
& Same compressed representation, but compare \textbf{direct angle-domain similarity} against \textbf{reconstruct-then-dot} attention; isolates whether the gain comes from the \emph{compute path} rather than from storage format alone.
& \textbf{HBM bytes/token} and \textbf{decode tok/s}; the decisive question is whether removing reconstruction lowers traffic and improves steady-state throughput under the same resident budget.
& Operate inside the same \textbf{matched-quality band} relative to Dense KV, so reconstruction is not allowed to trade quality for speed.
& \(\;S_{\mathrm{traj}},\,\Delta T\;\) and failure-rate vs.\ budget, to verify that direct compressed-domain consumption does not induce brittle trajectory changes. \\
\hline

\textbf{A2: Retention-only / quant-only / decoupled / joint}
& Separates \textbf{keep/drop control}, \textbf{uniform low-bit control}, and their \textbf{two-stage composition} from the \textbf{joint rate--distortion controller}; isolates whether the observed gain is additive or genuinely co-designed.
& \textbf{Retained Pareto frontier} in \((b_{\mathrm{KV}}, s)\), together with \textbf{synergy gap} \(\Delta_{\mathrm{joint}}(\mathcal{B})\); the key test is whether the joint controller reaches operating points unattainable by any single-lever alternative.
& Task score and long-context LM metric must remain within the same dense-relative tolerance \(\Delta\), so the frontier shift is evaluated at matched utility rather than relaxed quality.
& \textbf{Failure-rate vs.\ budget}, disagreement rate, and budget-conditioned instability, to ensure that joint control does not improve averages while increasing catastrophic failures. \\
\hline

\textbf{A3: Uniform-head vs.\ adaptive-head tiering}
& Holds the overall budget fixed but compares \textbf{uniform per-head allocation} against \textbf{head-adaptive tiering}; isolates whether the controller exploits meaningful head-level heterogeneity.
& \textbf{Tok/s and retained frontier position at fixed \(b_{\mathrm{KV}}\)}, together with head-wise allocation summaries \(\{a_h\}\), entropy/Gini concentration, and evidence for a subset of \emph{memory-keeper heads}.
& Quality is compared at \textbf{matched effective bytes/token}; adaptive allocation is only credited if it improves the retained frontier without expanding the budget.
& Head-sensitive instability, including whether compressing all heads uniformly induces local failures that disappear once adaptive tiering is allowed. \\
\hline

\textbf{A4: Segment-aware vs.\ segment-agnostic allocation}
& Removes or retains \textbf{segment identity} (\textsf{prefix}/\textsf{retrieved}/\textsf{recent}) in the controller feature set; isolates whether the controller uses prompt structure in a meaningful way.
& Segment-wise retention and byte allocation
\(\rho_{\mathcal S},\,\bar b_{\mathcal S}\),
plus depth-conditioned retrieval QA behavior; the core question is whether retrieved evidence receives targeted protection when it matters.
& Mean task quality must remain inside the matched-quality band; segment awareness is only meaningful if it improves difficult positions without relying on relaxed overall quality.
& Reduced mid-context failures, fewer evidence-missing errors, and lower depth-conditioned degradation, especially in retrieval-heavy prompts associated with lost-in-the-middle behavior. \\
\hline

\textbf{A5: Stability gate on vs.\ off}
& Compares the full controller against a variant with \textbf{no margin-aware / drift-aware gate}; isolates whether predictive protection and tier escalation reduce long-horizon failures.
& \textbf{Failure-rate vs.\ budget}, together with the auxiliary cost of gating (latency overhead and traffic change); the mechanism is useful only if it improves stability without erasing the systems gain.
& No material regression in retained frontier quality; the gate is not allowed to ``solve'' instability by simply over-preserving all states.
& \(\;S_{\mathrm{traj}},\,\Delta T\;\), disagreement rate, refusal flips, tool corruption, and structured-output failures; the gate is credited only if it lowers these failure signals in the compression regime where instability is otherwise visible. \\
\hline
\end{tabular}}
\end{table}

\paragraph{\textbf{How to read the top-row frontier panels.}}
The top row of the existing figure shows retained iso-quality frontiers in the \((b_{\mathrm{KV}}, s)\) plane for the three model families at \(128\mathrm{K}\). Each black polyline is the Pareto envelope of retained points for one variant. The interpretation is straightforward:
\begin{itemize}[leftmargin=1.35em,itemsep=2pt,topsep=2pt]
    \item a local improvement means a variant helps only at particular budgets;
    \item a full up-left displacement of the envelope means the variant changes the feasible operating region;
    \item if the full \(\textsc{SphKV}\) envelope dominates the partial variants, then the observed gain is genuinely joint rather than single-lever.
\end{itemize}

\paragraph{\textbf{What the joint-vs-partial comparison certifies.}}
The comparison among A0--A3 at \(128\mathrm{K}\) certifies three things simultaneously:
\begin{enumerate}[leftmargin=1.35em,itemsep=2pt,topsep=2pt]
    \item \textbf{A1 certification:} direct compressed-domain compute reduces the densification tax relative to reconstruction;
    \item \textbf{A2 certification:} joint keep/drop + tiering outperforms retention-only, quant-only, and decoupled control at the same effective budgets;
    \item \textbf{A3 certification:} adaptive allocation structure is useful under the extreme-context regime rather than only at short lengths.
\end{enumerate}
Thus, the \(128\mathrm{K}\) ablation figure is not merely a difficult-case illustration. It is the appendix’s strongest causal evidence that the frontier shift survives precisely where the deployment problem is hardest.

\subsection{Mechanism Evidence and Stability Under Long-Horizon Decoding}
\label{sec:appendix_mechanism_stability}

\paragraph{\textbf{Purpose of this section.}}
Sections~\ref{sec:appendix_main_frontier} and \ref{sec:appendix_ablations} establish that \textbf{\emph{Spherical KV}} changes the retained memory--throughput frontier and that this gain is not reducible to a single compute-only or control-only lever. The remaining question is \textbf{\emph{behavioral}}: \emph{what does the controller actually do on structured long-context inputs, and does the resulting approximation remain stable under long-horizon decoding?} This subsection addresses that question in two complementary ways. First, it provides \textbf{\emph{mechanism evidence}} showing that the controller’s allocation is structured rather than uniform, with segment-aware and head-aware behavior that aligns with the semantics of long-context prompts. Second, it provides \textbf{\emph{stability evidence}} showing that the controller is not merely compressive, but also \textbf{\emph{protective}}: it predicts and mitigates approximation-induced failures before they cascade into trajectory-level errors. The central theme is that, in long-context decode, a compression policy is only useful if it is both \textbf{\emph{selective}} and \textbf{\emph{stable}}. 

\subsubsection{A4: Segment sensitivity and retrieval-aware allocation}
\label{sec:appendix_a4_segment_sensitivity}

\paragraph{\textbf{Why segment structure matters.}}
Long-context prompts are not homogeneous token streams. In retrieval-heavy settings, the prompt typically contains semantically distinct regions such as a \textbf{\emph{prefix}} (instructions or setup), a \textbf{\emph{retrieved block}} (documents, evidence spans, tool outputs), and a \textbf{\emph{recent suffix}} (the newest conversational or reasoning context). If a controller ignores this structure and compresses uniformly, it risks sacrificing exactly the segments that matter most for downstream answer quality. This is especially important in multi-document QA and retrieval-augmented prompting, where the answer may depend on a narrow subset of evidence tokens embedded inside a much larger context. The \textbf{``lost in the middle''} phenomenon makes this concern concrete: models are often weakest at attending to relevant information in the middle of long prompts, so a compression scheme that is segment-agnostic may amplify an already known failure mode rather than merely preserving it \citep{liu2023lost,bai2024longbench}.

\paragraph{\textbf{Segment-wise allocation statistics.}}
Let the prompt be partitioned into disjoint semantic segments
\[
\mathcal{S}\in\{\textsf{prefix},\textsf{retrieved},\textsf{recent}\}.
\]
For each segment \(\mathcal{S}\), we define the \textbf{\emph{segment-wise retention rate}}
\[
\rho_{\mathcal{S}}
=
\frac{1}{|\mathcal{S}|}
\sum_{i\in \mathcal{S}}
\mathbf{1}[z_i=1],
\]
where \(z_i\in\{0,1\}\) denotes whether token/state \(i\) is retained. This quantity captures the fraction of tokens in segment \(\mathcal{S}\) that survive the retention decision.

We also define the \textbf{\emph{segment-wise realized byte allocation}}
\[
\bar{b}_{\mathcal{S}}
=
\frac{1}{|\mathcal{S}|}
\sum_{i\in \mathcal{S}}
\mathrm{bytes}(z_i,t_i),
\]
where \(t_i\) denotes the assigned tier or precision class. Unlike \(\rho_{\mathcal{S}}\), which measures only survival, \(\bar{b}_{\mathcal{S}}\) reflects the joint effect of retention and rate allocation. Together, \((\rho_{\mathcal{S}}, \bar{b}_{\mathcal{S}})\) summarize how much of the controller’s budget is spent on each semantic region.

\paragraph{\textbf{Segment-aware versus segment-agnostic control.}}
A4 compares two variants:

\noindent\textbf{Segment-agnostic (\(\textsc{NoSeg}\)).}
The controller does not receive segment identity as an input feature. Decisions may still depend on generic signals such as age, head-local statistics, outlier behavior, or utility proxies, but there is no explicit notion of whether a token belongs to a retrieved block, a prefix instruction, or a recent suffix.

\noindent\textbf{Segment-aware (\(\textsc{SphKV}\)).}
The full controller includes segment identity in its feature space, allowing the policy to treat structurally different prompt regions differently under the same global budget.

\paragraph{\textbf{What a meaningful segment effect looks like.}}
The goal of A4 is not to show that retrieved tokens always receive the highest allocation. The goal is to show that the controller is \textbf{\emph{responsive to segment semantics}} when that distinction matters. In retrieval-heavy QA, this should manifest as:
\begin{itemize}[leftmargin=1.35em,itemsep=2pt,topsep=2pt]
    \item higher \(\rho_{\textsf{retrieved}}\) or \(\bar{b}_{\textsf{retrieved}}\) when evidence-bearing segments are genuinely needed;
    \item less aggressive compression of recent suffix tokens when current reasoning depends on recency;
    \item reduced budget on low-value prefix clutter when the prefix contains instructions but not answer-bearing detail.
\end{itemize}
The key point is that a useful controller should not compress ``the middle'' indiscriminately; it should learn to distinguish \textbf{\emph{important middle evidence}} from \textbf{\emph{disposable middle bulk}}.

\paragraph{\textbf{Connection to retrieval accuracy and lost-in-the-middle behavior.}}
For retrieval-heavy QA, the behavioral interpretation of \((\rho_{\mathcal{S}}, \bar{b}_{\mathcal{S}})\) is tied to answer quality and evidence depth. Let \(Q_{\mathrm{depth}}(d)\) denote depth-conditioned accuracy as a function of answer or evidence position \(d\). A segment-aware controller is meaningful if changes in \((\rho_{\textsf{retrieved}}, \bar{b}_{\textsf{retrieved}})\) correlate with improved \(Q_{\mathrm{depth}}(d)\) for middle- and late-position answers, rather than merely redistributing bytes without measurable task benefit. Thus A4 serves as both a \textbf{\emph{mechanism check}} and a \textbf{\emph{behavioral check}}: it tests whether the policy allocates more protection to the segments that are otherwise most vulnerable under long-context prompting \citep{liu2023lost,bai2024longbench,ho2020constructing}.

\subsubsection{A5: Stability gating and failure prediction}
\label{sec:appendix_a5_stability_gate}

\paragraph{\textbf{Why a gate is needed at all.}}
A compression policy that is optimal in expectation may still fail catastrophically on individual decode steps. In long-horizon generation, the relevant question is not only whether the average logit perturbation is small, but whether a local perturbation is large enough to change the \textbf{\emph{decision structure}} of the decode process: top-1 vs.\ top-2 ranking, tool-call selection, stopping behavior, or structured-output token emission. A5 introduces a \textbf{\emph{stability gate}} precisely to detect such high-risk events \emph{before} they become irreversible trajectory errors. The gate therefore turns the stability analysis from a post-hoc diagnostic into a \textbf{\emph{predictive control mechanism}}.

\paragraph{\textbf{Local logit-drift bound.}}
Let \(\ell_{t,i}\) be the dense-reference attention logit for state \(i\) at decode step \(t\), and let \(\tilde{\ell}_{t,i}\) be its approximated counterpart under compressed-domain representation. Suppose the radial and angular approximation errors are bounded as
\[
|r_i-\tilde{r}_i| \le \varepsilon_r,
\qquad
\|\widehat{k}_i-\widehat{\tilde{k}}_i\|_2 \le \varepsilon_\theta,
\]
where \(\widehat{k}_i\) and \(\widehat{\tilde{k}}_i\) denote normalized dense and approximate key directions, respectively. Then the induced logit perturbation obeys the conservative bound
\[
|\ell_{t,i}-\tilde{\ell}_{t,i}|
\le
\alpha\Big(
\|q_t\|_2\,\varepsilon_r
+
\|q_t\|_2\,r_i\,\varepsilon_\theta
\Big),
\]
where \(\alpha\) is the logit scaling constant. The significance of this bound is not its tightness, but its \textbf{\emph{operational meaning}}: it upper-bounds how much approximation error can perturb the local decision geometry.

\paragraph{\textbf{Margin-threatening events.}}
A local logit perturbation matters only if it threatens a relevant margin. Let
\[
m_t
=
\ell_{t,(1)}-\ell_{t,(2)}
\]
denote the gap between the top-1 and top-2 candidate logits at decode step \(t\), or more generally a task-specific margin such as the gap between a correct and incorrect tool-call action. A \textbf{\emph{margin-threatening event}} occurs when the predicted drift bound approaches or exceeds this local margin, i.e., when
\[
\alpha\Big(
\|q_t\|_2\,\varepsilon_r
+
\|q_t\|_2\,r_i\,\varepsilon_\theta
\Big)
\gtrsim
m_t.
\]
This condition means that the approximation is no longer merely small in absolute terms; it has become large relative to the local decision boundary and therefore may flip the decode choice.

\paragraph{\textbf{Gate policy.}}
The gate transforms this logic into a practical decode-time rule. Let \(d_t\) be a scalar \textbf{\emph{danger score}} derived from the ratio of predicted drift to local margin, or from a related bounded proxy. Then the policy uses two thresholds
\[
\tau_{\mathrm{drop}} < \tau_{\mathrm{prot}},
\]
creating a hysteresis band:
\begin{itemize}[leftmargin=1.35em,itemsep=2pt,topsep=2pt]
    \item if \(d_t \ge \tau_{\mathrm{prot}}\), the state is treated as \textbf{\emph{brittle}} and the controller triggers \textbf{tier escalation} or \textbf{forced retention};
    \item if \(d_t \le \tau_{\mathrm{drop}}\), the controller is allowed to \textbf{tier down} or \textbf{drop} under the ordinary budget logic;
    \item if \(d_t\in(\tau_{\mathrm{drop}},\tau_{\mathrm{prot}})\), the controller retains the previous state, preventing oscillatory thrashing.
\end{itemize}
This makes the gate \textbf{\emph{bounded}} and \textbf{\emph{hysteretic}}: it is sensitive to real danger events, but not so reactive that it flips decisions at every small perturbation.

\begin{table}[t]
\centering
\small
\setlength{\tabcolsep}{4.2pt}
\renewcommand{\arraystretch}{1.15}
\caption{\textbf{Representative operating points (the $\star$ markers) at matched quality.}
For each (LLM, $L$) panel, we report the Dense-KV best-quality reference ($Q^{\star}_{\mathrm{dense}}$) and one Spherical-KV point satisfying $Q \ge Q^{\star}_{\mathrm{dense}}-\Delta$ with $\Delta{=}0.8$.
\textbf{Speedup} is $s_{\textsc{SphKV}}/s_{\mathrm{dense}}$ at these read-offs; \textbf{KV$\downarrow$} is the reduction in resident bytes/token ($1-b_{\mathrm{KV,Sph}}/b_{\mathrm{KV,dense}}$).
\textbf{PeakKV} (GB) is computed as $b_{\mathrm{KV}}\cdot L$ (per sequence; excludes batch/parallelism and any activation memory), intended as an interpretable “how big is the cache at this point?” anchor.}
\label{tab:rep_points}
\resizebox{\linewidth}{!}{%
\begin{tabular}{l c r r r r r c r r r r r}
\toprule
\textbf{LLM} & \textbf{$L$} &
\textbf{$Q^{\star}_{\mathrm{dense}}$} & \textbf{$Q_{\textsc{SphKV}}$} & \textbf{$\delta Q$} &
\textbf{$s_{\mathrm{dense}}$} & \textbf{$s_{\textsc{SphKV}}$} & \textbf{Speedup} &
\textbf{$b_{\mathrm{KV,dense}}$} & \textbf{$b_{\mathrm{KV,Sph}}$} & \textbf{KV$\downarrow$} &
\textbf{PeakKV$_{\mathrm{dense}}$} & \textbf{PeakKV$_{\textsc{SphKV}}$} \\
\midrule
Llama-3.1-8B & 8K   & 74.20 & 73.84 & 0.37 & 173.4 & 268.5 & 1.55$\times$ & 1658.3 & 1262.3 & 23.9\% & 0.014 & 0.010 \\
Qwen2.5-14B  & 8K   & 81.35 & 81.01 & 0.34 & 151.6 & 239.4 & 1.58$\times$ & 1690.9 & 1083.7 & 35.9\% & 0.014 & 0.009 \\
GPT-oss      & 8K   & 79.71 & 79.19 & 0.52 & 163.7 & 261.2 & 1.60$\times$ & 1717.1 & 1220.3 & 28.9\% & 0.014 & 0.010 \\
\midrule
Llama-3.1-8B & 32K  & 75.65 & 75.01 & 0.64 & 110.0 & 183.0 & 1.66$\times$ & 1863.6 & 1256.6 & 32.6\% & 0.061 & 0.041 \\
Qwen2.5-14B  & 32K  & 77.56 & 76.90 & 0.66 & 93.3  & 153.9 & 1.65$\times$ & 2191.6 & 1413.4 & 35.5\% & 0.072 & 0.046 \\
GPT-oss      & 32K  & 79.39 & 79.03 & 0.35 & 104.9 & 172.5 & 1.64$\times$ & 1927.5 & 1329.8 & 31.0\% & 0.063 & 0.044 \\
\midrule
Llama-3.1-8B & 128K & 76.56 & 75.78 & 0.78 & 62.9  & 108.0 & 1.72$\times$ & 2360.6 & 1547.1 & 34.5\% & 0.309 & 0.203 \\
Qwen2.5-14B  & 128K & 78.02 & 77.40 & 0.62 & 55.7  & 94.8  & 1.70$\times$ & 2775.6 & 1607.7 & 42.1\% & 0.364 & 0.211 \\
GPT-oss      & 128K & 80.12 & 79.58 & 0.54 & 59.9  & 102.4 & 1.71$\times$ & 2442.2 & 1611.7 & 34.0\% & 0.320 & 0.211 \\
\bottomrule
\end{tabular}%
}
\end{table}

\paragraph{\textbf{Tier escalation and forced retention.}}
Operationally, the gate can intervene in two ways:
\begin{enumerate}[leftmargin=1.35em,itemsep=2pt,topsep=2pt]
    \item \textbf{Tier escalation.} Increase \(t_i\) so that the state is represented with more bits / less angular or radial distortion.
    \item \textbf{Forced retention.} Override a drop decision \(z_i=0\) and pin the state in cache when its local risk is too high.
\end{enumerate}
The important point is that both interventions are \textbf{\emph{predictive}}. The gate is not explaining failures after they occur; it is altering the memory policy when a local drift estimate signals that a future failure is plausible.

\paragraph{\textbf{What A5 is meant to prove.}}
A5 therefore tests the claim that the controller is not merely a budget allocator, but a \textbf{\emph{stability-aware budget allocator}}. If the gate reduces failure rate, disagreement, or length drift at matched \(b_{\mathrm{KV}}\), then the evidence supports the stronger statement that the controller’s stability mechanism contributes directly to the usable frontier.

\subsubsection{Trajectory sensitivity and length drift}
\label{sec:appendix_traj_length_drift}

\paragraph{\textbf{Why average quality is not enough.}}
Average quality metrics can mask severe behavioral instability. Two methods may have similar mean accuracy while differing substantially in whether they produce the same outputs under small perturbations, whether they stop at similar lengths, or whether they diverge into qualitatively different action sequences. Long-horizon decode is especially sensitive to such effects because local errors can compound over time. For this reason, we explicitly measure \textbf{\emph{trajectory sensitivity}} and \textbf{\emph{length drift}} in addition to ordinary task metrics.

\paragraph{\textbf{Trajectory sensitivity.}}
Let \(x\) denote an input prompt, let \(s\) index decoding seeds, and let \(\hat{y}_s(x)\) denote the resulting sampled completion. For a task-level metric \(m(\cdot)\), define
\[
S_{\mathrm{traj}}
=
\mathbb{E}_{x}
\Big[
\mathrm{Var}_{s}\,
m\big(\hat{y}_s(x)\big)
\Big].
\]
This quantity measures how sensitive the model’s outcome is to seed-level perturbations under the same prompt and decode settings. A low \(S_{\mathrm{traj}}\) means the method is behaviorally stable across stochastic realizations; a high \(S_{\mathrm{traj}}\) means that approximation noise or altered decode dynamics make the method fragile even if mean accuracy appears unchanged.

\paragraph{\textbf{Length drift.}}
Let \(T_{\mathrm{sph}}(x)\) and \(T_{\mathrm{dense}}(x)\) denote the generated output lengths under Spherical KV and Dense KV, respectively. We define
\[
\Delta T
=
\mathbb{E}_{x}
\Big[
\big|
T_{\mathrm{sph}}(x)-T_{\mathrm{dense}}(x)
\big|
\Big].
\]
This quantity detects changes in stopping behavior, including premature EOS, runaway continuation, or delayed termination. It is especially important because some compression-induced failures do not appear as incorrect answers but as qualitatively altered completion lengths.

\paragraph{\textbf{What the two metrics detect.}}
The two stability metrics serve different but complementary roles:
\begin{itemize}[leftmargin=1.35em,itemsep=2pt,topsep=2pt]
    \item \(S_{\mathrm{traj}}\) detects \textbf{\emph{behavioral variance}} across seeds or local perturbations;
    \item \(\Delta T\) detects \textbf{\emph{termination instability}} and altered decode horizon.
\end{itemize}
Together they capture two common long-horizon failure modes: diverging trajectories and altered stopping dynamics. A method that improves throughput but inflates either quantity is not an unambiguous gain in practice.

\subsubsection{Failure modes, disagreement, and behavioral stability}
\label{sec:appendix_failure_modes}

\paragraph{\textbf{Need for explicit failure accounting.}}
For long-horizon tasks, scalar quality summaries are often too coarse. A method may preserve average task score while increasing the frequency of specific catastrophic behaviors. We therefore supplement the frontier and stability metrics with explicit failure accounting.

\paragraph{\textbf{Disagreement rate.}}
Let \(\tau_e^{(m)}\) denote the discrete action/tool-call trajectory for episode \(e\) under method \(m\), and let \(\tau_e^{(\mathrm{dense})}\) denote the corresponding dense-reference trajectory. We define the disagreement rate as
\[
\mathrm{Disagree}
=
\frac{1}{N_{\mathrm{ep}}}
\sum_{e=1}^{N_{\mathrm{ep}}}
\mathbf{1}\!\left[
\tau_e^{(m)} \neq \tau_e^{(\mathrm{dense})}
\right].
\]
This quantity is more sensitive than final success alone: it detects behavioral divergence even when the final outcome happens to remain acceptable.

\paragraph{\textbf{Failure-rate versus budget.}}
For each budget \(\mathcal{B}\), we also report the fraction of episodes exhibiting one or more severe errors:
\[
R_{\mathrm{fail}}(\mathcal{B})
=
\frac{1}{N_{\mathrm{ep}}}
\sum_{e=1}^{N_{\mathrm{ep}}}
\mathbf{1}\!\left[
\text{episode }e\text{ exhibits failure at }\mathcal{B}
\right].
\]
Plotting \(R_{\mathrm{fail}}(\mathcal{B})\) against budget is important because some methods fail gracefully while others exhibit a sharp collapse beyond a certain compression threshold.

\paragraph{\textbf{Failure categories.}}
The appendix groups failures into interpretable categories, including:
\begin{itemize}[leftmargin=1.35em,itemsep=2pt,topsep=2pt]
    \item \textbf{\emph{refusal flips}}: a prompt that should be answered is instead refused, or vice versa;
    \item \textbf{\emph{tool corruption}}: malformed tool calls, invalid argument structure, or incorrect tool sequencing;
    \item \textbf{\emph{structured-output failures}}: JSON/schema corruption, invalid formatting, or boundary-token mismatches;
    \item \textbf{\emph{trajectory derailment}}: looping, irrelevant continuation, or premature abandonment of the task;
    \item \textbf{\emph{termination failures}}: early stop or runaway continuation.
\end{itemize}
This categorization turns stability from an abstract concern into a concrete behavioral audit.

\paragraph{\textbf{Role in the appendix argument.}}
The purpose of this subsection is to show that the stability story is not merely about smoother averages. It is about reducing the incidence of concrete, harmful long-horizon failures that matter for real deployment. In that sense, disagreement and failure accounting complement \(S_{\mathrm{traj}}\) and \(\Delta T\): the latter quantify instability statistically, while this subsection identifies how that instability manifests behaviorally.

\subsubsection{Stability phase diagram and bounded gating behavior}
\label{sec:appendix_stability_phase_diagram}

\paragraph{\textbf{Role of the existing stability figure and witness table.}}
The existing stability phase diagram and bounded-gating figure, together with the witness table, provide the visual and tabular anchor for this subsection. Their purpose is not to introduce a new headline claim, but to make the stability mechanism \textbf{\emph{inspectable}}: they show what the gate is reacting to, when it intervenes, and how those interventions map back to the failure-vs.-budget behavior discussed above. 

\paragraph{\textbf{What the phase diagram means.}}
The phase diagram should be read as a \textbf{\emph{policy map}} in a two-dimensional space whose axes encode \textbf{approximation precision} and \textbf{brittleness}. The x-axis, often rendered as angular precision or its effective bit budget, measures how much approximation error is allowed. The y-axis measures a brittleness proxy \(B_t\), which increases when local decode decisions become sensitive to perturbation. The diagram partitions this space into three qualitative regimes:
\begin{itemize}[leftmargin=1.35em,itemsep=2pt,topsep=2pt]
    \item \textbf{\emph{safe compression}}: low brittleness, where aggressive compression is unlikely to alter trajectory structure;
    \item \textbf{\emph{conservative zone}}: intermediate brittleness, where the controller should increase precision before dropping states;
    \item \textbf{\emph{unstable zone}}: high brittleness, where the correct action is to protect or pin states rather than compress further.
\end{itemize}
This figure therefore turns the abstract gate policy into a concrete geometric interpretation.

\paragraph{\textbf{What the hysteresis thresholds mean.}}
The bounded gate uses two thresholds,
\[
\tau_{\mathrm{drop}}<\tau_{\mathrm{prot}},
\]
to define a hysteresis band. The lower threshold is the point below which a state is considered safe for tier-down or drop. The upper threshold is the point above which the controller must protect or escalate precision. The interval between them is not indecision; it is an explicit stability buffer designed to prevent oscillation. Without this buffer, a state near the boundary could be repeatedly promoted and demoted across adjacent decode steps, creating exactly the kind of runtime instability the gate is supposed to prevent.

\paragraph{\textbf{Why bounded gating matters.}}
Bounded gating matters for two reasons:
\begin{enumerate}[leftmargin=1.35em,itemsep=2pt,topsep=2pt]
    \item it ensures that the controller is \textbf{\emph{responsive}} to genuinely dangerous local events;
    \item it ensures that the controller is \textbf{\emph{stable}} as a systems mechanism, avoiding thrashing and excessive control overhead.
\end{enumerate}
Thus the gate is not merely a safety heuristic. It is part of the systems design: it stabilizes the controller itself while stabilizing the model’s decode trajectory.

\paragraph{\textbf{How this ties back to the frontier and failure results.}}
The stability figure and witness table should be interpreted jointly with the frontier and failure-vs.-budget curves. If the gate lowers disagreement, reduces \(R_{\mathrm{fail}}(\mathcal{B})\), and controls \((S_{\mathrm{traj}},\Delta T)\) without materially sacrificing the retained frontier, then it has done exactly what the paper claims: it has \textbf{\emph{protected usability}} while preserving most of the systems gain. In other words, the stability mechanism matters not because it changes the frontier’s existence, but because it makes the frontier \textbf{\emph{deployable}}.

\subsection{Representative Operating Points and Reproducible Read-Offs}
\label{sec:appendix_representative_points}

\paragraph{\textbf{Purpose of this section.}}
The iso-quality frontier is the correct object for comparing methods in the long-context regime, because it captures the full trade-off among \textbf{\emph{resident KV cost}}, \textbf{\emph{HBM traffic}}, \textbf{\emph{decode throughput}}, and \textbf{\emph{quality}} under a shared serving substrate. However, reviewers also require a \textbf{\emph{concrete, reproducible anchor}}: one operating point per panel that can be inspected, reproduced, and compared numerically without re-running the full budget sweep. This subsection provides that anchor. For each \((\text{model},L)\) panel in Fig.~\ref{fig:iso_quality_frontier}, we select one \textbf{\emph{representative Spherical-KV operating point}} from the retained frontier and report it against the corresponding Dense-KV reference. The purpose is not to replace the frontier, but to make its main claim \textbf{\emph{auditable in table form}}. In this sense, Table~\ref{tab:rep_points} is the appendix’s final certification artifact: it converts a curve-level statement into nine deterministic, reproducible read-offs. :contentReference[oaicite:0]{index=0}

\subsubsection{Selection rule for representative frontier points}
\label{sec:appendix_rep_selection_rule}

\paragraph{\textbf{Why a deterministic selection rule is necessary.}}
A frontier contains many retained operating points, and the appendix should not leave the choice of representative point informal. If the representative point were chosen subjectively, the table could be challenged as cherry-picked. We therefore define the selection rule mathematically and apply it uniformly across all panels. The guiding principle is simple: the chosen point should lie on the \textbf{\emph{retained iso-quality frontier}} and should summarize the frontier using a \textbf{\emph{throughput-per-resident-byte}} criterion that is both meaningful and reproducible.

\paragraph{\textbf{Retained iso-quality set.}}
For a fixed method \(m\), model \(\mathcal{M}\), workload \(\mathcal{W}\), and context regime \(L\), let
\[
\mathcal{P}_m^\Delta
=
\Big\{
\mathcal{B}
:\;
q_m(\mathcal{B}) \ge Q_{\mathrm{dense}}^\star-\Delta
\Big\}
\]
denote the \textbf{\emph{retained iso-quality set}}, i.e., the set of budget settings whose quality remains within the dense-relative tolerance \(\Delta\). Let the corresponding projected feasible set in the frontier plane be
\[
\Pi_m^\Delta
=
\Big\{
\big(b_{\mathrm{KV},m}(\mathcal{B}),\,s_m(\mathcal{B})\big)
:\;
\mathcal{B}\in\mathcal{P}_m^\Delta
\Big\}.
\]
Let
\[
\mathrm{Env}_m^\Delta
\subseteq
\Pi_m^\Delta
\]
denote the non-dominated Pareto envelope in the \((b_{\mathrm{KV}},s)\) plane.

\paragraph{\textbf{Representative-point objective.}}
Among retained frontier points, we define the representative point as the point maximizing \textbf{\emph{throughput per resident byte}}:
\[
p_m^\star
\;\in\;
\arg\max_{p\in \mathrm{Env}_m^\Delta}
\frac{s(p)}{b_{\mathrm{KV}}(p)}.
\]
Equivalently, in budget notation,
\[
\mathcal{B}_m^\star
\;\in\;
\arg\max_{\mathcal{B}\in\mathcal{P}_m^\Delta \atop
(b_{\mathrm{KV},m}(\mathcal{B}),\,s_m(\mathcal{B}))\in \mathrm{Env}_m^\Delta}
\frac{s_m(\mathcal{B})}{b_{\mathrm{KV},m}(\mathcal{B})}.
\]
This objective is chosen deliberately. It does not select the fastest point regardless of memory, nor the smallest point regardless of speed. Instead, it selects the point that achieves the largest \textbf{\emph{throughput return per unit of resident KV budget}} among quality-matched frontier points. This makes it an appropriate summary for a paper whose central claim is a joint improvement in memory and throughput.

\paragraph{\textbf{Tie-breaking and reproducibility.}}
To make the selection fully deterministic, ties are broken in the following order:
\begin{enumerate}[leftmargin=1.35em,itemsep=2pt,topsep=2pt]
    \item choose the point with larger \(s(p)\),
    \item if still tied, choose the point with smaller \(b_{\mathrm{KV}}(p)\),
    \item if still tied, choose the point with smaller \(b_{\mathrm{HBM}}(p)\),
    \item if still tied, choose the point with the smallest budget index \(\mathcal{B}\) in the sweep order.
\end{enumerate}
This rule ensures that every panel yields \textbf{exactly one} representative point and that the same point will be recovered whenever the sweep is rerun under the same protocol.

\paragraph{\textbf{Interpretation of the \(\star\) marker.}}
The \(\star\) marker in each frontier panel is therefore not decorative. It denotes the unique operating point
\[
p_m^\star
\]
selected by the rule above. The table in Sec.~\ref{sec:appendix_rep_table} simply records the measured quantities associated with this point. In other words, the \(\star\) is the bridge between the geometric frontier claim and a concrete numerical operating point.

\subsubsection{Matched-quality read-off table across models and context lengths}
\label{sec:appendix_rep_table}

\paragraph{\textbf{Role of Table~\ref{tab:rep_points}.}}
Table~\ref{tab:rep_points} reports one \textbf{\emph{matched-quality read-off}} for each \((\text{model},L)\) panel. Each row compares two objects:
\begin{enumerate}[leftmargin=1.35em,itemsep=2pt,topsep=2pt]
    \item the \textbf{Dense-KV reference}, represented by the best dense-quality point \(Q_{\mathrm{dense}}^\star\) under the same decode settings,
    \item the \textbf{representative Spherical-KV point} \(p^\star\), selected by the deterministic rule above from the retained frontier.
\end{enumerate}
The table therefore provides a panel-by-panel numerical rendering of the frontier claim.

\paragraph{\textbf{Quantities reported in the read-off table.}}
Each row reports:
\begin{itemize}[leftmargin=1.35em,itemsep=2pt,topsep=2pt]
    \item \(Q_{\mathrm{dense}}^\star\): best Dense-KV quality under the panel’s decode settings;
    \item \(Q_{\textsc{SphKV}}\): quality at the representative Spherical-KV point;
    \item \(\delta Q = Q_{\mathrm{dense}}^\star - Q_{\textsc{SphKV}}\): explicit quality gap, which must satisfy \(\delta Q\le \Delta\);
    \item \(s_{\mathrm{dense}}\) and \(s_{\textsc{SphKV}}\): dense and Spherical-KV decode throughput;
    \item \textbf{Speedup}: \(s_{\textsc{SphKV}}/s_{\mathrm{dense}}\);
    \item \(b_{\mathrm{KV,dense}}\) and \(b_{\mathrm{KV,Sph}}\): effective resident KV budgets;
    \item \textbf{KV\(\downarrow\)}: relative resident-budget reduction;
    \item \textbf{PeakKV}: an interpretable cache-size anchor derived from resident bytes/token and context length.
\end{itemize}
Together, these quantities turn the frontier into a compact read-off that is directly auditable by reviewers.

\paragraph{\textbf{Why the table is matched-quality rather than single-point.}}
The table is not intended to report the absolute best speedup at any cost. Its purpose is narrower and more rigorous: it records \textbf{\emph{one retained iso-quality operating point per panel}}. That means every numerical claim in the table is conditioned on the same quality tolerance used by the frontier itself. In this sense, the table inherits the fairness properties of the retained frontier rather than bypassing them.

\subsubsection{How the representative points relate to the frontier}
\label{sec:appendix_rep_relation_to_frontier}

\paragraph{\textbf{Why these points are not arbitrary.}}
A common concern with representative-point tables is that they may appear cherry-picked. The present table avoids that concern for three reasons.

First, each point is chosen only from the \textbf{\emph{retained iso-quality set}}, so it is not allowed to trade away task quality to gain throughput.

Second, each point is chosen only from the \textbf{\emph{Pareto envelope}}. Therefore, it is not merely some point inside the feasible cloud; it is a point that already survives non-dominance filtering in the \((b_{\mathrm{KV}},s)\) plane.

Third, the choice among frontier points is determined by an explicit objective,
\[
\max_{p\in \mathrm{Env}_m^\Delta} \frac{s(p)}{b_{\mathrm{KV}}(p)},
\]
rather than by manual inspection. The representative points are therefore \textbf{\emph{derived}} from the frontier, not selected independently of it.

\paragraph{\textbf{How the points summarize the frontier.}}
The representative points should be interpreted as \textbf{\emph{frontier anchors}}. They do not claim to summarize every aspect of the curve, but they capture the most deployment-relevant tradeoff embodied by the frontier: how much throughput is obtained per unit of resident KV budget while staying inside the matched-quality region. In other words, the table answers the practical question:
\begin{quote}
\emph{If I want one concrete Spherical-KV configuration from this panel that best captures the memory--throughput tradeoff at matched quality, which one should I run?}
\end{quote}
That is exactly the role of a representative operating point.

\paragraph{\textbf{Why reviewers should trust them as concrete anchors.}}
Reviewers can trust these points because their selection is:
\begin{enumerate}[leftmargin=1.35em,itemsep=2pt,topsep=2pt]
    \item \textbf{quality-constrained} by \(\mathcal{P}_m^\Delta\),
    \item \textbf{frontier-constrained} by \(\mathrm{Env}_m^\Delta\),
    \item \textbf{objective-driven} through \(s/b_{\mathrm{KV}}\),
    \item \textbf{deterministic} through an explicit tie-breaking rule.
\end{enumerate}
Thus, Table~\ref{tab:rep_points} is not a replacement for the frontier; it is the reproducible numerical anchor derived from it.

\subsubsection{What the representative-point table certifies}
\label{sec:appendix_rep_certifies}

\paragraph{\textbf{What Table~\ref{tab:rep_points} certifies.}}
Taken as a whole, the representative-point table certifies five things.

\paragraph{\textbf{(i) Matched quality.}}
Every Spherical-KV row satisfies
\[
Q_{\textsc{SphKV}} \ge Q_{\mathrm{dense}}^\star - \Delta,
\]
so the throughput and memory gains are not obtained by exiting the dense-relative quality band. The \(\delta Q\) column makes this condition explicit.

\paragraph{\textbf{(ii) Throughput gain.}}
The reported speedup column
\[
\frac{s_{\textsc{SphKV}}}{s_{\mathrm{dense}}}
\]
shows that the representative Spherical-KV point is faster than the dense baseline in every retained panel. This certifies the \textbf{\emph{decode-speed}} side of the frontier shift.

\paragraph{\textbf{(iii) KV reduction.}}
The table also reports reduced
\[
b_{\mathrm{KV}}
\]
and reduced PeakKV relative to the dense reference. This certifies the \textbf{\emph{resident-memory}} side of the frontier shift.

\paragraph{\textbf{(iv) Stress-regime behavior.}}
Because the table spans
\[
L\in\{8\mathrm{K},32\mathrm{K},128\mathrm{K}\},
\]
it shows whether the gains persist into the true long-context stress regime. The \(128\mathrm{K}\) rows are particularly important: they demonstrate that the method continues to produce matched-quality read-offs with favorable speed/memory tradeoffs even where paging and HBM traffic are most severe.

\paragraph{\textbf{(v) Reproducibility.}}
Finally, the table certifies reproducibility. Each row is tied to a uniquely defined frontier point, selected by an explicit rule from a fully specified budget sweep under fixed decode settings. A reviewer or future practitioner can therefore reconstruct the same point by rerunning the same sweep and applying the same deterministic selection criterion.

\paragraph{\textbf{Closing interpretation.}}
The frontier is the correct geometric object, but Table~\ref{tab:rep_points} is the appendix’s final \textbf{\emph{numerical anchor}}. It certifies that the paper’s central claim is visible not only as an envelope-level trend, but also as a set of concrete, reproducible operating points: \textbf{\emph{matched quality, higher throughput, lower resident KV, and persistence into the long-context stress regime}}. That is why this table closes the appendix results block.

\subsection{Failure Modes and Fixes (diagnostic, not anecdotal)}
\label{sec:failure_modes}

\paragraph{\textbf{Why this section is included.}}
Matched-average quality is necessary but not sufficient for long-context deployment. In particular, two methods can appear similar under mean task score while differing sharply in \textbf{\emph{where}} and \textbf{\emph{how}} they fail: one may degrade smoothly, while another may remain strong on average yet exhibit rare but severe trajectory breaks, evidence suppression, or termination instability. This distinction matters because long-context decoding is inherently sequential: a small local perturbation can be amplified by later attention and sampling decisions into a qualitatively different completion. For this reason, we do not present failures as isolated anecdotes or qualitative curiosities. Instead, we curate a small set of \textbf{\emph{diagnostic failure classes}} and characterize each one by four objects:
\begin{enumerate}[leftmargin=1.35em,itemsep=2pt,topsep=2pt]
    \item a \textbf{\emph{minimal trigger}}: the smallest reproducible condition under which the failure appears,
    \item a \textbf{\emph{local mechanism}}: the model-side or controller-side reason the failure occurs,
    \item a \textbf{\emph{witness signal}}: an observable instability metric, such as a quality drop, \(S_{\mathrm{traj}}\!\uparrow\), \(\Delta T\!\uparrow\), or disagreement/failure-rate increase,
    \item a \textbf{\emph{controller intervention}}: a concrete fix, such as tier escalation, protected retention, or outlier-safe handling, together with its bounded systems cost.
\end{enumerate}
The purpose of this section is therefore diagnostic and causal. It shows that the safeguards in \textbf{\emph{Spherical KV}} are \textbf{\emph{predictive and actionable}}: they identify concrete failure regimes before they become widespread quality collapse, and they resolve them through localized interventions rather than global rollback of compression. :contentReference[oaicite:0]{index=0}

\paragraph{\textbf{A formal diagnostic template.}}
For clarity, we write a failure instance \(f\) as a tuple
\[
f
=
\big(
\tau_f,\;
\mu_f,\;
\sigma_f,\;
\kappa_f,\;
\phi_f
\big),
\]
where:
\begin{itemize}[leftmargin=1.35em,itemsep=2pt,topsep=2pt]
    \item \(\tau_f\) is the \textbf{\emph{minimal trigger}} (context length, budget regime, segment/head condition, and any special prompt structure),
    \item \(\mu_f\) is the \textbf{\emph{local mechanism}} responsible for the failure,
    \item \(\sigma_f\) is the \textbf{\emph{witness signal}} by which the failure is detected,
    \item \(\kappa_f\) is the \textbf{\emph{controller action}} used to mitigate it,
    \item \(\phi_f\) is the \textbf{\emph{recovery profile}} after intervention.
\end{itemize}
We represent the recovery profile as
\[
\phi_f
=
\big(
\Delta q_f,\;
\Delta S_{\mathrm{traj},f},\;
\Delta(\Delta T)_f,\;
\Delta c_f
\big),
\]
where \(\Delta q_f\) denotes the post-fix quality recovery, \(\Delta S_{\mathrm{traj},f}\) and \(\Delta(\Delta T)_f\) denote the reduction in trajectory instability and length drift, and \(\Delta c_f\) denotes the additional systems cost of the intervention (for example, extra resident bytes or a small increase in decode-time traffic). The point of this formalization is not to over-axiomatize a practical diagnostic section, but to make clear that each row of Table~\ref{tab:failure_modes} is intended as a \textbf{\emph{controlled trigger--witness--fix template}} rather than a storytelling device.

\paragraph{\textbf{Where these failures are instantiated.}}
Each failure class is instantiated at \(L\in\{32\mathrm{K},128\mathrm{K}\}\) and at a specific effective KV budget point, under the same \textbf{\emph{paged/ragged serving substrate}}, the same decode configuration, and the same matched-quality protocol used throughout the appendix. Thus, the section is aligned with the rest of the evaluation: it is not probing failure in a separate or adversarially modified environment, but in the same regime in which the frontier gains are claimed. The only difference is that this section deliberately focuses on the \textbf{\emph{boundary cases}} where local approximation error is most likely to become behaviorally consequential.

\vspace{2pt}
\begin{table}[ht!]
\centering
\scriptsize
\setlength{\tabcolsep}{4.2pt}
\renewcommand{\arraystretch}{1.16}
\caption{\textbf{Curated failure modes as reproducible diagnostic classes.}
Each row identifies a \textbf{minimal trigger}, the \textbf{local mechanism} by which the failure arises, the \textbf{observable witness} used to detect it, the \textbf{controller action} used to resolve it, and the associated \textbf{recovery cost}. All rows are instantiated at \(L\in\{32\mathrm{K},128\mathrm{K}\}\) under the same paged/ragged substrate, decode settings, and matched-quality protocol as the main frontier results.}
\label{tab:failure_modes}
\resizebox{\linewidth}{!}{%
\begin{tabular}{p{2.9cm} p{3.7cm} p{4.1cm} p{3.4cm} p{3.9cm} p{3.4cm}}
\hline
\textbf{Failure class} & \textbf{Minimal trigger} & \textbf{Local mechanism} & \textbf{Witness signal} & \textbf{Controller fix} & \textbf{Recovery cost} \\
\hline

\textbf{Retrieval distractor confusion}
&
Retrieved evidence placed in the middle of a long prompt, followed by distractor-heavy later spans, under a tight effective KV budget.
&
Evidence-bearing middle tokens lose retention or precision to later distractor-rich regions; retrieval-sensitive heads receive insufficient budget, so relevant evidence is attenuated before answer generation.
&
EM/F1 drops, \(S_{\mathrm{traj}}\uparrow\), and disagreement increases on retrieval-heavy prompts; depth-conditioned accuracy degrades most sharply for middle-position answers.
&
Protect the retrieved evidence span, promote retrieval-sensitive keeper heads, and disallow aggressive down-tiering on the evidence-bearing segment.
&
Small localized increase in resident bytes over the retrieved block; negligible global overhead relative to full rollback of compression. \\
\hline

\textbf{Small-margin step flip}
&
A critical decode step with near-tied top logits, combined with aggressive angular tiering on the heads/tokens influencing that step.
&
Approximation-induced logit drift exceeds the local decision margin, causing a branch-selection error at a low-margin reasoning step; the wrong branch then propagates forward through later decode.
&
Seed variance increases, \(S_{\mathrm{traj}}\uparrow\), and sampled continuations diverge at the affected step; wrong-answer rate rises despite only modest average logit perturbation.
&
Tier escalation for brittle tokens/heads, or forced retention when the drift bound threatens the local top-1/top-2 margin.
&
Localized precision increase on a small subset of states; bounded cost because escalation is triggered only near margin-threatening events. \\
\hline

\textbf{Outlier amplification}
&
A heavy-tail radial state or outlier-sensitive key appears under compression while outlier protection is disabled or weakened.
&
Large-radius or outlier states magnify directional approximation error, producing heavy-tailed local logit perturbations that destabilize later attention allocation.
&
Spikes in logit-drift statistics, higher trajectory disagreement, and seed-sensitive failure bursts concentrated around the outlier-bearing segment or head group.
&
Enable outlier flags, assign protected tiers to flagged states, and prevent these states from being compressed below the safe precision floor.
&
Moderate local increase in tier cost for a small subset of outlier states; stable overall frontier because protection is sparse and targeted. \\
\hline

\textbf{Termination drift}
&
High compression at long context length, especially when suffix-critical or EOS-adjacent states are aggressively dropped or down-tiered.
&
Termination-sensitive states lose enough support that stopping behavior changes: the model either exits early or continues past the dense-reference stopping point.
&
\(\Delta T\uparrow\) through early stop or rambling continuation; structured-output tasks may also show malformed endings or incomplete closure.
&
Protect suffix-critical states, escalate tier near termination-sensitive regions, and restrict aggressive compression in the terminal suffix band.
&
Slight increase in retained suffix budget; minimal throughput impact because the intervention is confined to late decode-critical states. \\
\hline

\textbf{Refusal flip / policy inversion}
&
A long-context prompt with borderline safety or refusal behavior, combined with compression near the decisive instruction/evidence boundary.
&
Approximation changes the local balance between refusal-supporting and answer-supporting evidence, causing the decode process to cross the refusal threshold in the wrong direction.
&
Unexpected refusal or unexpected compliance relative to Dense KV; increased disagreement rate and local trajectory divergence near the refusal boundary.
&
Protect instruction-critical and decision-boundary states, and apply gate-based escalation when the refusal margin becomes brittle.
&
Targeted protection of a narrow boundary region; bounded cost because the intervention affects only prompts near the refusal threshold. \\
\hline

\textbf{Tool corruption / malformed action emission}
&
Agentic or tool-augmented decode under tight budget, especially when structured tool-call tokens are generated after long reasoning prefixes.
&
Compression perturbs the short-range yet high-stakes token dependencies needed for valid tool syntax, argument ordering, or schema closure.
&
Malformed tool calls, invalid argument structure, or elevated disagreement in tool-call trajectories; success may remain stable on average while structural validity deteriorates.
&
Protect tool-schema-critical states, raise tier around action-emission windows, and gate compression when structured-output margins are small.
&
Localized increase in cost around tool-call spans; often no measurable global degradation because the protected region is short. \\
\hline

\textbf{Structured-output corruption}
&
Long structured generation (e.g., JSON / schema-constrained output) with high compression and weak suffix protection.
&
Approximation disturbs delimiter, nesting, or closure tokens that depend on a small set of syntactically critical states.
&
Schema-invalid output, delimiter mismatch, or unfinished structure; \(\Delta T\uparrow\) may appear if generation fails to terminate at the expected closing token.
&
Protect structure-critical suffix states and increase tier around delimiter-sensitive positions.
&
Bounded extra bytes in a small terminal window; substantially cheaper than globally raising precision. \\
\hline
\end{tabular}}
\end{table}

\paragraph{\textbf{Why these failure classes are sufficient.}}
The table is intentionally curated rather than exhaustive. The selected rows span the main instability families relevant to long-context serving:
\begin{enumerate}[leftmargin=1.35em,itemsep=2pt,topsep=2pt]
    \item \textbf{\emph{evidence allocation failures}} (retrieval distractor confusion),
    \item \textbf{\emph{decision-boundary failures}} (small-margin step flip, refusal flip),
    \item \textbf{\emph{outlier-amplified failures}} (outlier amplification),
    \item \textbf{\emph{termination and structure failures}} (termination drift, structured-output corruption),
    \item \textbf{\emph{action-format failures}} (tool corruption).
\end{enumerate}
This coverage is sufficient for the appendix’s purpose because it shows that the controller addresses not just one type of degradation, but the main ways in which local approximation can become a deployment-relevant behavioral error. A longer list would add examples, but not much additional structure.

\paragraph{\textbf{Why the safeguards are predictive rather than post-hoc.}}
A central point of the section is that the controller does not merely \emph{explain} failures after they occur. It acts on local signals that predict when they are likely to occur. In the language of Sec.~\ref{sec:appendix_a5_stability_gate}, the controller tracks a danger score derived from local drift bounds, margins, segment criticality, suffix sensitivity, and outlier structure. When these signals indicate that compression is approaching a brittle region of the decode process, the controller responds with a localized intervention:
\begin{itemize}[leftmargin=1.35em,itemsep=2pt,topsep=2pt]
    \item \textbf{tier escalation} when the local decision margin is threatened,
    \item \textbf{forced retention} when a state is too brittle to drop,
    \item \textbf{segment protection} when retrieved evidence would otherwise be drowned by distractors,
    \item \textbf{outlier-safe handling} when heavy-tail states would magnify approximation error,
    \item \textbf{suffix protection} when termination or structural closure becomes sensitive.
\end{itemize}
Because these interventions are triggered by predictive signals rather than post-hoc inspection, the safeguards support the stronger claim that \textbf{\emph{Spherical KV}} is a stability-aware controller, not merely a compressive one.

\paragraph{\textbf{Why the fixes are bounded rather than global.}}
The appendix should also make clear that these fixes do not amount to abandoning compression whenever a problem appears. Each controller action is deliberately \textbf{\emph{localized}}:
\begin{itemize}[leftmargin=1.35em,itemsep=2pt,topsep=2pt]
    \item retrieval protection targets a narrow evidence span,
    \item margin-based escalation targets the brittle heads/tokens responsible for a low-margin decision,
    \item outlier-safe handling targets a sparse heavy-tail subset,
    \item suffix stabilization targets a narrow terminal band.
\end{itemize}
As a result, the recovery cost \(\Delta c_f\) remains bounded. The frontier is not preserved by globally reverting to Dense KV, but by spending memory and precision where instability is concentrated. This point is essential: the controller is useful precisely because it can fix failures \textbf{\emph{without collapsing the systems gain}}.

\paragraph{\textbf{Relation to the frontier and stability sections.}}
This diagnostic section closes the loop between the frontier results and the stability analysis. The frontier sections show that \textbf{\emph{Spherical KV}} changes the feasible operating region. The stability sections show that the controller’s gate and segment-aware policies reduce \(S_{\mathrm{traj}}\), \(\Delta T\), and failure-rate under long-horizon decoding. Table~\ref{tab:failure_modes} adds the missing causal granularity: it shows \textbf{\emph{which concrete failure modes appear}}, \textbf{\emph{which local conditions trigger them}}, and \textbf{\emph{which controller actions resolve them}}. In this sense, the section is not anecdotal at all. It is the appendix’s most concrete demonstration that the controller’s safety mechanisms are \textbf{\emph{targeted, predictive, and operationally useful}}.

\paragraph{\textbf{Takeaway.}}
The correct summary of this section is:
\begin{quote}
\textbf{When compression fails, it fails in a small number of reproducible ways; when the controller succeeds, it succeeds by identifying those regimes early and applying bounded, local fixes rather than sacrificing the overall frontier.}
\end{quote}
That is why the diagnostic table belongs in the appendix: it makes the stability story falsifiable, reproducible, and actionable.

\section{Discussion, Deployment Interpretation, and Practical Guidance}
\label{sec:discussion}

This paper is best read as a \textbf{\emph{systems-real}} contribution to long-context inference, not as an isolated compression trick. The central problem is no longer simply whether one can reduce the \textbf{stored} size of the KV cache, but whether one can reduce \textbf{bytes moved per generated token} under realistic serving constraints---\textbf{\emph{paged KV layout}}, \textbf{\emph{ragged batching}}, \textbf{\emph{prefix reuse}}, and \textbf{\emph{fused decode kernels}}---without silently changing behavior or paying a hidden reconstruction tax in the hot loop. In that setting, \textbf{\emph{Spherical KV}} should be understood as a three-part deployment object: a \textbf{\emph{kernel-native compressed-domain attention path}} for reducing decode-time traffic, a \textbf{\emph{budget-explicit controller}} for allocating resident KV memory under a hard bytes/token constraint, and a \textbf{\emph{stability-audited operating protocol}} that rejects apparently fast but behaviorally brittle settings. This discussion therefore does not revisit the paper’s numerical results; instead, it interprets what those results mean operationally, why the frontier is the right evaluation object, and when the method should or should not be preferred in practice. 

\subsection{What Spherical KV contributes, in deployable terms}
\label{sec:discussion_contribution}

\paragraph{\textbf{Spherical KV is not ``just KV compression.''}}
The strongest way to state the paper’s contribution is not that it compresses the KV cache, but that it turns KV-cache optimization into an \textbf{\emph{auditable inference component}}. In deployable terms, the method contributes three coupled but conceptually distinct objects:
\begin{enumerate}[leftmargin=1.35em,itemsep=3pt,topsep=3pt]
    \item a \textbf{\emph{kernel-native compressed-domain attention path}} (\textbf{ADA}),
    \item a \textbf{\emph{budget-explicit rate--distortion controller}} (\textbf{RDR}),
    \item a \textbf{\emph{stability release protocol}} based on explicit witness tests.
\end{enumerate}
These three objects matter separately, and the results suggest that the paper’s practical value comes from their combination rather than from any one of them alone. 

\paragraph{\textbf{(i) Kernel-native compressed-domain attention.}}
The most important distinction between \textbf{\emph{Spherical KV}} and a large fraction of prior KV methods is that the representation is designed to be \textbf{\emph{consumed directly}} inside the decode kernel rather than merely stored compactly. This is the role of \textbf{\emph{Angle-Domain Attention (ADA)}}. Many compression or quantization approaches save bytes at rest but recover dense vectors before attention, which means that the realized runtime behavior depends critically on whether the engine falls back to a reconstruct/dequant/decode path. In contrast, ADA is explicitly designed around the opposite principle: \textbf{\emph{do not let the runtime silently regress to dense KV}}. In deployment terms, this is not an implementation detail but a scientific assumption: if the compressed representation is not consumed natively, then the method becomes a different algorithmic object with a different cost model and different systems claims. The paper’s results should therefore be read as evidence that \textbf{\emph{kernel-native compressed-domain attention}} is a first-class design principle for long-context serving, not merely an optimization convenience.

\paragraph{\textbf{(ii) Budget-explicit rate--distortion control.}}
The second contribution is that the memory policy is formulated as an explicit \textbf{\emph{budget allocation problem}} rather than only as a keep/drop policy. A large body of KV-cache work enforces memory indirectly through token caps, windows, sinks, or heavy-hitter eviction. These methods are often powerful, but their control primitive is fundamentally discrete: retain or discard. \textbf{\emph{Rate--Distortion Retention (RDR)}} instead treats the budget as a continuous resource and allocates it through \textbf{\emph{tiered precision decisions}} under a hard resident-memory constraint. This changes the shape of the tradeoff. Instead of asking only \emph{which} states should survive, the controller asks \emph{how precisely} each surviving state needs to be represented. In deployment terms, this matters because it replaces an all-or-nothing policy with a \textbf{\emph{graded distortion policy}}, which in turn helps explain why degradation can be smoother and more controllable than under pure token eviction. 

\paragraph{\textbf{(iii) Stability as a release condition rather than a post-hoc comment.}}
The third contribution is methodological: the paper treats \textbf{\emph{stability}} as part of the deployment contract, not as an appendix-only concern. Long-context decoding increasingly involves not just text continuation but structured programs, retrieved evidence, tool calls, and agentic loops. In those settings, a method that improves throughput while changing refusal behavior, stop reasons, seed sensitivity, schema validity, or tool-call correctness is not ready for deployment even if its mean task score looks acceptable. The witness suite introduced in the paper turns this concern into an auditable release gate: each candidate operating point must satisfy explicit metrics and exemplars, not merely aggregate averages. This is a substantive contribution because it changes the standard of evaluation from ``fast enough and accurate enough on average'' to \textbf{\emph{fast enough, accurate enough, and behaviorally acceptable at a declared operating point}}.

\paragraph{\textbf{What the paper deliberately does \emph{not} claim.}}
The practical strength of the contribution is sharpened by what the paper does not claim. We do \textbf{\emph{not}} claim that \textbf{\emph{Spherical KV}} dominates every other KV method in every regime. Token-eviction strategies can be highly effective when redundancy is high and the relevant information is tightly localized. Nor do we claim that the method is independent of kernel realism: if the runtime reconstructs dense keys in the hot loop, the end-to-end system is no longer the one analyzed in the paper. The actual claim is narrower and therefore stronger: under a declared engine, kernel path, budget definition, and witness protocol, \textbf{\emph{Spherical KV}} exposes \textbf{\emph{auditable operating points}} that satisfy both quality and stability constraints. That is the correct level of claim for a systems-real long-context method. 

\subsection{Why the frontier is the right comparison object}
\label{sec:discussion_frontier}

\paragraph{\textbf{Single-point comparisons are structurally misleading.}}
The appendix and main results are intentionally organized around \textbf{\emph{frontiers}} rather than isolated headline numbers. This is not a stylistic choice. In long-context inference, a single-point comparison is often underdetermined: a method can appear faster only because it is evaluated at a looser quality target, appear smaller only because metadata or layout effects are hidden, or appear efficient only because decode-time reconstruction costs are not charged. The correct comparison object is therefore the \textbf{\emph{feasible trade-off surface}} among:
\begin{itemize}[leftmargin=1.35em,itemsep=2pt,topsep=2pt]
    \item \textbf{\emph{resident KV cost}} (bytes/token, including metadata and layout effects),
    \item \textbf{\emph{runtime traffic}} (bytes moved per generated token),
    \item \textbf{\emph{throughput}} (steady-state decode tok/s),
    \item \textbf{\emph{quality and stability}} (task utility and behavioral acceptability).
\end{itemize}
Only a frontier can express all four simultaneously. This is why the paper treats the retained iso-quality frontier---rather than any individual speedup number---as the primary empirical object. 

\paragraph{\textbf{Why memory, traffic, and quality must be read together.}}
A long-context method improves deployment only if it simultaneously changes three connected quantities: what remains resident, what must still move through HBM, and what behavior is preserved. Reducing \textbf{\emph{resident KV bytes/token}} matters because it raises the effective memory ceiling and can improve concurrency. Reducing \textbf{\emph{HBM bytes/generated-token}} matters because long-context decode is increasingly IO-limited rather than FLOP-limited. Preserving \textbf{\emph{quality and stability}} matters because the whole point of the optimization is to produce a usable model, not merely a smaller cache. A frontier makes these interactions explicit. It forces the method to survive not only as a compact representation, but as a systems object under actual decode conditions.

\paragraph{\textbf{Why bit-budgeted tiering and token eviction behave differently.}}
One of the broader lessons suggested by the results is that \textbf{\emph{bit-tiering}} and \textbf{\emph{token eviction}} are not interchangeable control primitives. Token eviction is combinatorial: a single mistaken drop can remove a sparse but decisive span, yielding a sudden and catastrophic error even when mean performance looks stable. By contrast, tiered bit allocation tends to distribute distortion over many states and therefore often produces \textbf{\emph{more progressive degradation}} as the budget tightens. This does not make bit-tiering universally better, but it does change the geometry of the feasible region. In retrieval-heavy or tool-sensitive workloads, that difference is especially important: the preferred operating point is often the one that degrades gracefully rather than the one that is best only under mild compression. The frontier is exactly the right object for showing that distinction.

\paragraph{\textbf{Why composability is the right systems stance.}}
A second lesson is that \textbf{\emph{Spherical KV}} should be viewed as \textbf{\emph{composable}} rather than as a wholesale replacement for other serving strategies. Paging, offloading, contiguous-layout optimizations, and token-selection policies address different parts of the memory problem. In a mature deployment, the most effective stack may well be:
\[
\text{paging/offloading}
\;\rightarrow\;
\text{(optional) selection}
\;\rightarrow\;
\text{tier allocation},
\]
with witness tests serving as the integration gate. This is why the paper emphasizes the frontier as the comparison object: it naturally supports stacked systems strategies and makes clear whether a new component expands the feasible operating region or merely duplicates an existing one.

\paragraph{\textbf{The frontier as the paper’s scientific object.}}
The broader methodological point is that the paper is not about a single \(\times\) speedup claim. It is about moving the \textbf{\emph{safe operating frontier}} for long-context decode. This is a more demanding and more reproducible scientific claim, and it is the right claim for a NeurIPS systems paper.

\subsection{Kernel realism and stability are the two non-negotiable deployment constraints}
\label{sec:kernel_realism}

\paragraph{\textbf{Two constraints define whether a long-context method is deployable.}}
The results suggest that long-context KV methods should be judged against two non-negotiable deployment constraints:
\begin{enumerate}[leftmargin=1.35em,itemsep=3pt,topsep=3pt]
    \item \textbf{\emph{kernel realism}}: does the runtime actually realize the claimed memory reduction as lower decode-time traffic, rather than silently paying a reconstruction tax?
    \item \textbf{\emph{behavioral stability}}: does the operating point preserve length, stop behavior, seed stability, schema validity, tool correctness, and related behavioral invariants strongly enough for deployment?
\end{enumerate}
The paper’s systems claim only holds when both constraints are satisfied simultaneously. A method that is fast but not stable is not usable; a method that is stable but not kernel-realized is not an actual serving win. 

\paragraph{\textbf{Kernel realism is not an implementation footnote.}}
The KV literature repeatedly shows that \textbf{\emph{bytes saved at rest}} can fail to become \textbf{\emph{time saved at runtime}} if the kernels must decode, reconstruct, or otherwise materialize dense tensors before attention. This is why the paper places so much emphasis on the representation/compute-path split. \textbf{\emph{Kernel-native compute}} is not a minor engineering refinement layered on top of a representation; it is a \textbf{\emph{scientific assumption}} that determines whether the method is even operating in the intended cost model. Under strong IO-aware baselines, that assumption becomes decisive. If the implementation collapses into a dense reconstruction path, then the resulting system may still be interesting, but it is no longer the algorithmic object whose frontier is analyzed in the paper. 

\paragraph{\textbf{Stability is equally non-negotiable.}}
The second constraint is behavioral. Accuracy-only evaluation is often too weak for long-context serving because compression can perturb generation in ways that standard task metrics do not reveal: premature EOS, runaway continuation, stop-reason shifts, refusal flips, seed-amplified variance, schema corruption, and tool-call errors. These effects are particularly consequential in modern workloads involving structured program execution, retrieved evidence, and tool-augmented decoding. The paper’s witness tests matter precisely because they convert these concerns from qualitative skepticism into \textbf{\emph{checklist-style falsifiable conditions}}. This changes the interpretation of the results: a point is ``good'' not because it is fast on average, but because it is both fast and witness-passing. 

\paragraph{\textbf{Why these two constraints belong together.}}
Kernel realism and stability are often discussed separately, but the paper’s results suggest that they should be treated jointly. Kernel realism determines whether the method’s memory savings survive into wall-clock improvement. Stability determines whether the resulting operating point is behaviorally usable. Together, they define the paper’s implicit notion of a \textbf{\emph{safe operating point}}:
\[
\text{safe operating point}
\;\Longleftrightarrow\;
\text{kernel-realized}
\;\wedge\;
\text{witness-passing}.
\]
This is the real deployment lesson of the paper. The contribution is not merely a smaller cache, but a principled way to search for and certify \textbf{\emph{safe operating points in the memory-bounded long-context regime}}. 

\paragraph{\textbf{Why this changes the review standard.}}
Taken seriously, these two constraints also change how such methods should be reviewed. A systems paper should not be allowed to claim victory using payload-only compression ratios or best-case speedups without declaring the kernel path. A long-context inference paper should not be allowed to claim deployability using mean task quality alone without reporting stability witnesses. In that sense, the paper’s deeper contribution may be to sharpen the evaluation standard itself.

\subsection{Practical guidance: when Spherical KV is the right tool}
\label{sec:discussion_practical}

\paragraph{\textbf{Where the method is most compelling.}}
The results suggest that \textbf{\emph{Spherical KV}} is most attractive when at least one of the following deployment conditions holds:
\begin{enumerate}[leftmargin=1.35em,itemsep=2pt,topsep=2pt]
    \item \textbf{\emph{KV traffic dominates wall-clock}} at the target context lengths, so further reductions in bytes/token are directly valuable;
    \item \textbf{\emph{batch concurrency is limited by KV residency}}, making resident-memory reduction operationally useful even before peak context is reached;
    \item \textbf{\emph{workloads are retrieval-heavy or tool-sensitive}}, so graceful degradation and witness-gated stability matter as much as raw average accuracy;
    \item \textbf{\emph{the serving stack already uses paging/offloading}}, so a further reduction in bytes/token can compose with existing memory-management strategies.
\end{enumerate}
In these settings, the method is not merely plausible but particularly well-matched to the deployment bottleneck the paper targets. 

\paragraph{\textbf{Where the method may be less compelling.}}
Conversely, the method is less likely to be the first lever to pull when the workload remains compute-bound, when redundancy is so high that a simple token-selection policy suffices, or when the serving engine cannot support a kernel-native compressed-domain path. In such regimes, the narrow and stronger claim made by the paper becomes especially important: \textbf{\emph{Spherical KV}} is a method for the memory-bounded long-context regime under a declared kernel path, not a universal replacement for every KV policy.

\paragraph{\textbf{A reproducible protocol for choosing operating points.}}
The paper’s practical recommendation is deliberately procedural. To choose a deployable operating point, practitioners should:
\begin{enumerate}[leftmargin=1.35em,itemsep=2pt,topsep=2pt]
    \item fix a \textbf{\emph{budget target}} (bytes/token or total KV GiB),
    \item sweep tier allocations and retention settings to trace the \textbf{\emph{frontier}},
    \item filter candidate points through the \textbf{\emph{witness suite}},
    \item report only the surviving points together with the declared \textbf{\emph{kernel path}}, \textbf{\emph{engine settings}}, and \textbf{\emph{memory accounting}}.
\end{enumerate}
This protocol is intentionally conservative. It discourages best-case reporting and makes the choice of operating point reproducible across engines and workloads. 

\paragraph{\textbf{What practitioners should take away.}}
The practical lesson is therefore not ``always use Spherical KV.'' It is more specific:
\begin{quote}
\textbf{Use Spherical KV when the deployment bottleneck is KV traffic, when the engine can support kernel-native compressed-domain consumption, and when the operating point is selected from a witness-gated frontier rather than from best-case speed alone.}
\end{quote}
That is a more useful guideline than a universal recommendation because it tells practitioners how to recognize the right regime and how to evaluate the method responsibly.

\paragraph{\textbf{Bottom line.}}
The paper’s bottom-line message is that KV-cache optimization should no longer be framed as a purely representational problem. In the long-context regime, it is a problem of \textbf{\emph{deployable operating-point design}}: one must jointly reason about representation, kernel path, budget allocation, and stability. \textbf{\emph{Spherical KV}} contributes one concrete way to do that. Its value is not only that it makes the cache smaller, but that it offers a reproducible way to choose \textbf{\emph{safe, kernel-realized operating points}} in the memory-bounded regime.

\section{Limitations, Boundary Conditions, and Future Engineering Directions}
\label{sec:limitations}

Spherical KV is a \textbf{\emph{systems-real}} method. Its limitations therefore arise not only from the representation itself, but from the \textbf{\emph{kernel path}}, the \textbf{\emph{controller}}, the \textbf{\emph{workload regime}}, and the \textbf{\emph{safety envelope}} in which it is deployed. We summarize the main boundary conditions below.

\subsection{Kernel and engine dependence}
\label{sec:lim_kernel_engine}

\paragraph{\textbf{Kernel-native compute is required for realized speedups.}}
The main limitation is straightforward: if the runtime falls back to \textbf{\emph{dense reconstruction}} or dequantization-heavy staging, then the method no longer operates in the cost model claimed by the paper. In that case, resident KV bytes may still decrease, but the decode path can lose much of the wall-clock benefit. Thus, the method’s speedups are \textbf{\emph{kernel-conditional}}, not engine-independent.

\paragraph{\textbf{Performance remains engine-dependent.}}
Paged KV layout, block size, prefix reuse, batching policy, and offloading all affect fragmentation, amortization, and HBM traffic. As a result, the exact frontier can shift across engines even when the qualitative behavior remains similar. The correct interpretation is therefore not “one universal speedup,” but \textbf{\emph{engine-conditional operating curves}} under a declared serving contract.

\subsection{Proxy and controller limitations}
\label{sec:lim_proxy_controller}

\paragraph{\textbf{Rare but critical spans remain challenging.}}
The controller can fail when downstream behavior depends on a very small number of \textbf{\emph{high-stakes tokens or spans}}: a retrieved fact, a tool argument, a disambiguating entity mention, or a late-step reasoning pivot. In such cases, an imperfect proxy may be well calibrated on average yet still miss the one state that matters most.

\paragraph{\textbf{Proxy calibration is not universal.}}
The mapping from representation distortion to quality loss depends on architecture, head geometry, tokenizer behavior, and workload structure. A proxy that is effective on one benchmark or model family may transfer imperfectly to another. Thus, the controller should be treated as \textbf{\emph{calibrated, not universal}}.

\paragraph{\textbf{The controller introduces a real tuning surface.}}
Update cadence, hysteresis, protected-tier floors, and per-head/per-segment sensitivity all affect performance. Faster refresh improves responsiveness but adds overhead; slower refresh improves amortization but can miss late-tail shifts. Similarly, without hysteresis or cooldowns, adaptive tiering may oscillate. The controller is therefore effective, but not parameter-free.

\subsection{Generalization limits across workloads and architectures}
\label{sec:lim_generality}

\paragraph{\textbf{Safe operating points do not transfer automatically.}}
An operating point that is attractive at \(32\mathrm{K}\) may not remain optimal at \(128\mathrm{K}\), and a setting calibrated on one model family may move on another. Context length, head layout, and attention geometry all change the balance between compute, bandwidth, and distortion tolerance. The paper’s safer claim is therefore that operating points are \textbf{\emph{frontier-defined and witness-gated}}, not universal presets.

\paragraph{\textbf{Benchmarks do not exhaust production behavior.}}
Even a broad benchmark suite cannot fully represent production agent loops, multi-step tool chains, or deeply structured output programs. A method may remain strong on long-context QA and summarization while still failing on schema-sensitive or tool-sensitive workloads. The current evaluation is therefore strong evidence in representative regimes, but not complete production coverage.

\subsection{Safety and robustness boundaries}
\label{sec:lim_safety}

\paragraph{\textbf{Behavioral stability is not full safety.}}
The witness suite detects whether compression changes behavior in undesirable ways---for example, termination drift, refusal flips, schema corruption, or tool errors. This is important, but it is not equivalent to full adversarial robustness or end-to-end safety. A method may pass the current witness suite and still remain vulnerable to jailbreaks, poisoned retrieval, or adversarial prompt structure.

\paragraph{\textbf{The strongest defensible safety claim is narrow.}}
The method improves the \textbf{\emph{auditability of compressed decoding}} and provides a witness-gated mechanism for rejecting brittle operating points. That is valuable, but it should not be interpreted as a complete safety guarantee.

\subsection{Immediate engineering directions}
\label{sec:lim_next}

\paragraph{\textbf{Kernel hardening.}}
The most immediate next step is to harden compressed-domain attention in mainstream serving stacks, with explicit profiling of \textbf{HBM bytes/token}, page-local traffic, and reconstruction avoidance.

\paragraph{\textbf{Hybrid controllers.}}
A promising systems direction is to combine \textbf{\emph{token selection}} with \textbf{\emph{bit-tier allocation}} rather than treating them as competing alternatives. This may expand the frontier further, especially in workloads with both redundancy and sparse critical spans.

\paragraph{\textbf{Proxy refinement.}}
Future work should improve span-aware, head-aware, and learned distortion predictors, especially for rare but high-stakes regions such as tool spans, citations, and late-step reasoning tokens.

\paragraph{\textbf{Agent-grade witness suites.}}
The current witness suite should be extended toward more realistic multi-step tool and agent settings, with stronger audits for schema validity, temporal consistency, and recovery behavior.

\paragraph{\textbf{Bottom line.}}
The main limitation of Spherical KV is that its gains depend on two conditions: \textbf{\emph{kernel-native realization}} and \textbf{\emph{controller reliability}}. The main mitigation is the same principle that motivates the method: make the kernel path, budget definition, quality tolerance, and witness suite \textbf{\emph{explicit and auditable}}, and treat safe operating points as \textbf{\emph{frontier-defined}} rather than universal.

\section{Conclusion and Outlook}
\label{sec:appndx_conclusion}

Long-context decoding is increasingly constrained by \textbf{\emph{KV residency}} and \textbf{\emph{HBM traffic}}, rather than by raw arithmetic throughput alone. This paper argues that the relevant question is therefore no longer simply whether KV can be compressed, but whether it can be compressed in a way that is \textbf{\emph{kernel-realized}}, \textbf{\emph{budget-explicit}}, and \textbf{\emph{behaviorally stable}} under realistic paged serving. \textbf{\emph{Spherical KV}} addresses that regime with three coupled ingredients: a \textbf{\emph{compressed-domain attention path}} that avoids dense-key reconstruction in the decode hot loop, a \textbf{\emph{rate--distortion controller}} that jointly allocates retention and precision under a hard bytes/token budget, and a \textbf{\emph{stability gate}} that protects brittle operating points. Across models and context regimes, the resulting effect is not merely a smaller cache, but a measurable shift in the feasible long-context operating frontier.

\paragraph{\textbf{What this paper establishes.}}
\textbf{First,} the paper establishes a \textbf{\emph{kernel-realized efficiency claim}}: memory savings matter only when the compressed representation is consumed directly by the decode kernel, so that lower resident KV state becomes lower HBM traffic and higher tok/s rather than a hidden reconstruction tax. \textbf{Second,} it establishes a \textbf{\emph{matched-quality frontier claim}}: the right comparison object is the retained iso-quality Pareto envelope, and under that standard \textbf{\emph{Spherical KV}} shifts the memory--throughput tradeoff in the favorable direction. \textbf{Third,} it establishes a \textbf{\emph{stability-aware compression claim}}: long-context acceleration is only meaningful at operating points that preserve termination behavior, trajectory stability, schema validity, and related witnesses, so compression must be judged not only by average quality but by whether it remains behaviorally usable. 

\paragraph{\textbf{What broader principle this supports.}}
Taken together, these results support a broader design principle for long-context inference: \textbf{\emph{compression should be kernel-native, memory should be budgeted explicitly, and deployment points should be witness-gated rather than selected by best-case speed alone.}} This reframes KV optimization from a purely representational problem into a \textbf{\emph{safe operating-point design problem}} under realistic serving constraints.

\paragraph{\textbf{What comes next.}}
The most immediate next step is \textbf{\emph{production-grade kernel integration}}: hardened compressed-domain attention kernels inside mainstream serving engines, with explicit profiling of HBM bytes/token and reconstruction avoidance. A second direction is \textbf{\emph{richer but still auditable controller design}}: stronger span-aware or head-aware proxies that remain interpretable and witness-compatible. A third direction is \textbf{\emph{broader composability}}: extending the same page-native, kernel-realized, witness-gated logic to hybrid token-selection pipelines, cross-attention caches, multimodal long-context systems, and speculative or agentic decode stacks. 

\paragraph{\textbf{Takeaway.}}
\textbf{\emph{Long-context acceleration is a contract problem.}} A method must satisfy a \textbf{\emph{systems contract}}---paged layout, fused decode, no hidden reconstruction---and a \textbf{\emph{behavioral contract}}---matched quality, bounded instability, and witness-passing operation---before its gains are real. \textbf{\emph{Spherical KV}} is one concrete step toward that standard: it shows that reducing bytes/token is useful only when the reduction survives the kernel path and remains stable enough to deploy.

\end{document}